\pdfoutput=1
\documentclass[twoside,11pt]{article} % 
\usepackage{jmlr2e}

\usepackage{amsmath, amssymb, natbib}
\usepackage{algorithm}
\usepackage{algorithmic}
\usepackage{paralist}
\usepackage{multirow}
\usepackage{makecell}

\def \x {\mathbf{x}}
\def \c {\mathbf{c}}
\def \g {\mathbf{g}}
\def \y {\mathbf{y}}
\def \ze {\mathbf{0}}
\def \onv {\mathbf{1}}
\def \K {\mathcal{K}}
\def \X {\mathcal{X}}
\def \B {\mathcal{B}}
\def \R {\mathbb{R}}
\def \E {\mathbb{E}}
\DeclareMathOperator*{\Reg}{Reg}

\usepackage{algorithm}
\usepackage{algorithmic}
\usepackage{bm}

\def \Z {\mathcal{Z}}
\def \z {\mathbf{z}}
\def \u {\mathbf{u}}

\def \D {\mathcal{D}}
\def \v {\mathbf{v}}
\def \xx {\mathbf{\underline{x}}}

\DeclareMathOperator*{\argmin}{argmin}
\DeclareMathOperator*{\argmax}{argmax}
\DeclareMathOperator*{\Pro}{Pr}
\newtheorem{assumption}{Assumption}
\newtheorem{thm}{Theorem}
\newtheorem{cor}{Corollary}

\usepackage{times}

\begin{document}
\title{Improved Approximate Regret for Decentralized Online Continuous Submodular Maximization via Reductions}

\author{%
\name Yuanyu Wan \email wanyy@zju.edu.cn\\
\name Yu Shen \email yus@zju.edu.cn\\
\addr School of Software Technology, Zhejiang University, Ningbo, China\\
\addr State Key Laboratory of Blockchain and Data Security, Zhejiang University, Hangzhou, China\\
\name Dingzhi Yu \email yudz@lamda.nju.edu.cn\\
\addr School of Artificial Intelligence, Nanjing University, Nanjing, China\\
\name Bo Xue \email boxue4-c@my.cityu.edu.hk\\
\addr Department of Computer Science, City University of Hong Kong, Hong Kong, China\\
\name Mingli Song \email brooksong@zju.edu.cn\\
\addr School of Software Technology, Zhejiang University, Ningbo, China\\
\addr State Key Laboratory of Blockchain and Data Security, Zhejiang University, Hangzhou, China
}

\maketitle

\begin{abstract}%
To expand the applicability of decentralized online learning, previous studies have proposed several algorithms for decentralized online continuous submodular maximization (D-OCSM)---a non-convex/non-concave setting with continuous DR-submodular reward functions. However, there exist large gaps between their approximate regret bounds and the regret bounds achieved in the convex setting. Moreover, if focusing on projection-free algorithms, which can efficiently handle complex decision sets, they cannot even recover the approximate regret bounds achieved in the centralized setting. In this paper, we first demonstrate that for D-OCSM over general convex decision sets, these two issues can be addressed simultaneously. Furthermore, for D-OCSM over downward-closed decision sets, we show that the second issue can be addressed while significantly alleviating the first issue. Our key techniques are two reductions from D-OCSM to decentralized online convex optimization (D-OCO), which can exploit D-OCO algorithms to improve the approximate regret of D-OCSM in these two cases, respectively. 
% To expand the applicability of decentralized online learning, previous studies have proposed several algorithms for decentralized online continuous submodular maximization (D-OCSM)---a non-convex/non-concave case with only continuous DR-submodular reward functions. However, there exist large gaps between their approximate regret bounds and the regret bounds achieved in the convex case. Moreover, if focusing on projection-free algorithms, which can efficiently handle complex decision sets, they cannot even recover the approximate regret bounds achieved in the centralized setting. In this paper, we first demonstrate that for D-OCSM over general convex decision sets, these two issues can be addressed simultaneously. Furthermore, for D-OCSM over downward-closed decision sets, we show that the second issue can be addressed while significantly alleviating the first issue. Our key techniques are two reductions from D-OCSM to decentralized online convex optimization (D-OCO), which can exploit D-OCO algorithms to improve the approximate regret of D-OCSM in these two cases, respectively. 
\end{abstract}

% \begin{keywords}%
%   Decentralized Online Learning, Continuous Submodular Maximization, Regret Bound, Reductions%
% \end{keywords}

\section{Introduction}
Decentralized online learning (DOL) is a powerful framework for modeling various real-time distributed applications \citep{D-OCO-Survey}. In general, it is formulated as a multi-round game between an adversary and $n$ local learners in a network defined by an undirected graph $\mathcal{G}=([n],\mathcal{E})$ with the edge set $\mathcal{E}\subseteq[n]\times[n]$. At each round $t$, each local learner $i\in[n]$ first needs to choose a decision $\x_{t}^i$ from a set $\K\subseteq \R^d$, and then receives a loss or reward function $f_t^i(\x):\mathcal{X}\mapsto\R$ selected by the adversary,\footnote{As discussed in Section \ref{sec-preliminaries}, the adversary can be categorized into three types---fully adaptive, weakly adaptive, and oblivious---based on the extent to which these functions depend on the decisions. By default, we consider a fully adaptive adversary in D-OCO, but an oblivious adversary in D-OCSM.} where $\mathcal{X}\supseteq \K$. Let $f_t(\x)=\sum_{j=1}^nf_t^j(\x)$ denote the global function of round $t$. All local learners aim to maximize the cumulative global reward or minimize the cumulative global loss over the total $T$ rounds, i.e, $\sum_{t=1}^Tf_t(\x_t^i),\forall i\in[n]$. To this end, they are allowed to communicate once per round via a single step of the gossip protocol \citep{Xiao-Gossip04}, i.e., computing a weighted average of some local variables based on a given weight matrix $A\in\R^{n\times n}$. Over the past decade, there has been a growing research interest in developing and analyzing DOL algorithms \citep{DAOL_TKDE,D-ODA,wenpeng17,Wan-ICML-2020,Wan-22-JMLR,Liao-Arxiv,wang2023distributed,zhang2022communication,wan2024nearly,wan2024optimal,COLT25-Wan,Wang-ICML25}.

% To this end, they are allowed to communicate $O(1)$ times per round, each time via a single step of the gossip protocol \citep{Xiao-Gossip04}, i.e., computing a weighted average of some local variables based on a given weight matrix $A\in\R^{N\times N}$. Over the past decade, there has been a growing research interest in developing and analyzing DOL algorithms \citep{DAOL_TKDE,D-ODA,wenpeng17,Wan-ICML-2020,Wan-22-JMLR,Liao-Arxiv,wang2023distributed,zhang2022communication,wan2024nearly,wan2024optimal,COLT25-Wan}. 

The vast majority of these studies assume that the decision set and all loss functions are convex. This setting is known as decentralized online convex optimization (D-OCO), and usually chooses the regret of each local learner $i$ as the performance metric, i.e.,
\begin{equation}
\label{reg_DOCO}
\Reg(T,i)=\sum_{t=1}^Tf_{t}(\x_t^i)-\min\limits_{\x\in\K}\sum_{t=1}^Tf_{t}(\x).
\end{equation}
A nearly optimal D-OCO algorithm with ${O}(n\rho^{-1/4}\sqrt{T\log n})$ regret has been proposed \citep{wan2024nearly,wan2024optimal}, where $\rho<1$ is the spectral gap of the matrix $A$. Moreover, to handle some complex decision sets, projection-free D-OCO algorithms, which only utilize one linear optimization step per round to update each decision, have been proposed to achieve ${O}(nT^{3/4}+n\rho^{-1/4}\sqrt{T\log n})$ regret in general \citep{wan2024optimal} and ${O}(n^{5/4}\rho^{-1/2}T^{2/3})$ regret for oblivious and smooth functions \citep{wang2023distributed}. However, the objective may not be convex in many practical applications. To address this limitation, a promising attempt is to investigate DOL with  continuous DR-submodular reward functions, a rich subclass of non-convex/non-concave objectives. This setting is known as decentralized online continuous submodular maximization (D-OCSM). 

% Due to the hardness of maximizing DR-submodular functions even in the centralized offline setting \citep{bian2017guaranteed}, the performance metric for D-OCSM is changed to the $\alpha$-regret of each local learner $i$, i.e.,
Due to the inherent hardness of maximizing continuous DR-submodular functions \citep{bian2017guaranteed}, the performance metric for D-OCSM is changed to the $\alpha$-regret of each local learner $i$, i.e.,
\begin{equation}
\label{reg_DOCSM}
\Reg(T,i,\alpha)=\alpha\max\limits_{\x\in\K}\sum_{t=1}^Tf_{t}(\x)-\sum_{t=1}^Tf_{t}(\x_t^i)
\end{equation}
where $\alpha\in(0,1]$ is an approximation ratio.\footnote{One may notice that the definition in \eqref{reg_DOCSM} is different from that in \eqref{reg_DOCO} even if $\alpha=1$. This is due to the convention that D-OCO focuses on loss functions, but D-OCSM focuses on reward functions.}~
The best achievable $\alpha$ depends on the types of both functions and the decision set, where the latter is still assumed to be convex. For D-OCSM with monotone functions, the existing projection-based and projection-free algorithms can respectively achieve $O(n^{5/4}\rho^{-1/2}\sqrt{T})$ \citep{zhang2022communication} and $O(n^{5/4}\rho^{-1/2}T^{3/4})$ \citep{Liao-Arxiv} bounds on the $(1-1/e)$-regret. If functions are non-monotone, the existing projection-based and projection-free algorithms can still enjoy the same two bounds, but only for the $((1-\nu)/4)$-regret \citep{D-online-upper-LO}, where $\nu=\min_{\x\in\K}\|\x\|_{\infty}$. However, these results are unsatisfactory for two reasons. First, there exist large gaps from the regret bounds of D-OCO, especially in terms of $n$ and $\rho$. Second, if focusing on projection-free algorithms, they cannot even recover the best results achieved in the centralized setting---a special case with $n=1$, i.e., an $(1-1/e)$-regret bound of $O(T^{2/3})$ for monotone and smooth functions, and an $O(T^{2/3})$ bound on the $((1-\nu)/4)$-regret and $(1/e)$-regret for smooth functions over general convex and downward-closed decision sets, respectively \citep{pedramfar2024unified}.\footnote{We focus on computation-efficient algorithms that only require $O(1)$ (stochastic) gradients and projections (or linear optimization steps) per round. If $O(\sqrt{T})$ stochastic gradients and linear optimization steps per round are allowed, the $O(T^{2/3})$ bound can be further improved to $O(\sqrt{T})$ \citep{pedramfar2024unified}.}~Thus, it is natural to ask whether the approximate regret of D-OCSM can be further improved, thereby alleviating or even addressing these two issues.

In this paper, we provide an affirmative answer to the above question. Note that in recent advances on D-OCO \citep{wan2024nearly,wan2024optimal}, the critical technique for achieving the nearly optimal dependence on $n$ and $\rho$ is an accelerated gossip strategy \citep{Acc_Gossip,Ye2020}. 
%, which enjoys a faster average consensus among local learners than the standard gossip \citep{Xiao-Gossip04}. 
As a result, one natural attempt towards this goal is to refine the existing D-OCSM algorithms and also extend these uncovered centralized algorithms by using this technique. However, this may give rise to significant technical challenges, depending on the specific details of these algorithms. In contrast, we propose two reductions from D-OCSM to D-OCO, which allow us to achieve the desired results more elegantly. Specifically, we first show that, over general convex decision sets, the $((1-v)/4)$-regret of D-OCSM can be bounded by the regret of any D-OCO algorithm on linear losses generated by a boosting technique \citep{Zhang-Arxiv24} if it satisfies some consensus property.\footnote{It is easy to verify that such a reduction also holds for the $(1-1/e)$-regret of D-OCSM if functions are monotone. However, due to the page limit, we defer the corresponding results to Appendix \ref{sec:a_monotone_variant_of_our_first_reduction}.} Furthermore, we develop a decentralized variant of a Meta-Frank-Wolfe framework \citep{pedramfar2024unified}, which converts D-OCSM to $L$ D-OCO problems, each with $T/L$ oblivious and linear losses. We show that by using any D-OCO algorithm with an $\Reg(T)$ regret bound on $T$ oblivious and linear losses as well as some consensus property, it can achieve an $(1/e)$-regret bound of $O(nT/L+L\Reg(T/L))$ for D-OCSM with smooth functions and downward-closed decision sets. Based on these two reductions, we obtain the following substantial improvements for D-OCSM.\footnote{Note that all D-OCO algorithms used below satisfy the consensus property required by our reductions.} %, which are summarized below.
\begin{compactitem}
\item First, for general functions and decision sets, combining the first reduction with the nearly optimal D-OCO algorithm \citep{wan2024nearly,wan2024optimal} and the projection-free D-OCO algorithm in \citet{wan2024optimal}, we improve the $((1-v)/4)$-regret of projection-based and projection-free D-OCSM to ${O}(n\rho^{-1/4}\sqrt{T\log n})$ and $O(nT^{3/4}+n\rho^{-1/4}\sqrt{T\log n})$, respectively.
% derive a projection-based D-OCSM algorithm and a projection-free D-OCSM algorithm, which enjoy ${O}(n\rho^{-1/4}\sqrt{T\log n})$ and $O(nT^{3/4}+n\rho^{-1/4}\sqrt{T\log n})$ bounds on the $((1-v)/4)$-regret, respectively.
% improve the existing $O(n^{5/4}\rho^{-1/2}\sqrt{T})$ and bounds of projection-on  to . 
% derive a projection-based  algorithm 
% , which improves the existing $O(n^{5/4}\rho^{-1/2}\sqrt{T})$ bound on the $((1-v)/4)$-regret to ${O}(n\rho^{-1/4}\sqrt{T\log n})$. 
% This bound reduces the existing $O(N^{5/4}\rho^{-1/2}\sqrt{T})$ bound of projection-based D-OCSM algorithms \citep{zhang2022communication,D-online-upper-LO} in terms of $N$ and $\rho$, and bridges their gap to the nearly optimal D-OCO algorithm.
\item Second, for D-OCSM with smooth functions and general decision sets, we use the first reduction to derive a projection-free algorithm that improves the $((1-v)/4)$-regret to ${O}(nT^{2/3}+n\rho^{-1/4}\sqrt{T\log n})$. To this end, we develop an improved variant of the projection-free D-OCO algorithm in \citet{wang2023distributed}, and establish an ${O}(nT^{2/3}+n\rho^{-1/4}\sqrt{T\log n})$ regret bound on oblivious and smooth functions, as well as the linear losses in the first reduction. %This improvement to projection-free OCO algorithms may be of independent interest.
% \item Third, combining the second reduction with the nearly optimal D-OCO algorithm \citep{wan2024nearly,wan2024optimal} and  tuning the best $L$, we derive a projection-based D-OCSM algorithm with an ${O}(N(\log N)^{1/3}\rho^{-1/6}T^{2/3})$ bound on the $(1/e)$-regret for non-monotone functions and downward-closed decision sets. 
%To the best of our knowledge, it is the first D-OCSM algorithm to achieve a guarantee on the $(1/e)$-regret for these functions and decision sets.
\item Finally, for D-OCSM with smooth functions and downward-closed sets, we use the second reduction to derive a projection-free algorithm with an ${O}(n(\log (nT))^{1/3}\rho^{-1/3}T^{2/3})$ bound on the $(1/e)$-regret. This is completed by developing a new projection-free D-OCO algorithm with an ${O}(n\rho^{-1/2}\sqrt{T\log (nT)})$ regret bound on oblivious and linear losses, 
 % and the required consensus property, 
 and tuning the best $L$ in the second reduction. 
%To the best of our knowledge, it is the first D-OCSM algorithm to achieve a guarantee on the $(1/e)$-regret for these functions and decision sets. 
\end{compactitem}

\begin{table}[t]
 \centering
 \caption{Summary of our results and the previous best results on D-OCSM.  Abbreviations: convex $\to$ cvx, smooth $\to$ sm, downward-closed $\to$ dc, and Projection $\to$ Proj.}
 \label{tab1}
 \begin{tabular}{|c|c|c|c|c|}
    \hline
    Assumptions & References & $\alpha$ & $\Reg(T,i,\alpha)$ & Proj-free?\\
    \hline
    \multirow{4}{*}{\makecell{$\K$: cvx}} &\citet{D-online-upper-LO} & $(1-\nu)/4$ & $O(n^{5/4}\rho^{-1/2}\sqrt{T})$ & $\times$ \\
    \cline{2-5}
    ~ & Corollary \ref{cor1:label} & $(1-\nu)/4$ & ${O}(n\rho^{-1/4}\sqrt{T\log n})$  & $\times$\\
    \cline{2-5}
    ~ & \citet{D-online-upper-LO} & $(1-\nu)/4$ & $O(n^{5/4}\rho^{-1/2}T^{3/4})$ & $\checkmark$\\
    \cline{2-5}
    ~ & Corollary \ref{cor2:label} & $(1-\nu)/4$ & ${O}(nT^{3/4}+n\rho^{-1/4}\sqrt{T\log n})$ & $\checkmark$\\
    % \hline
    % \multirow{4}{*}{\makecell{$f_{t}^i(\x)$: non-mte\\$\K$: cvx}} &\citet{D-online-upper-LO} & $(1-\nu)/4$ & $O(n^{5/4}\rho^{-1/2}\sqrt{T})$ & $\times$\\
    % \cline{2-5}
    % ~ &Theorem xxx & $(1-\nu)/4$ & ${O}(n\rho^{-1/4}\sqrt{T\log n})$ & $\times$\\
    %  \cline{2-5}
    % ~ &\citet{D-online-upper-LO} & $(1-\nu)/4$ & $O(n^{5/4}\rho^{-1/2}T^{3/4})$ &  $\checkmark$\\
    %  \cline{2-5}
    % ~ &Theorem xxx & $(1-\nu)/4$ & ${O}(nT^{2/3})$  &  $\checkmark$\\
    \hline
    {\makecell{$f_{t}^i(\x)$: sm\\$\K$: cvx}} &Corollary \ref{cor3:label}& $(1-\nu)/4$ & ${O}(nT^{2/3}+n\rho^{-1/4}\sqrt{T\log n})$ &  $\checkmark$\\
    \hline
    {\makecell{$f_{t}^i(\x)$: sm \\$\K$: cvx \& dc}}  &Corollary \ref{cor5} & $1/e$ & ${O}(n(\log (nT))^{1/3}\rho^{-1/3}T^{2/3})$&  $\checkmark$\\
    %  \multirow{2}{*}{\makecell{$f_{t}^i(\x)$: sm \\$\K$: cvx \& dc}} &Corollary \ref{cor4} & $1/e$ & ${O}(n(\log n)^{1/3}\rho^{-1/6}T^{2/3})$ &  $\times$\\
    % \cline{2-5}
    % ~ &Corollary \ref{cor5} & $1/e$ & ${O}(n(\log (nT))^{1/3}\rho^{-1/3}T^{2/3})$&  $\checkmark$\\
    \hline
 \end{tabular}
\end{table}

A detailed comparison between our results and previous studies on D-OCSM is presented in Table \ref{tab1}. For general decision sets, the approximate regret bounds of our D-OCSM algorithms have matched the best regret bounds achieved in D-OCO, and can recover the best results achieved in the centralized setting. This implies that the two aforementioned issues are addressed simultaneously. For downward-closed decision sets, our projection-free D-OCSM algorithm achieves the first guarantee on the $(1/e)$-regret, and addresses the second aforementioned issue while significantly alleviating the first one. Moreover, our two projection-free D-OCO algorithms for oblivious smooth and linear losses may be of independent interest.

\section{Related Work}
In this section, we provide a brief review of related work on D-OCO and D-OCSM, covering both the special case with $n=1$ and the general case.

\subsection{Decentralized Online Convex Optimization (D-OCO)}
% D-OCO with $N=1$ reduces to the online convex optimization (OCO) problem, which has been extensively studied since the pioneering work of \citet{Zinkevich2003}. It is well-known that the optimal regret bound of OCO is $O(\sqrt{T})$ \citep{Abernethy08}, and can be achieved by two classical algorithms: online gradient descent (OGD) \citep{Zinkevich2003} and follow-the-regularized-leader (FTRL) \citep{Shai07,Hazan_2007}. However, both OGD and FTRL require a projection operation in each round, which could be time-consuming for complex decision sets. Intriguingly, if loss functions are linear and weakly adaptive, there exists a projection-free algorithm called follow-the-perturbed-leader (FTPL) \citep{FTPL06}, which enjoys the $O(\sqrt{T})$ regret bound by only using one linear optimization step per round. 

% To efficiently handle the general OCO with complex decision sets, \citet{Hazan2012} propose a projection-free algorithm called online Frank-Wolfe (OFW), which has an $O(T^{3/4})$ regret bound. The key idea is to approximate each update of FTRL via one linear optimization step. Later, various projection-free OCO algorithms have been proposed to achieve improved regret bounds by exploiting special properties of loss functions and decision sets \citep{Garber16,hazan2020faster,SC_OFW,Garber_SOFW,zak-arXiv-22,Garber22,Garber-COLT23}. The most relevant one to this paper is the online smooth projection-free (OSPF) algorithm \citep{hazan2020faster}, which is an extended version of FTPL and can achieve an $O(T^{2/3})$ regret bound for oblivious and smooth functions.
D-OCO with $n=1$ reduces to online convex optimization (OCO), which has been extensively studied since \citet{Zinkevich2003}. It is well-known that the optimal regret bound of OCO is $O(\sqrt{T})$ \citep{Abernethy08}, and can be achieved by two classical algorithms:~online gradient descent (OGD) \citep{Zinkevich2003} and follow-the-regularized-leader (FTRL) \citep{Shai07,Hazan_2007}. However, both OGD and FTRL require a projection operation in each round, which could be time-consuming for complex decision sets. If functions are linear and weakly adaptive, there exists a projection-free algorithm called follow-the-perturbed-leader (FTPL) \citep{FTPL06}, which enjoys the $O(\sqrt{T})$ regret bound by only using one linear optimization step per round. For the general OCO, the first projection-free algorithm is online Frank-Wolfe (OFW) \citep{Hazan2012}, which has an $O(T^{3/4})$ regret bound. The key idea is to approximate each update of FTRL via one linear optimization step. For oblivious and smooth functions, \citet{hazan2020faster} propose an online smooth projection-free (OSPF) algorithm with an $O(T^{2/3})$ regret bound, which is an extended version of FTPL.

% Later, various projection-free OCO algorithms have been proposed to achieve improved regret bounds by exploiting special properties of loss functions and decision sets \citep{Garber16,hazan2020faster,SC_OFW,Garber_SOFW,zak-arXiv-22,Garber22,Garber-COLT23}. The most relevant one to this paper is the online smooth projection-free (OSPF) algorithm \citep{hazan2020faster}, which is an extended version of FTPL and can achieve an $O(T^{2/3})$ regret bound for oblivious and smooth functions.
% The general D-OCO is an extension of OCO for exploiting locally light computations in a network. 
Compared with OCO, the critical challenge of the general D-OCO is that each local learner no longer has direct access to the global function. To tackle this challenge, \citet{DAOL_TKDE} propose a decentralized variant of OGD, and achieve an $O(n^{5/4}\rho^{-1/2}\sqrt{T})$ regret bound. Their key idea is to first apply the standard gossip step \citep{Xiao-Gossip04} over the decisions of local learners, and then perform a gradient descent step according to the local function. After that, many other OCO algorithms, including FTRL, FTPL, OFW, and OSPF, have been extended into D-OCO by incorporating the standard gossip step \citep{D-ODA,wenpeng17,Wan-ICML-2020,Wan-22-JMLR,wang2023distributed}. However, none of them can improve the regret bound of D-OGD. Very recently, \citet{wan2024nearly,wan2024optimal} propose an accelerated decentralized variant of FTRL (AD-FTRL) with an improved regret bound of $O(n\rho^{-1/4}\sqrt{T\log n})$, and demonstrate its near optimality by establishing an $\Omega(n\rho^{-1/4}\sqrt{T})$ lower bound. The critical idea of AD-FTRL is to exploit the accelerated gossip strategy \citep{Acc_Gossip,Ye2020}, which enjoys a faster average consensus among local learners than the standard gossip step. Furthermore, \citet{wan2024optimal} have also developed a projection-free variant of AD-FTRL, which can achieve an $O(n{T}^{3/4}+n\rho^{-1/4}\sqrt{T\log n})$ regret bound. In contrast, the previous decentralized variants of OFW \citep{wenpeng17,Wan-ICML-2020,Wan-22-JMLR} only achieve an $O(n^{5/4}\rho^{-1/2}{T}^{3/4})$ regret bound. We also notice that the decentralized OSPF (D-OSPF) in \citet{wang2023distributed} achieves an $O(n^{5/4}\rho^{-1/2}T^{2/3})$ regret bound for oblivious and smooth functions.

\subsection{Decentralized Online Continuous Submodular Maximization (D-OCSM)}
% D-OCSM with $N=1$ reduces to online continuous submodular maximization (OCSM), which has received considerable attention recently \citep{chen2018projection,chen2018online,zhang2019online,Thang-AAAI21,zhang2022stochastic,Niazadeh-MS23,Mualem-AISTATS23,Liao-Arxiv,Zhang-AISTATS23,Zhang-Arxiv24,pedramfar2024unified,Pedramfar-NIPS24}. For brevity, we only discuss two categories of algorithms, which are the most related to this work. The first category includes online gradient ascent (OGA) and its variants. OGA is a naive counterpart of OGD for OCSM, and \citet{chen2018online} first show that it can use stochastic gradients to achieve an $O(\sqrt{T})$ bound on the $(1/2)$-regret for monotone functions and general convex sets. A boosting variant of OGA (BOGA) is proposed by \citet{zhang2022stochastic}, which can achieve the same bound on the $(1-1/e)$-regret for these functions and sets. Their key idea is to run OGA over an auxiliary function $F_t(\x)$ such that $\langle\nabla F_t(\x),\y-\x\rangle\geq (1-1/e) f_t(\y)-f_t(\x)$, and its stochastic gradient can be estimated from a stochastic gradient of the original function. This is the so-called non-oblivious boosting technique. \citet{Liao-Arxiv} propose a projection-free variant of BOGA, while relaxing the $(1-1/e)$-regret bound to $O(T^{3/4})$. Moreover, \citet{Zhang-Arxiv24} generalize BOGA into the case with non-monotone functions and general convex sets, and establish an $O(\sqrt{T})$ bound on the $((1-v)/4)$-regret.
D-OCSM with $n=1$ reduces to online continuous submodular maximization (OCSM), which has also received considerable attention recently \citep{chen2018projection,chen2018online,zhang2019online,Thang-AAAI21,zhang2022stochastic,Niazadeh-MS23,Mualem-AISTATS23,Liao-Arxiv,Zhang-AISTATS23,Zhang-Arxiv24,pedramfar2024unified,Pedramfar-NIPS24}. For brevity, we only discuss two categories of algorithms, which are the most related to this work. The first category includes online gradient ascent (OGA) and its variants. OGA is a naive counterpart of OGD for OCSM, and \citet{chen2018online} first show that it can use stochastic gradients to achieve an $(1/2)$-regret bound of $O(\sqrt{T})$ for monotone functions. A boosting variant of OGA (BOGA) is proposed by \citet{zhang2022stochastic}, which can achieve an $(1-1/e)$-regret bound of $O(\sqrt{T})$ for these functions. Their key idea is to run OGA over an auxiliary function $F_t(\x)$ such that $\langle\nabla F_t(\x),\y-\x\rangle\geq (1-1/e) f_t(\y)-f_t(\x)$, and its stochastic gradient can be estimated from a stochastic gradient of the original function. \citet{Liao-Arxiv} propose a projection-free variant of BOGA, while relaxing the $(1-1/e)$-regret bound to $O(T^{3/4})$. Moreover, \citet{Zhang-Arxiv24} generalize BOGA into the case with non-monotone functions, and establish an $((1-v)/4)$-regret bound of $O(\sqrt{T})$.

The second category includes several versions of a Meta-Frank-Wolfe (MFW) algorithm, and their guarantees further rely on the smoothness of the functions. Specifically, \citet{chen2018online} propose the initial version of MFW, and achieve an $(1-1/e)$-regret bound of $O(\sqrt{T})$ for monotone functions. Later, a series of extensions for MFW are proposed to handle stochastic gradients \citep{chen2018projection}, and non-monotone functions \citep{Zhang-AISTATS23,Mualem-AISTATS23}, while achieving the $O(\sqrt{T})$ bound on specific approximation ratios. However, MFW and these extensions require much more than $O(1)$ (stochastic) gradients and linear optimization steps at each round. To make MFW computation-efficient, \citet{zhang2019online} combine the stochastic version \citep{chen2018projection} with a blocking update mechanism, which allocates the original computational costs of one round into every round of a block. In this way, they achieve an $(1-1/e)$-regret bound of $O(T^{4/5})$ for monotone functions. \citet{pedramfar2024unified} propose a generalized and improved variant of this computation-efficient MFW, which reduces the $(1-1/e)$-regret bound to $O(T^{2/3})$. For non-monotone functions over general and downward-closed decision sets, it can enjoy the $O(T^{2/3})$ bound on the $((1-\nu)/4)$-regret and $(1/e)$-regret, respectively. 
% Note that these two computation-efficient versions of MFW are truly projection-free. 
Compared with \citet{zhang2019online}, the critical change of \citet{pedramfar2024unified} is to simply use each stochastic gradient, instead of combining a variance reduction technique. 

The general D-OCSM is first investigated by \citet{zhu2021projection}, who propose a decentralized variant of the stochastic MFW in \citet{chen2018projection}. However, this decentralized algorithm inherits the computational issue, and requires much more than $O(1)$ communications at each round, which violates the standard communication constraint. To address these issues, \citet{zhang2022communication} propose a decentralized variant of the first computation-efficient MFW in \citet{zhang2019online}, and achieve an $(1-1/e)$-regret bound of $O(n\rho^{-1/2}T^{4/5}+n^{3/2}\rho^{-1}T^{2/5})$ for smooth and monotone functions. Moreover, \citet{zhang2022communication} also propose a decentralized variant of BOGA (D-BOGA), which reduces the $(1-1/e)$-regret to $O(n^{5/4}\rho^{-1/2}\sqrt{T})$ even for non-smooth and monotone functions. Later, \citet{Liao-Arxiv} propose a projection-free variant of D-BOGA, which achieves an $(1-1/e)$-regret bound of $O(n^{5/4}\rho^{-1/2}T^{3/4})$ for the same type of functions. \citet{D-online-upper-LO} generalize D-BOGA and the projection-free variant into the case with non-monotone functions. They keep the original bounds of these two algorithms, but only for the $((1-\nu)/4)$-regret.

% \subsection{Discussions}
\paragraph{Discussions}
First, besides the specific algorithms based on the boosting technique, \citet{Pedramfar-NIPS24} recently provide a reduction from the approximate regret of OCSM to the regret of OCO on linear losses generated by this technique. The reduction can be simply derived from the  property of the auxiliary function $F_t(\x)$. However, in the decentralized setting, the exploitation of this technique is still limited to combining with OGA-type algorithms. In this paper, our first reduction will address this limitation, and bring significant improvements to D-OCSM.  
%Although our first reduction is a natural extension of this result, the benefits for D-OCSM . 
Second, we notice that all the existing versions of MFW for both OCSM and D-OCSM are actually a general framework that converts the original problem to several OCO problems with some oblivious and linear losses. The type of losses implies that FTPL \citep{FTPL06} is sufficient to solve these OCO problems with the desired regret. For this reason, they can update the decision by only using linear optimization steps, and the computation-efficient versions \citep{zhang2019online,pedramfar2024unified,zhang2022communication} are truly projection-free, i.e., one linear optimization step per round. However, there exists a gap between the current best MFW for D-OCSM \citep{zhang2022communication} and OCSM \citep{pedramfar2024unified}. 
% Moreover, it is actually more difficult to extend the version of \citet{pedramfar2024unified} into the decentralized setting because the variance reduction technique is not used. 
In this paper, our decentralized variant of MFW will fill this gap, and provide a novel reduction from D-OCSM to D-OCO, instead of OCO.

% , it is easy to verify that the power of the non-oblivious boosting technique is actually beyond only combining OGA-type algorithms. More specifically, have provided . Our first reduction is inspired .

\section{Main Results}
We first introduce the necessary preliminaries of D-OCSM, including some notations and assumptions, and then present our two reductions as well as the improved approximate regret bounds. All proofs can be found in the appendix.
\subsection{Preliminaries}
\label{sec-preliminaries}
Let $\mathbf{e}_i$ denote the $i$-th basic vector, and $\|\cdot\|$ denote the Euclidean norm. For any $\x,\y\in \mathbb{R}^d$, we use $\x\leq \y$ to denote that any components of $\x$ are not larger than the corresponding components of $\y$. D-OCSM usually requires the following assumptions.
% Following previous studies on D-OCSM \citep{zhang2022communication,Liao-Arxiv,D-online-upper-LO} and D-OCO \citep{wan2024nearly,wan2024optimal}, we introduce the following assumptions.
\begin{assumption}
\label{assum:bounded-set}
The feasible set 
% $\mathcal{X}$ contains the origin, i.e., $\mathbf{0}\in\mathcal{X}$, and 
satisfies $\mathcal{X} = \Pi_{i=1}^{d}\mathcal{X}_i \in \mathbb{R}_+^d$, where $\mathcal{X}_i=[0,1]$. Moreover, the decision set $\K\subseteq \mathcal{X}$ is convex, and its radius is bounded by $R$, i.e., $\|\x\|\leq R$ for any $\x\in\K$.
% Moreover, the decision set $\K\subseteq \mathcal{X}$ contains the origin, i.e., $\mathbf{0}\in\K$, and its radius is bounded by $R$, i.e., $\|\x\|\leq R$ for any $\x\in\K$.
\end{assumption}
\begin{assumption}
\label{assum:DR-submodular}
At each round $t\in[T]$, the reward function $f_t^i(\x)$ of each learner $i\in[n]$ is continuous DR-submodular and non-negative over $\mathcal{X}$, i.e., i) $
  f_t^i(\x+z\mathbf{e}_j) - f_t^i(\x) \ge f_t^i(\y+z\mathbf{e}_j) - f_t^i(\y)$, for any $\x\leq \y \in \mathcal{X}$, and any $z\in\mathbb{R}_+$, $j\in[d]$ such that $\x+z\mathbf{e}_j$ and $\y+z\mathbf{e}_j$ still belong to $\mathcal{X}$; ii) $f_t^i(\x)\geq 0$ for any $\x\in \mathcal{X}$. 
\end{assumption}
\begin{assumption}
  \label{assumption3} At each round $t\in[T]$, each learner $i\in[n]$ can query the stochastic gradient $\tilde{\nabla}f^i_t(\x)$, which is unbiased and bounded, i.e., $\mathbb{E}[\tilde{\nabla}f_{t}^i(\x)|\x] = \nabla f_{t}^i(\x)\text{~and~}\Vert \tilde{\nabla}f_{t}^i(\x)\Vert \le G$, for any $\x\in \mathcal{X}$. Moreover, all reward functions are chosen by an oblivious adversary.
\end{assumption}
\begin{assumption}
  \label{assumption4}
  The communication matrix $A$ is supported on the graph $\mathcal{G} = ([n], \mathcal{E})$, symmetric, and doubly stochastic, i.e., i) $A_{ij}>0$ only if $(i,j)\in \mathcal{E}$ or $i=j$; ii) $A^\top = A$; and iii) $\sum_{i\in[n]}A_{ij}=\sum_{j\in[n]}A_{ij}=1$. Moreover, $A$ is positive semidefinite, and its second largest singular value denoted by $\sigma_2(A)$ is strictly smaller than $1$.
  % the second largest magnitude of its eigenvalues is strictly smaller than $1$, i.e., $\beta = \max(|\lambda_2(\A)|,|\lambda_n(\A)|)<1$, where $\lambda_i(\A)$ denotes the $i$-th largest eigenvalue of $\A$.
  % it satisfies that $1 = \lambda_1(\A) \ge \lambda_2(\A) \ge \cdots \ge \lambda_n(\A) \ge -1$. For convenience, we set  to be the second largest magnitude of the eigenvalues of $\bold A$ throughout the paper.
\end{assumption}
\textbf{Remark.} First, the specific feasible set $\mathcal{X}$ in Assumption \ref{assum:bounded-set} is introduced for a direct application of the boosting technique \citep{Zhang-Arxiv24}, though it is actually can be generalized to $\mathcal{X} = \Pi_{i=1}^{d}\mathcal{X}_i $ with $\mathcal{X}_i=[0,c_i]$ for any $c_i>0$ as in many previous studies \citep{chen2018online,zhang2022stochastic}. Second, as indicated in Assumption \ref{assum:DR-submodular}, we focus on a subclass of continuous submodular functions that exhibit diminishing returns. This property ensures that they are concave along any non-negative direction and any non-positive direction \citep{bian2017guaranteed}. Thus, it is essential for maximizing these functions. 
% In contrast, the non-negativity of functions is only used for simplifying the analysis. 
Third, following the convention of D-OCSM and OCSM, we will only use stochastic gradients to update the decision. If the adversary is fully adaptive, i.e., functions $f_t^1(\x),\dots,f_t^n(\x)$ can even depend on decisions $\x_t^1,\dots,\x_t^n$, the unbiasedness of stochastic gradients may not hold. To address this issue, one may consider a weakly adaptive adversary, whose functions $f_t^1(\x),\dots,f_t^n(\x)$ can depend on any decisions of previous rounds, but are independent of decisions $\x_t^1,\dots,\x_t^n$. However, for algorithms with the blocking update mechanism, the weakly adaptive assumption will be contradictory, because decisions $\x_t^1,\dots,\x_t^n$ may be the same as the previous decisions. Therefore, in Assumption \ref{assumption3}, we consider an oblivious adversary, i.e., all functions are chosen beforehand, which is independent of the randomness of algorithms. 
%, and thus are independent of the randomness of algorithms. 
%Finally, the assumptions on the matrix $A$ are following .
% In Assumption \ref{assumption3}, the oblivious adversary is introduced To ensure the unbiasedness, 
Besides, the following assumptions will be further utilized to improve the approximate regret of D-OCSM.
\begin{assumption}
\label{assum:smooth}
At each round $t\in[T]$, the reward function $f_t^i(\x)$ of each learner $i\in[n]$ is $\beta$-smooth over $\mathcal{X}$, i.e., $
  \Vert \nabla f_t^i(\x)-\nabla f_t^i(\y)\Vert\leq \beta\Vert\x-\y\Vert$ for any $\x,\y \in \mathcal{X}$.
\end{assumption}
\begin{assumption}
\label{assum:DCsets}
The decision set $\K\subseteq\mathcal{X}$ contains the origin $\mathbf{0}$, and is downward-closed, i.e., there exists a lower bound $\underline{\u}\in \K$ such that i) $\underline{\u}\leq \y$ for any $\y\in\K$; and ii) $\underline{\u}\leq\x\leq \y$ implies $\x\in\K$ for any $\y\in\K$ and $\x\in\mathbb{R}^d$.
\end{assumption}

% \begin{assumption}
% \label{assum:smooth}
% At each round $t\in[T]$, the reward function $f_t^i(\x)$ of each learner $i\in[n]$ is $\beta$-smooth over $\mathcal{X}$, i.e., $
%   \Vert \nabla f_t^i(\x)-\nabla f_t^i(\y)\Vert\leq \beta\Vert\x-\y\Vert$ for any $\x,\y \in \mathcal{X}$.
% \end{assumption}

\subsection{Reduction Based on the Boosting Technique}
Our first reduction is inspired the boosting technique of \citet{Zhang-Arxiv24}. Specifically, combining each function $f_t^i(\x)$ with this technique, it is easy to derive the following lemma, where $\xx = \argmin_{\x\in\K}\|\x\|_\infty$.
\begin{lemma}
\label{lem:non-oblvious-nonm}
(Corollary 16 of \citet{Zhang-Arxiv24}) Under Assumptions \ref{assum:bounded-set} and \ref{assum:DR-submodular}, for any $i\in[n]$ and $t\in[T]$, there exists a function $F_t^i(\x)$ defined by its gradient 
\begin{equation}
    \label{definition: non-monotone gradient F}
    \nabla F_t^i(\x)=\int_0^1\frac{1}{8(1-\frac{z}{2})^3}\nabla f_t^i\left(\frac{z}{2}(\x-\xx)+\xx\right)dz
\end{equation}
% \begin{equation*}
%      F_t^i(\x)= \int_0^1\frac{1}{4z\left(1-\frac{z}{2})\right)^3}\left(f_t^i\left(\frac{z}{2}\cdot(\x-\xx)+\xx\right)-f_t^i(\xx)\right)dz.
% \end{equation*}
such that for any $\x,\y \in \K$, it holds that
\begin{equation}
    \label{non-monotone linear reduction}
    \langle\nabla F_t^i(\x),\y-\x\rangle \ge \frac{1-\|\xx\|_\infty}{4}f_t^i(\y)-f_t^i\left(\frac{\x+\xx}{2}\right).
\end{equation}
\end{lemma}
Let $\Z$ denote the distribution over $[0,1]$ such that $\Pro(z\le c)=\int_0^c(1-\frac{u}{2})^{-3}/3~du$ for $z\sim \Z$ and any $c\leq 1$. 
% Let $Z$ denote a random variable in $[0,1]$, which satisfies  
Combining \eqref{definition: non-monotone gradient F} with Assumption \ref{assumption3}, an unbiased estimation of $\nabla F_t^i(\x)$ can be computed by sampling $z_t^i\sim\Z$ and setting $\tilde{\nabla} F_t^i(\x)=\frac{3}{8}\tilde{\nabla} f_t^i(\x^\prime)$, where $\x^\prime=(z_t^i/2)(\x-\xx)+\xx$. 

Then, from \eqref{non-monotone linear reduction}, in the special case with $n=1$, it is natural to generate a preparatory decision $\hat{\x}_t^1$ by running an OCO algorithm over linear losses $\{\ell_t^1(\x)\}_{t\in[T]}$, where $\ell_t^1(\x)=\langle-\tilde{\nabla} {F}_t^1(\hat{\x}_t^1), \x\rangle$, and play the decision $\x_t^1=(\hat{\x}_t^1+\xx)/2$. Similarly, for the general D-OCSM, each local learner $i$ should also maintain a preparatory decision $\hat{\x}_t^i$ and play the decision $\x_t^i=(\hat{\x}_t^i+\xx)/2$. However, it is more complicated to find an appropriate $\hat{\x}_t^i$, because a straightforward application of \eqref{non-monotone linear reduction} will result in the following upper bound
\begin{equation}
\label{simple-use}
  \alpha\sum_{t=1}^Tf_{t}(\x)-\sum_{t=1}^Tf_{t}(\x_t^i)\leq \sum_{t=1}^T\sum_{j=1}^n\langle\nabla F_t^j(\hat{\x}_t^i),\x-\hat{\x}_t^i\rangle 
\end{equation}
for any $\x\in\K$, where $\alpha=(1-\nu)/4$ and $\nu=\|\xx\|_\infty$. Due to the mismatch between $\nabla F_t^j(\cdot)$ and $\hat{\x}_t^i$, it is very difficult and even impossible to minimize this bound in the decentralized setting. Fortunately, we notice that this mismatch can be corrected, i.e., replacing $\nabla F_t^j(\hat{\x}_t^i)$ in \eqref{simple-use} with $\nabla F_t^j(\hat{\x}_t^j)$, by combining \eqref{non-monotone linear reduction} with the Lipschitz continuity of $f_t^i(\x)$, which is derived from Assumption \ref{assumption3}. This implies that we can generate the preparatory decision by running a D-OCO algorithm $\D$ over linear losses $\{\ell_t^i(\x)\}_{t\in[T],i\in[n]}$, where $\ell_t^i(\x)=\langle-\tilde{\nabla} {F}_t^i(\hat{\x}_t^i), \x\rangle$. 
\begin{algorithm}[t]
    \caption{Decentralized Online Submodular-to-Convex Reduction} 
    \label{DNSCR}
    \begin{algorithmic}[1]
        \STATE \textbf{Input:} a D-OCO algorithm $\D$ 
        %with the initial decision $\mathbf{0}$
        % \STATE Get an instance $\D_i$ of $\D$, get $\x_i(1)=0$ from $\D_i$, $\xx = \argmin_{\x\in\K}\|\x\|_\infty$
        \FOR{$t = 1,\dots,T$}
        \FOR{each local learner $i\in[n]$}
            \STATE Sample $z_t^i\sim\Z$, where $\Pro(z_t^i\le c)=\int_0^c(1-\frac{u}{2})^{-3}/3~du$ for any $c\leq 1$
            \STATE Get $\hat{\x}_t^i$ from $\D^i$, play $\x_t^i=(\hat{\x}_t^i+\xx)/{2}$, and query $\tilde{\nabla}f_{t}^i\left(\frac{z_t^i}{2}(\hat{\x}_t^i-\xx)+\xx\right)$
            \STATE Set $\tilde{\nabla} {F}_t^i(\hat{\x}_t^i)=\frac{3}{8}\tilde{\nabla} {f}_t^i\left(\frac{z_t^i}{2}(\hat{\x}_t^i-\xx)+\xx\right)$, and send $\ell_t^i(\x)=\langle-\tilde{\nabla} {F}_t^i(\hat{\x}_t^i), \x\rangle$ to $\D^i$ 
            % \STATE Define $F_{t,i}(\x)=\langle \g_i(t), \x \rangle$ and send it to $\D_i$ once $\g_i(t)$ is available, $\forall i\in [N]$
        \ENDFOR
        \ENDFOR
    \end{algorithmic}
\end{algorithm}

Based on these discussions, we are ready to reduce D-OCSM to D-OCO, and the detailed procedure is outlined in Algorithm \ref{DNSCR}, where $\D^i$ denotes the $i$-th local learner of $\D$. To demonstrate the power of this reduction, we first establish the following guarantee.
\begin{thm}
\label{thm: DSCR}
Let $\alpha=(1-\|\xx\|_\infty)/4$. Under Assumptions \ref{assum:bounded-set}, \ref{assum:DR-submodular}, and \ref{assumption3}, for any $i\in [n]$ and $\x\in\K$, Algorithm \ref{DNSCR} ensures  
\begin{equation}
\label{eq:regret of DNSCR}
\E\left[\alpha\sum_{t=1}^Tf_{t}(\x)-\sum_{t=1}^Tf_{t}(\x_t^i) \right] 
  \leq \E\left[\sum_{t=1}^T\sum_{j=1}^n\left(\ell_t^j(\hat{\x}_t^i)-\ell_t^j(\x)\right)\right]+\E\left[\sum_{t=1}^T\sum_{j=1}^nG\|\hat{\x}_t^j-\hat{\x}_t^i\|\right].
\end{equation}
\end{thm}
\textbf{Remark.} In the right side of \eqref{eq:regret of DNSCR}, the first term is the regret of $\D$ on $\{\ell_t^i(\x)\}_{t\in[T],i\in[n]}$, and the second term is the cumulative consensus error among $\D^1,\dots,\D^n$. Therefore, the above theorem provides a reduction from the $((1-v)/4)$-regret of D-OCSM to the regret of any D-OCO algorithm on linear losses, as long as its cumulative consensus error is sufficiently small. 

Actually, in many existing D-OCO algorithms, including the state-of-the-art ones for general convex functions: AD-FTRL and its projection-free variant \citep{wan2024nearly,wan2024optimal}, the cumulative consensus error is included as a part of the regret bound. Combining Theorem \ref{thm: DSCR} with them, we have the following two corollaries, where $C=\lceil\sqrt{2}\ln(\sqrt{14n})/((\sqrt{2}-1)\sqrt{\rho})\rceil$ is a constant and $\rho=1-\sigma_2(A)$ is the spectral gap. 

% Combining Theorem \ref{thm: DSCR} with them, it is easy to derive the following two corollaries.
%As shown in the following two corollaries, we can simply use Theorem \ref{thm: DSCR} to
%Therefore, it is easy to derive the following results for D-OCSM.
\begin{cor}
\label{cor1:label}
Let $\alpha=(1-\|\xx\|_\infty)/4$. Under Assumptions \ref{assum:bounded-set}, \ref{assum:DR-submodular}, \ref{assumption3}, and \ref{assumption4}, by setting $\D$ as AD-FTRL \citep{wan2024nearly,wan2024optimal}, for any $i\in [n]$, Algorithm \ref{DNSCR} ensures 
\[
  \E\left[\Reg(T,i,\alpha) \right]\leq 2GR\sqrt{5TC}.
\]

\end{cor}
\begin{cor}
\label{cor2:label}
Let $\alpha=(1-\|\xx\|_\infty)/4$. Under Assumptions \ref{assum:bounded-set}, \ref{assum:DR-submodular}, \ref{assumption3}, and \ref{assumption4}, by setting $\D$ as the projection-free variant of AD-FTRL \citep{wan2024optimal}, for any $i\in [n]$, Algorithm \ref{DNSCR} ensures
\[
  \E\left[\Reg(T,i,\alpha) \right]\leq 2nGR\sqrt{6TC }+(4\sqrt{3}+13)nGRT^{3/4}.
\]
\end{cor}
\textbf{Remark.} From Corollaries \ref{cor1:label} and \ref{cor2:label}, we derive a projection-based algorithm and a projection-free algorithm for D-OCSM with general functions and decision sets, which can enjoy ${O}(n\rho^{-1/4}\sqrt{T\log n})$ and $O(nT^{3/4}+n\rho^{-1/4}\sqrt{T\log n})$ bounds on the $((1-v)/4)$-regret, respectively. These two bounds are significantly better than the $O(n^{5/4}\rho^{-1/2}\sqrt{T})$  and $O(n^{5/4}\rho^{-1/2}T^{3/4})$ bounds achieved by existing projection-based and projection-free D-OCSM algorithms for general functions and decision sets \citep{D-online-upper-LO}. Such improvements mainly owe to the accelerated gossip strategy used in the two D-OCO algorithms. Our first reduction provides a very simple way to exploit this strategy.
\begin{algorithm}[t]
    \caption{Accelerated Decentralized Online Smooth Projection-free Algorithm} 
    \label{AD-OSPA}
    \begin{algorithmic}[1]
        \STATE \textbf{Input:} block size $L$, communication budget $K$, mixing ratio $\theta$, and perturbation parameter $\eta$ 
        % weight matrix $\mathbf{A} = [a_{ij}]\in \mathbb{R}_{+}^{N\times N}$, linear optimization oracle $\mathcal{O_K}(\cdot)$
        \STATE \textbf{Initialization:} set $\z_1^i=\z_{1,K-1}^{i}=\mathbf{0}$, and $\x_1^i=\x_2^i=\x\in\K, \forall i\in[n]$
        % For any $q\in [Q]$, initialize $\z_i(1)=\a_{i,k}^{0}(1)=\a_{i,k}^{M-1}(1)=0$.
        \FOR{$q=1,\dots,T/L$}
        % \STATE If $q\geq 2$, set $\z_{q,0}^i=\z_{q-1}^{i}+\g_{q-1}^i$ and $\z_{q,-1}^i=\z_{q-1,L-1}^{i}+\g_{q-1}^i,\forall i\in[n]$
            \FOR{$t=(q-1)L+1,\dots,qL$}
            \FOR{each local learner $i\in[n]$}
            \STATE Play $\x_q^i$, query $\nabla f_t^i(\x_q^i)$, and set $k=t-(q-1)L-1$
            \STATE If $q\geq 2$ and $k<K$, update $\z_{q,k+1}^i = (1+\theta)\sum_{j=1}^nA_{ij}\z_{q,k}^j-\theta\z_{q,k-1}^i$
            \ENDFOR
            \ENDFOR
            \FOR{each local learner $i\in[n]$}
            \STATE Set $\g_q^i=\sum_{t\in\mathcal{T}_q}\nabla f_t^i(\x_q^i)$, $\z_{q+1,0}^i=\z_{q}^{i}+\g_{q}^i$, and $\z_{q+1,-1}^i=\z_{q,K-1}^{i}+\g_{q}^i$
            \STATE If $q\ge 2$, set $\z_q^i=\z_{q,K}^i$, and sample $\v_{q,1}^i,\dots,\v_{q,L}^i\sim\mathcal{B}$ 
             \STATE If $q\ge 2$, compute
            %where $\mathcal{B}=\{\x\in\mathbb{R}^d|\|\x\|\leq 1\}$
            $\x_{q+1}^i=\frac{1}{L}\sum_{m=1}^L \argmax_{\x\in\K}\langle-\z_{q-1}^i+\eta \v_{q,m}^i,\x\rangle$
            % \mathcal{O}_{\K}(-\z_{q-1}^i+\eta \v_{q,m}^i)$
            \ENDFOR
            
                % \FOR{$t=(q-1)L+1,\cdots,qL$}
                %     \STATE Play $\x_{i}(t)=\x_{q,i}$
                %     \STATE $\g_i(t)=\nabla f_{t,i}(\x_i(t))$
                % \ENDFOR
                % \STATE $\d_i(q)=\sum_{t\in\mathcal{T}_q}\g_i(t)$
        \ENDFOR
    \end{algorithmic}
\end{algorithm}

Furthermore, we notice that the best existing projection-free algorithm for OCSM with smooth functions enjoys an $((1-v)/4)$-regret bound of $O(T^{2/3})$ \citep{pedramfar2024unified}. In the following, we will demonstrate that our first reduction is sufficient to bridge the gap between Corollary \ref{cor2:label} and this centralized result, though a more natural idea may be to extend this centralized algorithm. The key insight is that the linearization in our first reduction does not affect the exploitation of the smoothness in the projection-free D-OCO algorithm of \citet{wang2023distributed}, namely D-OSPF. Nonetheless, a direct application of D-OSPF is unsatisfactory, because it only uses the standard gossip step. To address this issue, we develop an improved variant of D-OSPF, which is outlined in Algorithm \ref{AD-OSPA} with four parameters $L$, $K$, $\theta$, and $\eta$. Moreover, in Algorithm \ref{AD-OSPA}, $\mathcal{B}=\{\x\in\mathbb{R}^d|\|\x\|\leq 1\}$ denotes the unit ball, and $\mathcal{T}_q=\{ (q-1)L+1,\dots,qL\}$ for any $q\in[T/L]$. Following D-OSPF, we adopt a blocking update mechanism, i.e., dividing $T$ rounds into $T/L$ blocks, and only maintaining a decision $\x_q^i$ for each local learner $i$ during all rounds in each block $q$. Our critical change is to maintain $\z_q^i\approx\sum_{\tau=1}^{q-1}\frac{1}{n}\sum_{i=1}^n\sum_{t\in\mathcal{T}_\tau}\nabla f_t^i(\x_\tau^i)$ via the accelerated gossip strategy, i.e., steps 7, 11, and 12 in Algorithm \ref{AD-OSPA}, which are simply following AD-FTRL and its projection-free variant \citep{wan2024nearly,wan2024optimal}. 

Combining the original analysis of D-OSPF with the property of the accelerated gossip strategy, we can establish an $O(nT^{2/3}+n\rho^{-1/4}\sqrt{T\log n})$ regret bound for projection-free D-OCO with oblivious and smooth losses (see Theorem \ref{thm1-AD-OSPF} in the appendix), as well as the linear losses constructed by our first reduction. However, we still cannot derive a desired result for D-OCSM, because the cumulative consensus error of Algorithm \ref{AD-OSPA} actually is bounded by $O(nT^{5/6}+n\rho^{-1/4}\sqrt{T\log n})$, where the $nT^{5/6}$ term is caused by the sampling error of $\x_{q+1}^i$. To this end, we show that besides Theorem \ref{thm: DSCR}, our first reduction enjoys the following guarantee for smooth functions.
\begin{thm}
\label{thm: DSCR-smooth}
Let $\alpha=(1-\|\xx\|_\infty)/4$. Under Assumptions \ref{assum:bounded-set}, \ref{assum:DR-submodular}, \ref{assumption3}, and \ref{assum:smooth}, for any $i\in [n]$ and $\x\in\K$, Algorithm \ref{DNSCR} ensures 
\begin{equation*}
\begin{split}
\E\left[\alpha\sum_{t=1}^Tf_{t}(\x)-\sum_{t=1}^Tf_{t}(\x_t^i) \right]\leq& \E\left[\sum_{t=1}^T\sum_{j=1}^n\left(\ell_t^j(\hat{\x}_t^i)-\ell_t^j(\x)\right)\right]+\E\left[\sum_{t=1}^T\sum_{j=1}^n\frac{\beta}{8}\|\hat{\x}_t^i-\hat{\x}_t^j\|^2\right]\\
&+\E\left[\sum_{t=1}^T\sum_{j=1}^n\left\langle\frac{1}{2}\nabla f_t^j(\x_t^j)-\nabla F_t^j(\hat{\x}_t^j),\hat{\x}_t^j-\hat{\x}_t^i\right\rangle\right].
  \end{split}
\end{equation*}
\end{thm}
\textbf{Remark.} Different from Theorem \ref{thm: DSCR}, the second term in the above bound is regarding the square consensus error. Moreover, the last term is newly added, but its effect is similar to the second term, due to the smoothness of $f_t^j(\x)$ and $F_t^j(\x)$ (see Lemma \ref{lem2:non-oblvious-nonm} in the appendix). Actually, the cumulative square consensus error of Algorithm \ref{AD-OSPA} is included as a part of its regret bound.

Therefore, combining Theorem \ref{thm: DSCR-smooth} with Algorithm \ref{AD-OSPA}, we have the following corollary. %, where $C=\sqrt{2}\ln(\sqrt{14n})/((\sqrt{2}-1)\sqrt{\rho})$.
\begin{cor}
\label{cor3:label}
Let $\alpha=(1-\|\xx\|_\infty)/4$ and $R^\prime=R(G+\beta R)$. Under Assumptions \ref{assum:bounded-set}, \ref{assum:DR-submodular}, \ref{assumption3}, \ref{assumption4}, and \ref{assum:smooth}, by setting $\D$ as Algorithm \ref{AD-OSPA}, for any $i\in [n]$, Algorithm \ref{DNSCR} ensures 
\[
  \E\left[\Reg(T,i,\alpha) \right]\leq 8n(\sqrt{2}T^{2/3}+\sqrt{TC})\sqrt{d}R^\prime+7\beta nT^{2/3}R^2+4n(2T^{1/3}+C)R^\prime.
\] 
\end{cor}
\textbf{Remark.} From Corollary \ref{cor3:label}, we derive a projection-free algorithm for D-OCSM with smooth functions and general decision sets, which improves the $O(nT^{3/4}+n\rho^{-1/4}\sqrt{T\log n})$ bound in Corollary \ref{cor2:label} to $O(nT^{2/3}+n\rho^{-1/4}\sqrt{T\log n})$. Moreover, it recovers the existing $((1-v)/4)$-regret bound of $O(T^{2/3})$ for projection-free OCSM with these functions and decision sets \citep{pedramfar2024unified}. 

%To the best of our knowledge, this is the first work to achieve such a result without the MFW framework.
%To the best of our knowledge, .

\subsection{Reduction Based on Decentralized Meta-Frank-Wolfe}
One limitation of our first reduction is that it cannot exploit the downward-closed property of the decision set to achieve the $(1/e)$-regret guarantee for smooth functions. To address this limitation, we propose to extend the computation-efficient MFW of \citet{pedramfar2024unified} into D-OCSM. 

Specifically, we also adopt the blocking update mechanism with a block size $L$. Let $\odot$ denote the element-wise product. At the beginning of each block $q$, we perform the following iterations
\begin{equation}
\label{iterations-MFW}
\x_{q,k+1}^i=\x_{q,k}^i+\frac{1}{L}\v_{q,k}^i\odot(\mathbf{1}-\x_{q,k}^i)~\text{for}~k=1,\dots, L
\end{equation}
with $\x_{q,1}^i=\ze$ and some $\v_{q,k}^i\in\K$, and set the decision as $\x_q^i=\x_{q,L+1}^i$. Let $\nabla_{t,k}^{i,j}=\nabla f_t^j(\x_{q,k}^i)\odot(\x_{q,k}^i-\onv)$ for any $t\in\mathcal{T}_q$. Due to the convergence of \eqref{iterations-MFW} (see Lemma \ref{lem1-DMFW} in the appendix), under Assumptions \ref{assum:bounded-set}, \ref{assum:DR-submodular}, \ref{assum:smooth}, and \ref{assum:DCsets}, for any $\x\in\K$, we can verify that
\begin{equation}
\label{pre-reduction}
\begin{split}
\sum_{q=1}^{T/L}\sum_{t\in\mathcal{T}_q}\left(\frac{1}{e} f_t(\x)-f_t(\x_q^i)\right)=O\left(\sum_{k=1}^L\sum_{q=1}^{T/L}\sum_{j=1}^n\left\langle\frac{1}{L}\sum_{t\in\mathcal{T}_q}\nabla_{t,k}^{i,j},\v_{q,k}^i-\x\right\rangle+\frac{nT}{L}\right).
\end{split}
\end{equation}
Moreover, let $(t_{q,1},\dots, t_{q,L})$ be a random permutation of $\mathcal{T}_q=\{ (q-1)L+1,\dots,qL\}$. Under Assumption \ref{assumption3}, we can verify that $\tilde{\nabla}f_{t_{q,k}}^j(\x_{q,k}^j)\odot(\x_{q,k}^j-\onv)$ is an unbiased estimation of $\frac{1}{L}\sum_{t\in\mathcal{T}_q}\nabla_{t,k}^{j,j}$. Thus, in the special case with $n=1$, for each $k\in[L]$, we can find $\{\v_{q,k}^1\}_{q\in[T/L]}$ by performing OCO over linear losses $\{\ell_{q,k}^1(\x)\}_{q\in[T/L]}$, where $\ell_{q,k}^1(\x)=\langle\tilde{\nabla}f_{t_{q,k}}^1(\x_{q,k}^1)\odot(\x_{q,k}^1-\onv), \x\rangle$. Indeed, this is how the computation-efficient MFW reduces OCSM to $L$ OCO problems. % \citep{pedramfar2024unified}.
\begin{algorithm}[t]
  \caption{Decentralized Meta-Frank-Wolfe}
    \label{ADMFW}
    \begin{algorithmic}[1]
        \STATE \textbf{Input:} a D-OCO algorithm $\D$, block size $L$
        \STATE Create $L$ instances of $\D$, and denote the $k$-th instance by $\D_k$
        % \STATE Set $\x_1^i=0$, $\x_1^i\uu =\mathbf{0}$.
        % \FOR{$i=1,\ldots,N$}
            \FOR{$q=1,\dots,T/L$}
            \STATE Get $\v_{q,k}^i$ from $\D_k^i,\forall k\in[L]$, compute $\{\x_{q,k}^i\}_{k\in[L+1]}$ as in \eqref{iterations-MFW}, and set $\x_{q}^i=\x_{q,L+1}^i,\forall i\in[n]$
            % \FOR{each local learner $i\in[n]$}
            % \STATE Set $\x_{q,1}^i=\ze$, and get $\v_{q,k}^i$ from $\D_k^i,\forall k\in[L]$
            % \STATE Update $\x_{q,k+1}^i = \x_{q,k}^i+ \frac{1}{L} \v_{q,k}^i\odot(\mathbf{1}-\x_{q,k}^i)$ for $k=1,\dots,L$, and set $\x_{q}^i=\x_{q,L+1}^i$
            % \ENDFOR
                \STATE Let $(t_{q,1},\dots, t_{q,L})$ be a random permutation of $\mathcal{T}_q=\{ (q-1)L+1,\dots,qL\}$
                \FOR{$t=(q-1)L+1,\dots,qL$}
                \FOR{each local learner $i\in[n]$}
                    \STATE Play $\x_{q}^i$, find the corresponding $k\in [L]$ such that $t=t_{q,k}$, and query $\tilde{\nabla}f_{t}^i(\x_{q,k}^i)$
                    \STATE Send $\ell_{q,k}^i(\x)=\langle\tilde{\nabla}f_{t}^i(\x_{q,k}^i)\odot(\x_{q,k}^i-\mathbf{1}),\x\rangle$ to $\D_k^i$
                \ENDFOR
                \ENDFOR
            \ENDFOR
        % \ENDFOR
    \end{algorithmic}
\end{algorithm}

However, in the general D-OCSM, the choice of update directions is more challenging due to the mismatch of $\nabla f_t^j(\cdot)$ and $\x_{q,k}^i$ in $\nabla_{t,k}^{i,j}$. To this end, we first utilize the smoothness of $f_t^j(\x)$ to correct this mismatch, i.e., replacing $\nabla_{t,k}^{i,j}$ in \eqref{pre-reduction} with $\nabla_{t,k}^{j,j}$, which also results in an additional term about $\|\x_{q,k}^i-\x_{q,k}^j\|$. Interestingly, we  notice that this term is bounded as \[O\left(\sum_{\tau=1}^{k-1}\frac{\|\v_{q,\tau}^j-\v_{q,\tau}^i\|}{L}\right)\] even though iterations in \eqref{iterations-MFW} are performed locally. 
% \[\|\x_{q,k}^i-\x_{q,k}^j\|=O\left(\frac{1}{L}\sum_{\tau=1}^{k-1}\|\v_{q,\tau}^j-\v_{q,\tau}^i\|\right)\]
Then, for each $k\in[L]$, we only need to find $\{\v_{q,k}^i\}_{q\in[T/L],i\in[n]}$ by performing D-OCO over linear losses $\{\ell_{q,k}^i(\x)\}_{q\in[T/L],i\in[n]}$, where $\ell_{q,k}^i(\x)=\langle\tilde{\nabla}f_{t_{q,k}}^i(\x_{q,k}^i)\odot(\x_{q,k}^i-\onv), \x\rangle$. In this way, D-OCSM is converted to $L$ D-OCO problems. The detailed procedure is outlined in Algorithm \ref{ADMFW}, and its theoretical guarantee is presented below.

% \textbf{Remark.} By substituting $\v_k=\argmax_{\x\in\K}\langle\nabla f(\x_{k})\odot(\onv-\x_{k}),\x\rangle$ into \eqref{main-lem1-DMFW}, it can be simplified to $(1/e)f(\x)-f(\x_{L+1})=O(1/L)$, i.e., an $(1/e)$-approximation convergence rate of $O(1/L)$ for offline optimization. In this case, the iterations in Lemma \ref{lem1-DMFW} are known as the classical Frank-Wolfe (FW) method \citep{FW-56}. 

%More interestingly, from \eqref{main-lem1-DMFW}, we actually .
\begin{thm}
\label{thm1-DMFW}
Under Assumptions \ref{assum:bounded-set}, \ref{assum:DR-submodular}, \ref{assumption3}, \ref{assumption4}, \ref{assum:smooth}, and \ref{assum:DCsets}, for any $i\in [n]$ and $\x\in\K$, Algorithm \ref{ADMFW} ensures 
\begin{equation*}
\begin{split}
\E\left[\sum_{q=1}^{T/L}\sum_{t\in\mathcal{T}_q}\left(\frac{1}{e} f_t(\x)-f_t(\x_q^i)\right)\right]
\leq& \E\left[\sum_{k=1}^L\left(1-\frac{1}{L}\right)^{L-k}\sum_{q=1}^{T/L}\sum_{j=1}^n\left(\ell_{q,k}^j(\v_{q,k}^i)-\ell_{q,k}^j(\x)\right)\right]\\
&+\E\left[4(\beta+G)R\sum_{k=1}^L\sum_{q=1}^{T/L}\sum_{j=1}^n\|\v_{q,k}^j-\v_{q,k}^i\|\right]+\frac{n\beta TR^2}{2L}.
\end{split}
\end{equation*}
\end{thm}
\textbf{Remark.} For any $k\in[L]$, linear losses $\{\ell_{q,k}^i(\x)\}_{q\in[T/L],i\in[n]}$ in Algorithm \ref{ADMFW} actually are oblivious to the $k$-th instance of the D-OCO algorithm $\D$, i.e., $\D_k$. This is because such losses will not change even if only the previous $k-1$ instances of $\D$ are maintained. Therefore, from Theorem \ref{thm1-DMFW}, if $\D$ has an $\Reg(T)$ regret bound over $T$ oblivious and linear losses, and a sufficiently small cumulative consensus error, Algorithm \ref{ADMFW} can enjoy an $(1/e)$-regret bound of $O(nT/L+L\Reg(T/L))$ for D-OCSM with smooth functions and downward-closed decision sets.
\begin{algorithm}[t]
    \caption{Decentralized Follow-The-Perturbed-Leader} 
    \label{D-FTPL}
    \begin{algorithmic}[1]
        \STATE \textbf{Input:} block size $L$, perturbation parameter $\eta$ 
        % weight matrix $\mathbf{A} = [a_{ij}]\in \mathbb{R}_{+}^{N\times N}$, linear optimization oracle $\mathcal{O_K}(\cdot)$
        % \STATE \textbf{Initialization:} set $\z_1^i=\mathbf{0},\forall i\in[n]$
        % For any $q\in [Q]$, initialize $\z_i(1)=\a_{i,k}^{0}(1)=\a_{i,k}^{M-1}(1)=0$.
        \STATE \textbf{Initialization:} set $\x_1^1=\dots=\x_1^n=\hat{\x}^1_1=\dots=\hat{\x}_1^n\in\K$
        \FOR{$q=1,\dots,T/L$}
        % \FOR{$i=1,\dots,n$}
        \STATE Compute $\x_{q+1,1}^i=\hat{\x}_{q+1}^i=\argmax_{\x\in\K}\langle-\sum_{\tau=1}^{q-2}\g_{\tau}^i+\eta\v_{q}^i,\x\rangle$, where $\v_{q}^i\sim\mathcal{B}$, $\forall i\in[n]$
        % \ENDFOR
        \FOR{$t=(q-1)L+1,\dots,qL$}
            \FOR{each local learner $i\in[n]$}
            \STATE Play $\x_q^i$, query $\nabla f_t^i(\x_q^i)$, and set $k_x=t-(q-1)L,k_g=t-(q-1)L-L/2$
            %, and update $\z_{t+1}^i = \sum_{j=1}^nA_{ij}\z_{t}^j+\nabla f_t^i(\x_t^i)$
            \STATE If $k_x\leq L/2$, update $\x^i_{q+1,k_x+1}=\sum_{j=1}^nA_{ij}\x^j_{q+1,k_x}$
            \STATE If $k_g\geq 1$ and $q\geq 2$, update $\g^i_{q-1,k_g+1}=\sum_{j=1}^nA_{ij}\g^j_{q-1,k_g}$
            \ENDFOR
        \ENDFOR
        % \FOR{$i=1,\dots,n$}
        \STATE Set $\g_{q-1}^i=\g^i_{q-1,L/2+1}$ if $q\geq 2$, $\x_{q+1}^i=\x_{q+1,L/2+1}^i$, and $\g_{q,1}^i=\sum_{t\in\mathcal{T}_q}\nabla f_t^i(\x_q^i),\forall i\in[n]$
        % \ENDFOR
        \ENDFOR
    \end{algorithmic}
\end{algorithm}

Due to the oblivious property of linear losses, it is appealing to combine  Algorithm \ref{ADMFW} with FTPL-type D-OCO algorithms, to derive projection-free D-OCSM algorithms almost without sacrifice on the approximate regret. Actually, one can verify that Algorithm \ref{AD-OSPA} with $L=C$ reduces to a FTPL-type D-OCO algorithm with ${O}(n\rho^{-1/4}\sqrt{T\log n})$ regret for oblivious and linear losses (see Theorem \ref{thm2-AD-OSPF} in the appendix). However, both Algorithm \ref{AD-OSPA} with $L=C$ and the existing FTPL-type D-OCO algorithm, i.e., D-OSPF \citep{wang2023distributed} with $L=1$, cannot be used in Algorithm \ref{ADMFW}, because their sampling errors would make their consensus errors unbounded. To tackle this challenge, as outlined in Algorithm \ref{D-FTPL}, we further develop a new decentralized variant of FTPL for oblivious and linear losses. It is inspired by a black-box reduction from D-OCO to OCO \citep{COLT25-Wan}. The key idea is to locally run FTPL with the delayed approximation of average gradients to generate a preparatory decision $\hat{\x}_q^i$, i.e., step 4, and generate the decision $\x_q^i$ by applying multiple standard gossip steps to preparatory decisions, i.e., step 8. In this way, it enjoys an ${O}(n\rho^{-1/2}\sqrt{T\log (nT)})$ regret bound and an $O(n)$ consensus error bound (see Theorem \ref{thm-DFTPL} and Lemma \ref{lem1:D-FTPL} in the appendix). Finally, combining Theorem \ref{thm1-DMFW} with it, we have the following corollary.
\begin{cor}
\label{cor5}
Let $\alpha=1/e$ and $C^\prime=2\lceil \ln(\sqrt{n}T/L)\rho^{-1}\rceil$. Under Assumptions \ref{assum:bounded-set}, \ref{assum:DR-submodular}, \ref{assumption3}, \ref{assumption4}, \ref{assum:smooth}, and \ref{assum:DCsets}, by setting $\D$ as Algorithm \ref{D-FTPL}, for any $i\in [n]$, Algorithm \ref{ADMFW} with $L$ such that both $T/L$ and $T/(LC^\prime)$ are integers ensures
\[
\E\left[\Reg(T,i,\alpha) \right]
\leq 15nRG\sqrt{dTC^\prime L}+16(\beta+G)nR^2L+\frac{n\beta TR^2}{2L}.
\]
This bound becomes $\E\left[\Reg(T,i,\alpha) \right]={O}(n(\log (Tn))^{1/3}\rho^{-1/3}T^{2/3})$ with $L=\Theta((T/C^\prime)^{1/3})$.
\end{cor}
\textbf{Remark.} From Corollary \ref{cor5}, we derive a projection-free D-OCSM algorithm that enjoys an $(1/e)$-regret bound of ${O}(n(\log (Tn))^{1/3}\rho^{-1/3}T^{2/3})$ for smooth functions and downward-closed decision sets. This is the first $(1/e)$-regret guarantee for D-OCSM, and can match the $(1/e)$-regret bound of $O(T^{2/3})$ achieved by the existing projection-free OCSM algorithm \citep{pedramfar2024unified}.

\section{Conclusion}
This paper proposes two reductions from D-OCSM to D-OCO. Based on these reductions, we have proposed a series of improved D-OCSM algorithms. Specifically, our two projection-free algorithms for smooth functions are the first to recover the best results achieved in the centralized setting. Moreover, all our algorithms for general decision sets are the first to achieve approximate regret bounds matching the current best regret bounds of D-OCO. 

% There still are several directions for future research, which are discussed in the appendix due to
% the limitation of space.

%This paper proposes two reductions that allow us to minimize the approximate regret of D-OCSM by simply exploiting algorithms for D-OCO---a more extensively studied problem.

\bibliography{ref}

\newpage
\appendix
\section{Theoretical Guarantees of Our Projection-free D-OCO Algorithms}
In this section, we present the theoretical guarantees of our two projection-free D-OCO algorithms, i.e., Algorithms \ref{AD-OSPA} and \ref{D-FTPL}, which can be combined with our two reductions to derive the corresponding results for D-OCSM.

Recall that $C=\lceil\sqrt{2}\ln(\sqrt{14n})/((\sqrt{2}-1)\sqrt{\rho})\rceil$ and $\mathcal{T}_q=\{ (q-1)L+1,\dots,qL\}$. Moreover, with some abuse of notations, we define $G=\max_{\x\in\K,i\in[n],t\in[T]}\|\nabla f_t^i(\x)\|$ and $R=\max_{\x\in\K}\|\x\|$ in this section. For Algorithm \ref{AD-OSPA}, we first introduce the following two lemmas:~one about the regret on linearized losses and the other about some consensus properties. 
\begin{lemma}
\label{lem1-AD-OSPF:label}
Let $\bar{\y}_q=\mathbb{E}_{\v\in\B}\left[\argmax_{\x\in\K}\left\langle-\bar{\z}_{q}+\eta \v,\x\right\rangle\right]$ for any $q\in[T/L]$, where $\bar{\z}_q=\sum_{\tau=1}^{q-1}\bar{\g}_\tau$ and $\bar{\g}_q=\frac{1}{n}\sum_{i=1}^n\g_q^i$. Suppose the decision set $\K$ is convex. Then, Algorithm \ref{AD-OSPA} ensures
\[
\sum_{q=1}^{T/L}\sum_{t\in\mathcal{T}_q}\sum_{j=1}^n\langle \nabla f_{t}^j(\x_q^j),\x_q^i-\x\rangle\leq  2\eta nR+\frac{ dnRTLG^2}{2\eta}+\sum_{q=1}^{T/L}\sum_{t\in\mathcal{T}_q}\sum_{j=1}^n\langle \nabla f_{t}^j(\x_q^j),\x_q^i-\bar{\y}_q\rangle.
\]
\end{lemma}
\begin{lemma}
\label{lem2-AD-OSPF:label}
Suppose the decision set $\K$ is convex, and Assumption \ref{assumption4} holds.  For any $q=3,\dots,T/L$ and $i,j\in[n]$, Algorithm \ref{AD-OSPA} with $\theta=1/(1+\sqrt{1-\sigma_2^2(A)})$ and $K=C\leq L$ ensures \[\E\left[\|{\x}_q^i-{\x}_q^j\|^2\right]\leq \frac{32R^2}{L}+\frac{48 dLGR^2}{\eta}.\]
Moreover, let $f(\x):\K\mapsto\mathbb{R}$ be an $\beta$-smooth function over $\K$ and $G^\prime=\max_{\x\in\K}\|\nabla f(\x)\|$. If $f(\x)$ is oblivious to Algorithm \ref{AD-OSPA}, for any $q=3,\dots,T/L$ and $i,j\in[n]$, Algorithm \ref{AD-OSPA} with $\theta=1/(1+\sqrt{1-\sigma_2^2(A)})$ and $K=C\leq L$  ensures
\begin{equation*}
\begin{split}
\E\left[\left|\langle \nabla f(\tilde{\x}_q^j),\x_q^i-\bar{\y}_q\rangle\right|\right]
\leq\frac{4\beta \lambda R^2}{L}+\frac{5dLRGG^\prime}{\eta}
% ,~\E\left[\left|\langle\nabla f({\x}_q^j),{\x}_q^j-{\x}_q^i\rangle\right|\right]
% \leq\frac{8\beta R^2}{L}+6\eta dLRGG^\prime.
\end{split}
\end{equation*}
where $\tilde{\x}_q^j=\lambda \x_q^j+(1-\lambda)\xx$ for any $\lambda\in[0,1]$, and $\bar{\y}_q$ follows the definition in Lemma \ref{lem1-AD-OSPF:label}.
\end{lemma}
\textbf{Remark.} Note that that Lemmas \ref{lem1-AD-OSPF:label} and \ref{lem2-AD-OSPF:label} do not require the convexity of functions $\{f_t^i(\x)\}_{i\in[n],t\in[T]}$ and $f(\x)$. Therefore, they can be used to further analyze the upper bound provided in Theorem \ref{thm: DSCR-smooth}, and finally derive the $((1-v)/4)$-regret bound of $O(nT^{2/3}+n\rho^{-1/4}\sqrt{T\log n})$ for D-OCSM with smooth functions, as presented in Corollary \ref{cor3:label}.  

If these functions are convex, i.e., D-OCO, we can combine these two lemmas to derive the following theorem.
\begin{thm}
\label{thm1-AD-OSPF} Suppose the decision set $\K$ is convex, $f_t^i(\x)$ is convex and $\beta$-smooth over $\K$ for any $i\in[n],t\in[T]$, the adversary is oblivious, and Assumption \ref{assumption4} holds. Let $R^\prime=R(G+\beta R)$. For any $i\in[n]$, Algorithm \ref{AD-OSPA} with $\theta=1/(1+\sqrt{1-\sigma_2^2(A)})$, $K=C$, $L=\max\{\lceil T^{1/3}\rceil,C\}$, and $\eta=G\sqrt{dTL}$ ensures 
\begin{equation*}
\begin{split}
\E\left[\Reg(T,i)\right]\leq 24n(\sqrt{2}T^{2/3}+\sqrt{TC})\sqrt{d}R^\prime+20\beta nT^{2/3}R^2+4n(2T^{1/3}+C)R^\prime.
  \end{split}
\end{equation*}
\end{thm}
\textbf{Remark.} From Theorem \ref{thm1-AD-OSPF}, Algorithm \ref{AD-OSPA} improves the regret of projection-free D-OCO with oblivious and smooth losses from ${O}(n^{5/4}\rho^{-1/2}T^{2/3})$ \citep{wang2023distributed} to $O(nT^{2/3}+n\rho^{-1/4}\sqrt{T\log n})$.

Furthermore, if the functions $\{f_t^i(\x)\}_{i\in[n],t\in[T]}$ are oblivious and linear, i.e., $f_t^i(\x)=\langle\c_t^i,\x\rangle$, where $\c_t^i$ is chosen beforehand, the above theorem can be improved as below.
\begin{thm}
\label{thm2-AD-OSPF} Suppose the decision set $\K$ is convex, $f_t^i(\x)=\langle\c_t^i,\x\rangle$ for any $i\in[n],t\in[T]$, the adversary is oblivious, and Assumption \ref{assumption4} holds. For any $i\in[n]$, Algorithm \ref{AD-OSPA} with $\theta=1/(1+\sqrt{1-\sigma_2^2(A)})$, $L=K=C$, and $\eta=G\sqrt{dTL}$ ensures 
\begin{equation*}
\begin{split}
\E\left[\Reg(T,i)\right]\leq 8n\sqrt{dTC}RG+4nCRG.
  \end{split}
\end{equation*}
\end{thm}
\textbf{Remark.} From Theorem \ref{thm2-AD-OSPF}, Algorithm \ref{AD-OSPA} enjoys an $O(n\rho^{-1/4}\sqrt{T\log n})$ regret bound for projection-free D-OCO with oblivious and linear losses. However, because of $L=C$, the sampling error of $\x_{q+1}^i$ is a constant, which could make the cumulative consensus error of Algorithm \ref{AD-OSPA} unbounded. Therefore, as previously mentioned, Algorithm \ref{AD-OSPA} cannot be combined with Theorem \ref{thm1-DMFW}.

To address this issue, Algorithm \ref{D-FTPL} is proposed to achieve guarantees on both the consensus property and regret.
%Due to the property of the standard gossip step \citep{Xiao-Gossip04}, we can establish the following lemma.
\begin{lemma}
\label{lem1:D-FTPL}
Let $\bar{\g}_q=(1/n)\sum_{i=1}^n\sum_{t\in\mathcal{T}_q}\nabla f_t^i(\x_q^i)$ and $\bar{\x}_q=(1/n)\sum_{i=1}^n\hat{\x}_q^i$ for any $q\in[T/L]$. For any $q\in[T/L]$ and $i\in[n]$, Algorithm \ref{D-FTPL} with $L=2\lceil \ln(\sqrt{n}T)\rho^{-1}\rceil$ ensures
\[
\|\g_{q}^i-\bar{\g}_q\|\leq \frac{2LG}{T}~\text{and}~\|\x_{q}^i-\bar{\x}_q\|\leq \frac{2R}{T}.
\]
\end{lemma}
\begin{thm}
\label{thm-DFTPL}
Suppose the decision set $\K$ is convex, $f_t^i(\x)=\langle\c_t^i,\x\rangle$ for any $i\in[n],t\in[T]$, the adversary is oblivious, and Assumption \ref{assumption4} holds. For any $i\in[n]$, Algorithm \ref{D-FTPL} with $L=2\lceil \ln(\sqrt{n}T)\rho^{-1}\rceil$ and $\eta=G\sqrt{dTL}$ ensures 
\begin{equation*}
\begin{split}
\E\left[\Reg(T,i)\right]\leq 5nRG\sqrt{dTL}+4nL GR+6nGR.
  \end{split}
\end{equation*}
\end{thm}
\textbf{Remark.} From Theorem \ref{thm-DFTPL}, Algorithm \ref{D-FTPL} achieves an $O(n\rho^{-1/2}\sqrt{T\log (nT)})$ regret bound for projection-free D-OCO with oblivious and linear losses. Moreover, according to Lemma \ref{lem1:D-FTPL}, it is easy to derive an $O(n)$ cumulative consensus error bound. Finally, one may notice that there exists a gap of $O(\rho^{-1/4}\sqrt{\log T})$ between this regret bound and that in Theorem \ref{thm2-AD-OSPF}. This gap is caused by the simple application of multiple standard gossip steps in Algorithm \ref{D-FTPL}. Following \citet{COLT25-Wan}, it is possible to reduce this gap to only $O(\sqrt{\log T})$ by using the accelerated gossip strategy. However, if we apply this strategy to $\hat{\x}_q^i$, the output may be not contained in $\K$. To this end, a projection operation must be applied, which destroys the projection-free property of Algorithm \ref{D-FTPL}. Actually, if not pursuing the projection-free property, AD-FTRL \citep{wan2024nearly,wan2024optimal} is sufficient to be combined with Theorem \ref{thm1-DMFW}.

\section{A Monotone Variant of Our First Reduction}
\label{sec:a_monotone_variant_of_our_first_reduction}
In this section, we present a variant of our first reduction, which can  achieve the $(1-1/e)$-regret guarantees when the following monotone assumption on functions holds.
\begin{assumption}
\label{assum:mono}
At each round $t\in[T]$, the reward function $f_t^i(\x)$ of each learner $i\in[n]$ is monotone over $\mathcal{X}$, i.e., $
  f_t^i(\x)\leq f_t^i(\y)$ for any $\x\leq\y \in \mathcal{X}$, and satisfies $f_t^i(\ze)=\ze$.
\end{assumption}
Specifically, we notice that there exist simpler yet better auxiliary functions for monotone functions.
\begin{lemma}
\label{lem:non-oblvious-mon}
(Corollary 7 of \citet{Zhang-Arxiv24}) Under Assumptions \ref{assum:bounded-set}, \ref{assum:DR-submodular}, and \ref{assum:mono}, for any $i\in[n]$ and $t\in[T]$, there exists a function $F_t^i(\x)$ defined by its gradient $\nabla F_t^i(\x)=\int_0^1e^{z-1}\nabla f_t^i\left(z\x\right)dz$ 
% \begin{equation*}
%      F_t^i(\x)= \int_0^1\frac{1}{4z\left(1-\frac{z}{2})\right)^3}\left(f_t^i\left(\frac{z}{2}\cdot(\x-\xx)+\xx\right)-f_t^i(\xx)\right)dz.
% \end{equation*}
such that for any $\x,\y \in \K$, it holds that $\langle\nabla F_t^i(\x),\y-\x\rangle \ge (1-1/e)f_t^i(\y)-f_t^i\left(\x\right)$.
\end{lemma}
Let $\Z^\prime$ denote the distribution over $[0,1]$ such that $\Pro(z\le c)=\int_0^ce^{u-1}/(1-1/e)~du$ for $z\sim \Z^\prime$ and any $c\leq 1$. 
% Let $Z$ denote a random variable in $[0,1]$, which satisfies  
From Assumption \ref{assumption3}, an unbiased estimation of $\nabla F_t^i(\x)$ in Lemma \ref{lem:non-oblvious-mon} can be computed by sampling $z_t^i\sim\Z^\prime$ and setting $\tilde{\nabla} F_t^i(\x)=(1-1/e)\tilde{\nabla} f_t^i(z_t^i\x)$. Therefore, Algorithm \ref{DNSCR} can be naturally modified to Algorithm \ref{mono-DNSCR}. For the case without the smoothness assumption, we now have the following guarantee.
\begin{algorithm}[t]
    \caption{Decentralized Online Submodular-to-Convex Reduction for Monotone Functions} 
    \label{mono-DNSCR}
    \begin{algorithmic}[1]
        \STATE \textbf{Input:} a D-OCO algorithm $\D$ 
        %with the initial decision $\mathbf{0}$
        % \STATE Get an instance $\D_i$ of $\D$, get $\x_i(1)=0$ from $\D_i$, $\xx = \argmin_{\x\in\K}\|\x\|_\infty$
        \FOR{$t = 1,\dots,T$}
        \FOR{each local learner $i\in[n]$}
            \STATE Sample $z_t^i\sim\Z^\prime$, where $\Pro(z_t^i\le c)=\int_0^ce^{u-1}/(1-1/e)~du$ for any $c\leq 1$
            \STATE Get $\hat{\x}_t^i$ from $\D^i$, play $\x_t^i=\hat{\x}_t^i$, and query $\tilde{\nabla}f_{t}^i(z_t^i\hat{\x}_t^i)$
            \STATE Set $\tilde{\nabla} {F}_t^i(\hat{\x}_t^i)=(1-1/e)\tilde{\nabla}f_{t}^i(z_t^i\hat{\x}_t^i)$, and send $\ell_t^i(\x)=\langle-\tilde{\nabla} {F}_t^i(\hat{\x}_t^i), \x\rangle$ to $\D^i$ 
            % \STATE Define $F_{t,i}(\x)=\langle \g_i(t), \x \rangle$ and send it to $\D_i$ once $\g_i(t)$ is available, $\forall i\in [N]$
        \ENDFOR
        \ENDFOR
    \end{algorithmic}
\end{algorithm}
\begin{thm}
\label{thm: mono-DSCR}
Let $\alpha=(1-1/e)$. Under Assumptions \ref{assum:bounded-set}, \ref{assum:DR-submodular}, \ref{assumption3}, and \ref{assum:mono}, for any $i\in [n]$ and $\x\in\K$, Algorithm \ref{mono-DNSCR} ensures  
\begin{equation*}
\E\left[\alpha\sum_{t=1}^Tf_{t}(\x)-\sum_{t=1}^Tf_{t}(\x_t^i) \right] 
  \leq \E\left[\sum_{t=1}^T\sum_{j=1}^n\left(\ell_t^j(\hat{\x}_t^i)-\ell_t^j(\x)\right)\right]+\E\left[\sum_{t=1}^T\sum_{j=1}^n2G\|\hat{\x}_t^j-\hat{\x}_t^i\|\right].
\end{equation*}
\end{thm}
\textbf{Remark.} Different from Theorem \ref{thm: DSCR}, the above upper bound holds for the $(1-1/e)$-regret, instead of the $((1-v)/4)$-regret. Besides, the only difference is the constant in the last term. Therefore, Theorem \ref{thm: mono-DSCR} provides a reduction from the $(1-1/e)$-regret of D-OCSM to the regret of any D-OCO algorithm on linear losses, as long as its cumulative consensus error is sufficiently small. 

Combining Theorem \ref{thm: mono-DSCR} with AD-FTRL and its projection-free variant \citep{wan2024nearly,wan2024optimal}, we have the following corollaries, where $C=\lceil\sqrt{2}\ln(\sqrt{14n})/((\sqrt{2}-1)\sqrt{\rho})\rceil$.
\begin{cor}
\label{mono-cor1:label}
Let $\alpha=(1-1/e)$. Under Assumptions \ref{assum:bounded-set}, \ref{assum:DR-submodular}, \ref{assumption3}, \ref{assumption4}, and \ref{assum:mono}, by setting $\D$ as AD-FTRL \citep{wan2024nearly,wan2024optimal}, for any $i\in [n]$, Algorithm \ref{mono-DNSCR} ensures 
\[
  \E\left[\Reg(T,i,\alpha) \right]\leq 10GR\sqrt{TC}.
\]

\end{cor}
\begin{cor}
\label{mono-cor2:label}
Let $\alpha=(1-1/e)$. Under Assumptions \ref{assum:bounded-set}, \ref{assum:DR-submodular}, \ref{assumption3}, \ref{assumption4},  and \ref{assum:mono}, by setting $\D$ as the projection-free variant of AD-FTRL \citep{wan2024optimal}, for any $i\in [n]$, Algorithm \ref{mono-DNSCR} ensures
\[
  \E\left[\Reg(T,i,\alpha) \right]\leq 8nGR\sqrt{2TC}+44nGRT^{3/4}.
\]
\end{cor}
\textbf{Remark.} These two corollaries are similar to Corollaries \ref{cor1:label} and \ref{cor2:label}, except the difference of $\alpha$ and the constants in the upper bounds. From them, we can derive a projection-based algorithm and a projection-free algorithm for D-OCSM with monotone functions, which enjoy ${O}(n\rho^{-1/4}\sqrt{T\log n})$ and $O(nT^{3/4}+n\rho^{-1/4}\sqrt{T\log n})$ bounds on the $(1-1/e)$-regret, respectively. These two bounds are significantly better than the $O(n^{5/4}\rho^{-1/2}\sqrt{T})$  and $O(n^{5/4}\rho^{-1/2}T^{3/4})$ bounds achieved by existing projection-based and projection-free D-OCSM algorithms for monotone functions \citep{zhang2022communication,Liao-Arxiv}. 

Furthermore, if functions are also smooth, we can derive the following theorem and corollary.
\begin{thm}
\label{thm: mono-DSCR-smooth}
Let $\alpha=(1-1/e)$. Under Assumptions \ref{assum:bounded-set}, \ref{assum:DR-submodular}, \ref{assumption3}, \ref{assum:smooth},  and \ref{assum:mono}, for any $i\in [n]$ and $\x\in\K$, Algorithm \ref{mono-DNSCR} ensures 
\begin{equation*}
\begin{split}
\E\left[\alpha\sum_{t=1}^Tf_{t}(\x)-\sum_{t=1}^Tf_{t}(\x_t^i) \right]\leq& \E\left[\sum_{t=1}^T\sum_{j=1}^n\left(\ell_t^j(\hat{\x}_t^i)-\ell_t^j(\x)\right)\right]+\E\left[\sum_{t=1}^T\sum_{j=1}^n\frac{\beta}{2}\|\hat{\x}_t^i-\hat{\x}_t^j\|^2\right]\\
&+\E\left[\sum_{t=1}^T\sum_{j=1}^n\left\langle\nabla f_t^j(\hat{\x}_t^j)-\nabla F_t^j(\hat{\x}_t^j),\hat{\x}_t^j-\hat{\x}_t^i\right\rangle\right].
  \end{split}
\end{equation*}
\end{thm}
\begin{cor}
\label{mono-cor3:label}
Let $\alpha=(1-1/e)$ and $R^\prime=R(G+\beta R)$. Under Assumptions \ref{assum:bounded-set}, \ref{assum:DR-submodular}, \ref{assumption3}, \ref{assumption4}, \ref{assum:smooth},  and \ref{assum:mono}, by setting $\D$ as Algorithm \ref{AD-OSPA}, for any $i\in [n]$, Algorithm \ref{mono-DNSCR} ensures 
\[
  \E\left[\Reg(T,i,\alpha) \right]\leq 24n(\sqrt{2}T^{2/3}+\sqrt{TC})\sqrt{d}R^\prime+26\beta nT^{2/3}R^2+8n(2T^{1/3}+C)R^\prime.
\] 
\end{cor}
\textbf{Remark.} Theorem \ref{thm: mono-DSCR-smooth} is similar to Theorem \ref{thm: DSCR-smooth}, except the difference of $\alpha$ and some constants in the upper bound. Therefore, it can also be combined with Algorithm \ref{AD-OSPA} to establish Corollary \ref{mono-cor3:label}. More specifically, from Corollary \ref{mono-cor3:label}, we can derive a projection-free algorithm for D-OCSM with smooth and monotone functions, which improves the $O(nT^{3/4}+n\rho^{-1/4}\sqrt{T\log n})$ bound in Corollary \ref{mono-cor2:label} to $O(nT^{2/3}+n\rho^{-1/4}\sqrt{T\log n})$. It can recover the existing $(1-1/e)$-regret bound of $O(T^{2/3})$ for projection-free OCSM with these functions \citep{pedramfar2024unified}. 

\section{Analysis of Algorithm \ref{DNSCR}}
In this section, we present detailed proofs of the theoretical guarantees for Algorithm \ref{DNSCR}.
% we provide the detailed proofs of theoretical guarantees of Algorithm  

\subsection{Proof of Theorem \ref{thm: DSCR}}
For any $\x\in\K$, due to Assumption \ref{assumption3} and Jensen's inequality, it is easy to verify that
\begin{equation}
\label{bound-gradient}
\|\nabla f_t^i(\x)\|=\left\| \mathbb{E}[\tilde{\nabla} f_t^i(\x)|\x]\right\|\leq \mathbb{E}\left[\left.\left\|\tilde{\nabla} f_t^i(\x)\right\|\right|\x\right]\leq G.
\end{equation}
Combining \eqref{bound-gradient} with the convexity of $\K$, we can verify that $f_t^i(\x)$ is $G$-Lipschitz over $\K$, i.e., 
\begin{equation}
\label{Lipschitz-property}
|f_t^i(\x)-f_t^i(\y)|\leq G\|\x-\y\|,\forall \x,\y\in\K.
\end{equation}
Due to $f_t(\x)=\sum_{j=1}^nf_t^j(\x)$ and \eqref{Lipschitz-property}, for any $\x\in\K$, we have
\begin{equation}
\label{thm1-eq1}
\begin{split}
  \alpha\sum_{t=1}^Tf_{t}(\x)-\sum_{t=1}^Tf_{t}(\x_t^i)
  % =&\sum_{t=1}^T\sum_{j=1}^n\left(\alpha f_{t}^j(\x)-f_{t}^j(\x_t^i)\right)\\
  {\leq} \sum_{t=1}^T\sum_{j=1}^n\left(\alpha f_{t}^j(\x)-f_{t}^j(\x_t^j)+G\|\x_t^j-\x_t^i\|\right).
   % \leq& \sum_{t=1}^T\sum_{j=1}^n\left(\alpha f_{t}^j(\x)-f_{t}^j(\x_t^j)\right)+\sum_{t=1}^T\sum_{j=1}^nG\|\x_t^j-\x_t^i\|\\
  \end{split}
\end{equation}
% Let $\alpha=(1-\nu)/4$, where $\nu=\|\xx\|_\infty$. 
Recall that $\alpha=(1-\|\xx\|_\infty)/4$ and $\x_t^i=(\hat{\x}_t^i+\xx)/{2}$. Combining \eqref{thm1-eq1} with Lemma \ref{lem:non-oblvious-nonm}, for any $\x\in\K$, we have
\begin{equation}
\label{thm1-eq2}
\begin{split}
\alpha\sum_{t=1}^Tf_{t}(\x)-\sum_{t=1}^Tf_{t}(\x_t^i)  {\leq}& \sum_{t=1}^T\sum_{j=1}^n\left\langle\nabla F_t^j(\hat{\x}_t^j),\x-\hat{\x}_t^j\right\rangle+\sum_{t=1}^T\sum_{j=1}^nG\|\x_t^j-\x_t^i\|\\
  \leq &\sum_{t=1}^T\sum_{j=1}^n\left(\left\langle\nabla F_t^j(\hat{\x}_t^j),\x-\hat{\x}_t^i\right\rangle+\left(\left\|\nabla F_t^j(\hat{\x}_t^j)\right\|+\frac{G}{2}\right)\|\hat{\x}_t^j-\hat{\x}_t^i\|\right).
  % \leq &\sum_{t=1}^T\sum_{j=1}^n\left\langle\nabla F_t^j(\hat{\x}_t^j),\x-\hat{\x}_t^i\right\rangle+\sum_{t=1}^T\sum_{j=1}^nG\|\hat{\x}_t^j-\hat{\x}_t^i\|
  %+\sum_{t=1}^T\sum_{j=1}^n\frac{G}{2}\|\hat{\x}_t^j-\hat{\x}_t^i\|.
  % \leq &\sum_{t=1}^T\sum_{j=1}^n\left\langle\nabla F_t^j(\hat{\x}_t^j),\x-\hat{\x}_t^i\right\rangle+\sum_{t=1}^T\sum_{j=1}^n\left\|\nabla F_t^j(\hat{\x}_t^j)\right\|\|\hat{\x}_t^j-\hat{\x}_t^i\|+\sum_{t=1}^T\sum_{j=1}^nG\|\x_t^j-\x_t^i\|\\
   % \leq& \sum_{t=1}^T\sum_{j=1}^n\left(\alpha f_{t}^j(\x)-f_{t}^j(\x_t^j)\right)+\sum_{t=1}^T\sum_{j=1}^nG\|\x_t^j-\x_t^i\|\\
  \end{split}
\end{equation}
Moreover, from \eqref{definition: non-monotone gradient F} and the convexity of $\K$, for any $\x\in \mathcal{K}$, it is not hard to verify that 
\begin{equation}
\label{thm1-eq3}
\nabla F_t^i(\x)= \mathbb{E}_{z_t^i\sim\Z}\left.\left[\frac{3}{8}\tilde{\nabla} f_t^i\left(\frac{z_t^i}{2}(\x-\xx)+\xx\right)\right|\x\right].
\end{equation}
Combining \eqref{thm1-eq3} with Jensen's inequality and Assumption \ref{assumption3}, for any $\x\in \mathcal{K}$, we have
\begin{equation}
\label{thm1-eq4}
\left\|\nabla F_t^i(\x)\right\|\leq \mathbb{E}_{z_t^i\sim\Z}\left.\left[\left\|\frac{3}{8}\tilde{\nabla} f_t^i\left(\frac{z_t^i}{2}(\x-\xx)+\xx\right)\right\|\right|\x\right]\leq \frac{3G}{8}.
\end{equation}
% Due to \eqref{definition: non-monotone gradient F}, an unbiased estimation of $\nabla F_t^i(\x)$ can be computed by sampling $z_t^i$ from $Z$ and setting $\tilde{\nabla} F_t^i(\x)=\frac{3}{8}\tilde{\nabla} f_t^i(\x^\prime)$, where $\x^\prime=(z_t^i/2)(\x-\xx)+\xx$. 
Then, combining \eqref{thm1-eq2} with \eqref{thm1-eq4}, for any $\x\in \mathcal{K}$, we have
\begin{equation}
\label{thm1-eq5}
\begin{split}
\alpha\sum_{t=1}^Tf_{t}(\x)-\sum_{t=1}^Tf_{t}(\x_t^i)  
  \leq \sum_{t=1}^T\sum_{j=1}^n\left\langle\nabla F_t^j(\hat{\x}_t^j),\x-\hat{\x}_t^i\right\rangle+\sum_{t=1}^T\sum_{j=1}^nG\|\hat{\x}_t^j-\hat{\x}_t^i\|.
  % \leq &\sum_{t=1}^T\sum_{j=1}^n\left\langle\nabla F_t^j(\hat{\x}_t^j),\x-\hat{\x}_t^i\right\rangle+\sum_{t=1}^T\sum_{j=1}^nG\|\hat{\x}_t^j-\hat{\x}_t^i\|
  %+\sum_{t=1}^T\sum_{j=1}^n\frac{G}{2}\|\hat{\x}_t^j-\hat{\x}_t^i\|.
  % \leq &\sum_{t=1}^T\sum_{j=1}^n\left\langle\nabla F_t^j(\hat{\x}_t^j),\x-\hat{\x}_t^i\right\rangle+\sum_{t=1}^T\sum_{j=1}^n\left\|\nabla F_t^j(\hat{\x}_t^j)\right\|\|\hat{\x}_t^j-\hat{\x}_t^i\|+\sum_{t=1}^T\sum_{j=1}^nG\|\x_t^j-\x_t^i\|\\
   % \leq& \sum_{t=1}^T\sum_{j=1}^n\left(\alpha f_{t}^j(\x)-f_{t}^j(\x_t^j)\right)+\sum_{t=1}^T\sum_{j=1}^nG\|\x_t^j-\x_t^i\|\\
  \end{split}
\end{equation}
Finally, due to \eqref{thm1-eq3}, \eqref{thm1-eq5}, and $\ell_t^i(\x)=\langle-\tilde{\nabla} {F}_t^i(\hat{\x}_t^i), \x\rangle$, for any $\x\in \mathcal{K}$, we have
\begin{equation*}
\begin{split}
\E\left[\alpha\sum_{t=1}^Tf_{t}(\x)-\sum_{t=1}^Tf_{t}(\x_t^i) \right] 
  \leq \E\left[\sum_{t=1}^T\sum_{j=1}^n\left(\ell_t^j(\hat{\x}_t^i)-\ell_t^j(\x)\right)\right]+\E\left[\sum_{t=1}^T\sum_{j=1}^nG\|\hat{\x}_t^j-\hat{\x}_t^i\|\right].
  \end{split}
\end{equation*}

\subsection{Proof of Corollary \ref{cor1:label}}
Due to Assumption \ref{assumption3}, it is easy to verify that
\begin{equation}
\label{bound-estimated-stochastic}
\left\|\tilde{\nabla} {F}_t^i(\hat{\x}_t^i)\right\|= \frac{3}{8}\left\|\tilde{\nabla} {f}_t^i\left(\frac{z_t^i}{2}(\hat{\x}_t^i-\xx)+\xx\right)\right\|\leq \frac{3G}{8}
\end{equation}
which implies that the linear loss $\ell_t^i(\x)$ is $(3G/8)$-Lipschitz over $\K$. Because of this Lipschitz continuity, and Assumptions \ref{assum:bounded-set} and \ref{assumption4}, for any $\x\in\K$, we can directly use Theorem 1 of \citet{wan2024optimal} to derive the following upper bound
\begin{equation}
\label{cor1-eq1}
\sum_{t=1}^T\sum_{j=1}^n\left(\ell_t^j(\hat{\x}_t^i)-\ell_t^j(\x)\right)\leq \frac{189nCTG^2}{128h}+nhR^2
%\leq \frac{2nLTG^2}{h}+nhR^2
\end{equation}
where $C=\lceil\sqrt{2}\ln(\sqrt{14n})/((\sqrt{2}-1)\sqrt{\rho})\rceil$, and $h$ can be any positive constant.\footnote{Although \citet{wan2024optimal} actually assume that $\mathbf{0}\in\K$, it is trivial to verify that this assumption can be removed without any change to their original regret bounds, as well as the intermediate results on the consensus error. The only modification is to set the initial decision as $\argmin_{\x\in\K}\|\x\|$, instead of $\mathbf{0}$.}

Moreover, from (39), (40), and (41) in the proof of Theorem 1 of \citet{wan2024optimal}, it is also easy to derive the following consensus error bound
\begin{equation}
\label{cor1-eq2}
\sum_{t=1}^T\sum_{j=1}^n\|\hat{\x}_t^j-\hat{\x}_t^i\|\leq \frac{21nCTG}{8h}.
\end{equation}
By substituting \eqref{cor1-eq1} and \eqref{cor1-eq2} into Theorem \ref{DNSCR}, for any $\x\in\K$, we have
\begin{equation}
\label{cor1-eq3}
  \E\left[\alpha\sum_{t=1}^Tf_{t}(\x)-\sum_{t=1}^Tf_{t}(\x_t^i) \right] 
  \leq \frac{5nCTG^2}{h}+nhR^2.
\end{equation}
Finally, we can complete this proof by substituting $h=\sqrt{5CT}G/R$ into \eqref{cor1-eq3}.

\subsection{Proof of Corollary \ref{cor2:label}}
This proof is similar to the proof of Corollary \ref{cor1:label}. First, because of the Lipschitz continuity of $\ell_t^i(\x)$ derived from \eqref{bound-estimated-stochastic}, and Assumptions \ref{assum:bounded-set} and \ref{assumption4}, for any $\x\in\K$, we can directly use Theorem 5 of \citet{wan2024optimal} to derive the following upper bound
\begin{equation}
\label{cor2-eq1}
\sum_{t=1}^T\sum_{j=1}^n\left(\ell_t^j(\hat{\x}_t^i)-\ell_t^j(\x)\right)\leq \frac{243nLTG^2}{128h}+nhR^2+\frac{9nGRT}{2\sqrt{L+2}}
%\leq \frac{2nLTG^2}{h}+nhR^2
\end{equation}
where $L$ is an integer such that $T\geq L\geq C=\lceil\sqrt{2}\ln(\sqrt{14n})/((\sqrt{2}-1)\sqrt{\rho})\rceil$, and $h$ can be any positive constant.

Moreover, from the proof of Theorem 5 of \citet{wan2024optimal}, it is also easy to derive the following consensus error bound
\begin{equation}
\label{cor2-eq2}
\sum_{t=1}^T\sum_{j=1}^n\|\hat{\x}_t^j-\hat{\x}_t^i\|\leq \frac{27nLTG}{8h}+\frac{8nRT}{\sqrt{L+2}}.
\end{equation}
By substituting \eqref{cor2-eq1} and \eqref{cor2-eq2} into Theorem \ref{DNSCR}, for any $\x\in\K$, we have
\begin{equation}
\label{cor2-eq3}
  \E\left[\alpha\sum_{t=1}^Tf_{t}(\x)-\sum_{t=1}^Tf_{t}(\x_t^i) \right] 
  \leq \frac{6nLTG^2}{h}+nhR^2+\frac{13nGRT}{\sqrt{L}}.
\end{equation}
Finally, by substituting $h=\sqrt{6LT}G/R$ and $\sqrt{T}\leq L=\max\{\lceil\sqrt{T}\rceil,C\}\leq2\sqrt{T}+C$ into \eqref{cor2-eq3}, we have
\[
  \E\left[\Reg(T,i,\alpha) \right]\leq 2nGR\sqrt{6 TC}+(4\sqrt{3}+13)nGRT^{3/4}.
\]

% Note that as in \citet{wan2024optimal}, here we implicitly assume \[\sqrt{2}\ln(\sqrt{14n})/({(\sqrt{2}-1)\sqrt{1-\sigma_2(A)}})\leq \sqrt{T}\] 
% which is reasonable because $T$ is commonly much larger
% than other problem constants.

\subsection{Proof of Theorem \ref{thm: DSCR-smooth}}
As proved in \citet{chen2018online}, for any $\beta$-smooth function $f(\x):\K\mapsto\mathbb{R}$ over the convex set $\K$, it holds that
\begin{equation}
\label{smooth-property1}
|f(\y)-f(\x)-\langle\nabla f(\x),\y-\x \rangle|\leq \frac{\beta}{2}\|\x-\y\|^2,\forall\x,\y\in\K.
\end{equation}
Due to $f_t(\x)=\sum_{j=1}^nf_t^j(\x)$, Assumption \ref{assum:smooth}, and \eqref{smooth-property1}, for any $\x\in\K$, we have
\begin{equation}
\label{thm1-smooth-eq0}
\begin{split}
  \alpha\sum_{t=1}^Tf_{t}(\x)-\sum_{t=1}^Tf_{t}(\x_t^i)
  % =&\sum_{t=1}^T\sum_{j=1}^n\left(\alpha f_{t}^j(\x)-f_{t}^j(\x_t^i)\right)\\
  {\leq} \sum_{t=1}^T\sum_{j=1}^n\left(\alpha f_{t}^j(\x)-f_{t}^j(\x_t^j)+\langle\nabla f_t^j(\x_t^j),\x_t^j-\x_t^i\rangle+\frac{\beta}{2}\|\x_t^i-\x_t^j\|^2\right).
   % \leq& \sum_{t=1}^T\sum_{j=1}^n\left(\alpha f_{t}^j(\x)-f_{t}^j(\x_t^j)\right)+\sum_{t=1}^T\sum_{j=1}^nG\|\x_t^j-\x_t^i\|\\
  \end{split}
\end{equation}
Recall that $\alpha=(1-\|\xx\|_\infty)/4$ and $\x_t^i=(\hat{\x}_t^i+\xx)/{2}$. Combining \eqref{thm1-smooth-eq0} with Lemma \ref{lem:non-oblvious-nonm}, for any $\x\in\K$, we have
\begin{equation} 
\label{thm1-smooth-eq1}
\begin{split}
&\alpha\sum_{t=1}^Tf_{t}(\x)-\sum_{t=1}^Tf_{t}(\x_t^i)  \\
\leq & \sum_{t=1}^T\sum_{j=1}^n\left(\left\langle\nabla F_t^j(\hat{\x}_t^j),\x-\hat{\x}_t^j\right\rangle+\frac{1}{2}\langle\nabla f_t^j(\x_t^j),\hat{\x}_t^j-\hat{\x}_t^i\rangle+\frac{\beta}{8}\|\hat{\x}_t^i-\hat{\x}_t^j\|^2\right)\\
\leq & \sum_{t=1}^T\sum_{j=1}^n\left(\left\langle\nabla F_t^j(\hat{\x}_t^j),\x-\hat{\x}_t^i\right\rangle+\left\langle\frac{1}{2}\nabla f_t^j(\x_t^j)-\nabla F_t^j(\hat{\x}_t^j),\hat{\x}_t^j-\hat{\x}_t^i\right\rangle+\frac{\beta}{8}\|\hat{\x}_t^i-\hat{\x}_t^j\|^2\right).
  \end{split}
\end{equation}
Finally, combining \eqref{thm1-smooth-eq1} with \eqref{thm1-eq3} and $\ell_t^i(\x)=\langle-\tilde{\nabla} {F}_t^i(\hat{\x}_t^i), \x\rangle$, for any $\x\in \mathcal{K}$, we have
\begin{equation*}
\begin{split}
\E\left[\alpha\sum_{t=1}^Tf_{t}(\x)-\sum_{t=1}^Tf_{t}(\x_t^i) \right]\leq& \E\left[\sum_{t=1}^T\sum_{j=1}^n\left(\ell_t^j(\hat{\x}_t^i)-\ell_t^j(\x)\right)\right]+\E\left[\sum_{t=1}^T\sum_{j=1}^n\frac{\beta}{8}\|\hat{\x}_t^i-\hat{\x}_t^j\|^2\right]\\
&+\E\left[\sum_{t=1}^T\sum_{j=1}^n\left\langle\frac{1}{2}\nabla f_t^j(\x_t^j)-\nabla F_t^j(\hat{\x}_t^j),\hat{\x}_t^j-\hat{\x}_t^i\right\rangle\right].
  \end{split}
\end{equation*}

\subsection{Proof of Corollary \ref{cor3:label}}
To distinguish the decisions of Algorithms \ref{DNSCR} and \ref{AD-OSPA}, in this proof, we denote the decision $\x_q^i$ in Algorithm \ref{AD-OSPA} for any $q\in[T/L]$ and $i\in[n]$ as $\u_q^i$, where $L$ will be specified later. This implies that $\hat{\x}_t^i$ in Algorithm \ref{DNSCR} equals to $\u_q^i$ for any $t\in \mathcal{T}_q$, where $\mathcal{T}_q=\{(q-1)L+1,\dots,qL\}$.

Then, due to \eqref{bound-estimated-stochastic}, and Assumption \ref{assum:bounded-set}, for any $\x\in\K$, we can apply Lemma \ref{lem1-AD-OSPF:label} to derive that
\begin{equation}
\label{cor3-eq1}
\begin{split}
&\sum_{t=1}^T\sum_{j=1}^n\left(\ell_t^j(\hat{\x}_t^i)-\ell_t^j(\x)\right)=\sum_{q=1}^{T/L}\sum_{t\in\mathcal{T}_q}\sum_{j=1}^n\langle \nabla \ell_{t}^j(\u_q^j),\u_q^i-\x\rangle\\
\leq& 2\eta nR+\frac{9dnRTLG^2}{128\eta}+\sum_{q=1}^{T/L}\sum_{t\in\mathcal{T}_q}\sum_{j=1}^n\langle \nabla \ell_{t}^j(\u_q^j),\u_q^i-\bar{\y}_q\rangle\\
=&2\eta nR+\frac{9dnRTLG^2}{128\eta}+\sum_{q=1}^{T/L}\sum_{t\in\mathcal{T}_q}\sum_{j=1}^n\langle \tilde{\nabla} {F}_t^j(\u_q^j),\bar{\y}_q-\u_q^i\rangle
\end{split}
\end{equation}
where $\bar{\y}_q$ follows the definition in Lemma \ref{lem1-AD-OSPF:label}, $\eta$ will be specified later, and the first and last equalities are due to the definition of $\ell_t^j(\x)$.

By substituting \eqref{cor3-eq1} into Theorem \ref{thm: DSCR-smooth}, for any $\x\in\K$, we have
\begin{equation}
\label{cor3-eq2}
\begin{split}
&\E\left[\alpha\sum_{t=1}^Tf_{t}(\x)-\sum_{t=1}^Tf_{t}(\x_t^i) \right]-\left(2\eta nR+\frac{9dnRTLG^2}{128\eta}\right)\\
\leq& 
\E\left[\sum_{q=1}^{T/L}\sum_{t\in\mathcal{T}_q}\sum_{j=1}^n\langle \tilde{\nabla} {F}_t^j(\u_q^j),\bar{\y}_q-\u_q^i\rangle\right]+\E\left[\sum_{t=1}^T\sum_{j=1}^n\frac{\beta}{8}\|\hat{\x}_t^i-\hat{\x}_t^j\|^2\right]\\
&+\E\left[\sum_{t=1}^T\sum_{j=1}^n\left\langle\frac{1}{2}\nabla f_t^j(\x_t^j)-\nabla F_t^j(\hat{\x}_t^j),\hat{\x}_t^j-\hat{\x}_t^i\right\rangle\right]\\
% =& 
% \E\left[\sum_{q=1}^{T/L}\sum_{t\in\mathcal{T}_q}\sum_{j=1}^n\langle {\nabla} {F}_t^j(\u_q^j),\bar{\y}_q-\u_q^i\rangle\right]+\E\left[\sum_{q=1}^{T/L}\sum_{t\in\mathcal{T}_q}\sum_{j=1}^n\frac{\beta}{8}\|\u_q^i-\u_q^j\|^2\right]\\
% &+\E\left[\sum_{q=1}^{T/L}\sum_{t\in\mathcal{T}_q}\sum_{j=1}^n\left\langle\frac{1}{2}\nabla f_t^j\left(\frac{\u_q^j+\xx}{2}\right),\u_q^j-\u_q^i\right\rangle\right]+\E\left[\sum_{q=1}^{T/L}\sum_{t\in\mathcal{T}_q}\sum_{j=1}^n\left\langle\nabla F_t^j(\u_q^j),\u_q^i-\u_q^j\right\rangle\right]\\
=& 
\E\left[\sum_{q=1}^{T/L}\sum_{t\in\mathcal{T}_q}\sum_{j=1}^n\langle {\nabla} {F}_t^j(\u_q^j),\bar{\y}_q-\u_q^j\rangle\right]+\E\left[\sum_{q=1}^{T/L}\sum_{t\in\mathcal{T}_q}\sum_{j=1}^n\frac{\beta}{8}\|\u_q^i-\u_q^j\|^2\right]\\
&+\E\left[\sum_{q=1}^{T/L}\sum_{t\in\mathcal{T}_q}\sum_{j=1}^n\left\langle\frac{1}{2}\nabla f_t^j\left(\frac{\u_q^j+\xx}{2}\right),\u_q^j-\u_q^i\right\rangle\right]
  \end{split}
\end{equation}
where the equality is due to the definition of $\u_q^i$ and $\E[\tilde{\nabla} {F}_t^j(\u_q^j),\bar{\y}_q-\u_q^i]=\E[\nabla {F}_t^j(\u_q^j),\bar{\y}_q-\u_q^i]$ derived from \eqref{thm1-eq3} for any $t\in\mathcal{T}_q$.

For the second term in the right side of \eqref{cor3-eq2}, due to \eqref{bound-estimated-stochastic}, Lemma \ref{lem2-AD-OSPF:label} and $\|\u_q^i-\u_q^j\|^2\leq 4R^2$ for any $q\leq 2$, we have
\begin{equation}
\label{cor3-eq3}
\begin{split}
\E\left[\sum_{q=1}^{T/L}\sum_{t\in\mathcal{T}_q}\sum_{j=1}^n\frac{\beta}{8}\|\u_q^i-\u_q^j\|^2\right]\leq \frac{\beta nT}{4}\left(\frac{16R^2}{L}+\frac{9 dLGR^2}{\eta}\right) + n\beta LR^2.
\end{split}
\end{equation}
To analyze the first term in the right side of \eqref{cor3-eq2}, we introduce a lemma about $F_t^i(\x)$.
\begin{lemma}
\label{lem2:non-oblvious-nonm}
(Theorem 18 of \citet{Zhang-Arxiv24}) Under Assumptions \ref{assum:bounded-set}, \ref{assum:DR-submodular}, and \ref{assum:smooth}, for any $i\in[n]$ and $t\in[T]$, the function $F_t^i(\x)$ defined in Lemma \ref{lem:non-oblvious-nonm} can be written as
\begin{equation*}
     F_t^i(\x)= \int_0^1\frac{1}{4z\left(1-\frac{z}{2})\right)^3}\left(f_t^i\left(\frac{z}{2}(\x-\xx)+\xx\right)-f_t^i(\xx)\right)dz
\end{equation*}
and it is $(\beta/8)$-smooth over $\mathcal{X}$.
\end{lemma}
Due to the oblivious assumption of $f_t^i(\x)$, the function $F_t^i(\x)$ given by Lemma \ref{lem2:non-oblvious-nonm} is also oblivious. Combining this property, Lemma \ref{lem2:non-oblvious-nonm}, \eqref{thm1-eq4}, and \eqref{bound-estimated-stochastic}, we can use Lemma \ref{lem2-AD-OSPF:label} and $\langle {\nabla} {F}_t^j(\u_q^j),\bar{\y}_q-\u_q^j\rangle\leq 3RG/4$ for any $q\leq 2$ to derive that
\begin{equation}
\label{cor3-eq4}
\begin{split}
\E\left[\sum_{q=1}^{T/L}\sum_{t\in\mathcal{T}_q}\sum_{j=1}^n\langle {\nabla} {F}_t^j(\u_q^j),\bar{\y}_q-\u_q^j\rangle\right]\leq nT\left(\frac{\beta R^2}{2L}+\frac{45 dLRG^2}{64\eta}\right)+\frac{3nRGL}{2}.
\end{split}
\end{equation}
Similarly, for any $q\geq 3$ and $t\in\mathcal{T}_q$, due to the oblivious assumption and $\beta$-smoothness of $f_t^i(\x)$, \eqref{bound-gradient}, and \eqref{bound-estimated-stochastic}, we can use Lemma \ref{lem2-AD-OSPF:label} again to derive that
\begin{equation}
\label{cor3-eq5}
\begin{split}
&\E\left[\left\langle\nabla f_t^j\left(\frac{\u_q^j+\xx}{2}\right),\u_q^j-\u_q^i\right\rangle\right]\\
\leq&\E\left[\left\langle\nabla f_t^j\left(\frac{\u_q^j+\xx}{2}\right),\u_q^j-\bar{\y}_q\right\rangle\right]+\E\left[\left\langle\nabla f_t^j\left(\frac{\u_q^j+\xx}{2}\right),\bar{\y}_q-\u_q^i\right\rangle\right]\\
\leq &\frac{4\beta R^2}{L}+\frac{15 dLRG^2}{4\eta}.
\end{split}
\end{equation}
Combining \eqref{cor3-eq5} with $\langle\nabla f_t^j((\u_q^j+\xx)/2),\u_q^j-\u_q^i\rangle\leq 2RG$ for any $q\leq 2$, it is easy to verify that
\begin{equation*}
% \label{cor3-eq6}
\begin{split}
\E\left[\sum_{q=1}^{T/L}\sum_{t\in\mathcal{T}_q}\sum_{j=1}^n\left\langle\frac{1}{2}\nabla f_t^j\left(\frac{\u_q^j+\xx}{2}\right),\u_q^j-\u_q^i\right\rangle\right]\leq nT\left(\frac{2\beta R^2}{L}+\frac{15 dLRG^2}{8\eta}\right)+2nRGL.
\end{split}
\end{equation*}
Finally, by substituting \eqref{cor3-eq3}, \eqref{cor3-eq4}, and the above inequality into \eqref{cor3-eq2}, we have
\begin{equation*}
\begin{split}
\E\left[\Reg(T,i,\alpha) \right]\leq &2\eta nR+\frac{dnRTLG}{\eta}\left(\frac{339G}{128}+\frac{9\beta R}{4}\right)+\frac{13\beta nTR^2}{2L}+nRL\left(\frac{7G}{2}+\beta R\right)\\
\leq &8n(\sqrt{2}T^{2/3}+\sqrt{TC})\sqrt{d}R^\prime+7\beta nT^{2/3}R^2+4n(2T^{1/3}+C)R^\prime
  \end{split}
\end{equation*}
where the last inequality is due to $\eta=G^\prime \sqrt{dTL}$, $G^\prime = 3G/8$, $T^{1/3}\leq L=\max\{\lceil T^{1/3}\rceil,C\}\leq 2T^{1/3}+ C$, and $R^\prime=R(G+\beta R)$. Note that these parameters are following Theorem \ref{thm1-AD-OSPF} with a simple adjustment about the upper bound on the gradient norm in \eqref{bound-estimated-stochastic}.

\section{Analysis of Algorithm \ref{AD-OSPA}}
In this section, we present detailed proofs of the theoretical guarantees for Algorithm \ref{AD-OSPA}.

% provide the detailed proofs of theoretical guarantees of Algorithm 

\subsection{Proof of Lemma \ref{lem1-AD-OSPF:label}}
% \begin{equation*}
% % \label{ideal-decisions}
% \y_{q}^i=\mathbb{E}_{\v\in\B}\left[\argmax_{\x\in\K}\left\langle-\z_{q}^i+\frac{1}{\eta}\v,\x\right\rangle\right],\forall i\in[n]
% \end{equation*}
For any $\x\in\K$, we first have
\begin{equation}
\label{thm-AD-OSPF-eq1}
\begin{split}
  \sum_{q=1}^{T/L}\sum_{t\in\mathcal{T}_q}\sum_{j=1}^n\langle \nabla f_{t}^j(\x_q^j),\x_q^i-\x\rangle=&\sum_{q=1}^{T/L}\sum_{t\in\mathcal{T}_q}\sum_{j=1}^n\langle \nabla f_{t}^j(\x_q^j),\bar{\y}_q-\x+\x_q^i-\bar{\y}_q\rangle\\
  =&\sum_{q=1}^{T/L}n\langle \bar{\g}_q,\bar{\y}_q-\x\rangle+\sum_{q=1}^{T/L}\sum_{t\in\mathcal{T}_q}\sum_{j=1}^n\langle \nabla f_{t}^j(\x_q^j),\x_q^i-\bar{\y}_q\rangle.
\end{split}
\end{equation}
We notice that $\bar{\y}_1,\dots,\bar{\y}_{T/L}$ are actually the decisions generated by running an expected version of FTPL, i.e., Algorithm 3 in \citet{hazan2020faster}, over linear losses $\langle\bar{\g}_1,\x\rangle,\dots,\langle\bar{\g}_{T/L},\x\rangle$ and the decision set $\K$. Moreover, a regret bound of this algorithm over these linear losses has been derived in the proof of their Theorem 10.
\begin{lemma}
\label{lem-opfs-hazan} 
(Derived from the proof of Theorem 10 in \citet{hazan2020faster}) Let $\ell_1(\x)=\langle\nabla_1,\x\rangle,\dots,\ell_T(\x)=\langle\nabla_T,\x\rangle$ be a sequence of linear losses. Moreover, let $G=\max_{t\in[T]}\|\nabla_t\|$, $R=\max_{\x\in\K}\|\x\|$, and $\x_t^\ast=\E_{\v\in\B}[\argmax_{\x\in\K}\langle-\sum_{\tau=1}^{t-1}\nabla_\tau+\eta\v,\x\rangle]$ for any $t\in[T]$. Suppose the decision set $\K$ is convex, for any $\x\in\K$, we have
\[\sum_{t=1}^T\ell_t(\x_t^\ast)-\sum_{t=1}^T\ell_t(\x)\leq 2\eta R+\frac{dRTG^2}{2\eta}.\]
\end{lemma}
Due to Lemma \ref{lem-opfs-hazan} and $\|\bar{\g}_q\|\leq LG$, for any $\x\in\K$, we have
\begin{equation}
\label{thm-AD-OSPF-eq2}
\begin{split}
\sum_{q=1}^{T/L}\langle \bar{\g}_q,\bar{\y}_q-\x\rangle\leq 2\eta R+\frac{ dRTLG^2}{2\eta}.
\end{split}
\end{equation}
Finally, this proof can be completed by substituting \eqref{thm-AD-OSPF-eq2} into \eqref{thm-AD-OSPF-eq1}.

\subsection{Proof of Lemma \ref{lem2-AD-OSPF:label}}
We first introduce three useful lemmas, where the last one holds because Algorithm \ref{AD-OSPA} maintains $\z_q^i$ by following the same way of AD-FTRL in \citet{wan2024optimal}.
\begin{lemma}
\label{lem15hazan:label}
(Lemma 15 of \citet{hazan2020faster}) Let $Z_1,\dots,Z_L$ be i.i.d.~samples of a bounded random vector $Z\in\mathbb{R}^d$ with $\|Z\|\leq R$, and $\bar{Z}_L=\frac{1}{L}\sum_{m=1}^LZ_m$. For $\bar{Z}=\E[Z]$, we have \[\E_{Z_1,\dots,Z_K}\left[\|\bar{Z}_L-\bar{Z}\|^2\right]\leq\frac{4R^2}{L}.\]
\end{lemma}
\begin{lemma}
\label{lem1_ospf}
(Lemma 10 of \citet{Wan-Arxiv-DelayedOFW}) 
Let $L(\u)=\E_{\v\in\B}\left[\argmax_{\x\in\K}\left\langle\u+\eta \v,\x\right\rangle\right]$ and $R=\max_{\x\in\K}\|\x\|$. If the decision set $\K$ is convex, we have \[\|L(\u)-L(\y)\|\leq\frac{dR}{\eta}\|\u-\y\|,\forall\u,\y\in\mathbb{R}^d.\]
\end{lemma}
\begin{lemma}
\label{lem-fastMix}
(Lemma 2 of \citet{wan2024optimal})
Let $\bar{\z}_q=\sum_{\tau=1}^{q-1}\bar{\g}_\tau$, where $\bar{\g}_q=\frac{1}{n}\sum_{i=1}^n\g_q^i$. Moreover, let $G=\max_{\x\in\K,i\in[n],t\in[T]}\|\nabla f_t^i(\x)\|$. If Assumption \ref{assumption4} holds, Algorithm \ref{AD-OSPA} with $\theta=1/(1+\sqrt{1-\sigma_2^2(A)})$, $K=C$ ensures \[\left\|\z_q^i-\bar{\z}_q\right\|\leq 3LG,\forall i\in[n],q\in[T/L].\]
\end{lemma}
Then, let $\y_{q}^i=\mathbb{E}_{\v\in\B}\left[\argmax_{\x\in\K}\left\langle-\z_{q}^i+\eta \v,\x\right\rangle\right]$ for any $i\in[n],q\in[T/L]$. Moreover, let $\xi_q$ denote the randomness introduced at each block $q$ of Algorithm \ref{AD-OSPA}, and $\xi_{1:q}$ denote the randomness introduced at the first $q$ blocks. Note that conditioned on the randomness $\xi_{1:q-1}$, the sample $\x_{q+2,m}^i$ in Algorithm \ref{AD-OSPA} is an unbiased estimation of $\y_{q}^i$ for any $m\in[L]$. Therefore, for any $q=3,\dots,T/L$ and $i\in[n]$, we have
\begin{equation}
\label{lem2-AD-OSPF-eq1-pre}
\begin{split}
\E\left[\|{\x}_q^i-\bar{\y}_{q-2}\|^2\right]\leq&\E\left[2\|{\x}_q^i-{\y}^i_{q-2}\|^2+2\|{\y}^i_{q-2}-\bar{\y}_{q-2}\|^2\right]\\
=& \E\left[\E\left[2\|{\x}_q^i-{\y}^i_{q-2}\|^2|\xi_{1:q-3}\right]\right]+\E\left[2\|{\y}^i_{q-2}-\bar{\y}_{q-2}\|^2\right]\\
\leq&\frac{8R^2}{L}+\E\left[2\|{\y}^i_{q-2}-\bar{\y}_{q-2}\|^2\right]\\
\leq &\frac{8R^2}{L}+\E\left[4R\|{\y}^i_{q-2}-\bar{\y}_{q-2}\|\right]
\end{split}
\end{equation}
where the second inequality is due to Lemma \ref{lem15hazan:label} and $\x_{q}^i=\frac{1}{L}\sum_{m=1}^L\x_{q,m}^i$. Combining \eqref{lem2-AD-OSPF-eq1-pre} with Lemmas \ref{lem1_ospf} and \ref{lem-fastMix}, for any $q=3,\dots,T/L$ and $i\in[n]$, we have
\begin{equation*}
\begin{split}
\E\left[\|{\x}_q^i-\bar{\y}_{q-2}\|^2\right]\leq& \frac{8R^2}{L}+\E\left[\frac{4 dR^2}{\eta}\|\z_{q-2}^i-\bar{\z}_{q-2}\|\right]\leq \frac{8R^2}{L}+\frac{12 dLGR^2}{\eta}.
\end{split}
\end{equation*}
From the above inequality, for any $q=3,\dots,T/L$ and $i,j\in[n]$, it is easy to verify that
\begin{equation*}
\begin{split}
\E\left[\|{\x}_q^i-{\x}_q^j\|^2\right]\leq \E\left[2\|{\x}_q^i-\bar{\y}_{q-2}\|^2+2\|{\x}_q^j-\bar{\y}_{q-2}\|^2\right]\leq \frac{32R^2}{L}+\frac{48 dLGR^2}{\eta}.
\end{split}
\end{equation*}

Next, we proceed to analyze $\langle \nabla f(\tilde{\x}_q^j),\x_q^i-\bar{\y}_q\rangle$. For any $q=3,\dots,T/L$ and $i,j\in[n]$, it is not hard to verify that
\begin{equation}
\label{lem2-AD-OSPF-eq2-pre}
\begin{split}
&\E\left[\langle \nabla f(\tilde{\x}_q^j),\x_q^i-\bar{\y}_q\rangle\right]=\E\left[\langle \nabla f(\lambda \x_q^j+(1-\lambda)\xx),\x_q^i-{\y}^i_{q-2}+{\y}^i_{q-2}-\bar{\y}_q\rangle\right]\\
\leq &\E\left[\E\left[\langle \nabla f(\lambda \x_q^j+(1-\lambda)\xx),\x_q^i-{\y}^i_{q-2}\rangle|\xi_{1:q-3}\right]\right]+\E\left[G^\prime\|{\y}^i_{q-2}-\bar{\y}_q\|\right].
\end{split}
\end{equation}
Note that conditioned on the randomness $\xi_{1:q-3}$, $\x_q^i$ is independent of $\x_q^j$ for any $i\neq j\in[n]$, and it is also independent of ${\y}^i_{q-2}$. Moreover, due to the oblivious assumption, the function $f(\x)$ is independent of any randomness of Algorithm \ref{AD-OSPA}. Therefore, for any $q=3,\dots,T/L$ and $i\neq j\in[n]$, by using $\x_{q}^i=\frac{1}{L}\sum_{m=1}^L\x_{q,m}^i$ and the unbiasedness of $\x_{q,m}^i$, we can simplify \eqref{lem2-AD-OSPF-eq2-pre} to
\begin{equation}
\label{lem2-AD-OSPF-eq2}
\begin{split}
\E\left[\langle \nabla f(\tilde{\x}_q^j),\x_q^i-\bar{\y}_q\rangle\right]\leq &\E\left[G^\prime\|{\y}^i_{q-2}-\bar{\y}_q\|\right]\leq \E\left[G^\prime(\|{\y}^i_{q-2}-\bar{\y}_{q-2}\|+\|\bar{\y}_{q-2}-\bar{\y}_{q}\|)\right]\\
\leq& \frac{dRG^\prime}{\eta}\left(\E\left[\|{\z}^i_{q-2}-\bar{\z}_{q-2}\|\right]+\E\left[\|\bar{\z}_{q-2}-\bar{\z}_{q}\|\right]\right)\leq  \frac{5 dLRGG^\prime}{\eta}
\end{split}
\end{equation}
where the third inequality is due to Lemma \ref{lem1_ospf}, and the last one is due to Lemma \ref{lem-fastMix} and \[\|\bar{\z}_{q-2}-\bar{\z}_{q}\|=\|\bar{\g}_{q-2}+\bar{\g}_{q-1}\|\leq 2LG.\]

Otherwise, we consider the case with $i=j$. Let $\tilde{\y}^i_{q-2}=\lambda{\y}^i_{q-2}+(1-\lambda)\xx$. For any $q=3,\dots,T/L$ and $i\in[n]$, by combining \eqref{lem2-AD-OSPF-eq2-pre} with $\x_{q}^i=\frac{1}{L}\sum_{m=1}^L\x_{q,m}^i$ and the unbiasedness of $\x_{q,m}^i$, it is not hard to verify that
\begin{equation}
\label{lem2-AD-OSPF-eq3}
\begin{split}
&\E\left[\langle \nabla f(\tilde{\x}_q^i),\x_q^i-\bar{\y}_q\rangle\right]\\
\leq &\E\left[\E\left[\langle \nabla f(\lambda \x_q^i+(1-\lambda)\xx)-\nabla f(\tilde{\y}^i_{q-2}),\x_q^i-{\y}^i_{q-2}\rangle|\xi_{1:q-3}\right]\right]+\E\left[G^\prime\|{\y}^i_{q-2}-\bar{\y}_q\|\right]\\
\leq &\E\left[\E\left[\beta\lambda\|\x_q^i-{\y}^i_{q-2}\|^2|\xi_{1:q-3}\right]\right]+\E\left[G^\prime\|{\y}^i_{q-2}-\bar{\y}_q\|\right]\leq\frac{4\beta\lambda R^2}{L}+\frac{5dLRGG^\prime}{\eta }
\end{split}
\end{equation}
where the second inequality is due to the $\beta$-smoothness of $f(\x)$, and the last inequality is due to Lemma \ref{lem15hazan:label} and the upper bound of $\E\left[G^\prime\|\bar{\y}_{q-2}-\bar{\y}_{q}\|\right]$ in \eqref{lem2-AD-OSPF-eq2}.

Finally, combining \eqref{lem2-AD-OSPF-eq2} with \eqref{lem2-AD-OSPF-eq3} and noticing that they also hold if replacing $\x_q^i-\bar{\y}_q$ by $\bar{\y}_q-\x_q^i$, for any $q=3,\dots,T/L$ and $i,j\in[n]$, we have
\begin{equation*}
\begin{split}
\E\left[\left|\langle \nabla f(\tilde{\x}_q^j),\x_q^i-\bar{\y}_q\rangle\right|\right]
\leq&\frac{4\beta\lambda R^2}{L}+\frac{5dLRGG^\prime}{\eta }.
\end{split}
\end{equation*}

\subsection{Proof of Theorem \ref{thm1-AD-OSPF}}
Due to $f_t(\x)=\sum_{j=1}^nf_t^j(\x)$, the $\beta$-smoothness of $f_t^i(\x)$, and \eqref{smooth-property1}, for any $\x\in\K$ and $q\in[T/L]$, it is not hard to verify that
\begin{equation}
\label{thm-DSPF-eq1}
\begin{split}
  \sum_{t\in\mathcal{T}_q}\left( f_{t}(\x_q^i)-f_{t}(\x)\right){\leq} \sum_{t\in\mathcal{T}_q}\sum_{j=1}^n\left(f_{t}^j(\x_q^j)- f_{t}^j(\x)+\langle\nabla f_t^j(\x_q^j),\x_q^i-\x_q^j\rangle+\frac{\beta}{2}\|\x_q^i-\x_q^j\|^2\right).
   % {\leq}& \sum_{q=1}^{T/L}\sum_{t\in\mathcal{T}_q}\sum_{j=1}^n\left(\langle\nabla f_{t}^j(\x_q^j), \x_q^j-\x \rangle +\langle\nabla f_t^j(\x_q^j),\x_q^i-\x_q^j\rangle+\frac{\beta}{2}\|\x_q^i-\x_q^j\|^2\right)\\
   % =& \sum_{q=1}^{T/L}\sum_{t\in\mathcal{T}_q}\sum_{j=1}^n\left(\langle\nabla f_{t}^j(\x_q^j), \x_q^i-\x \rangle+\frac{\beta}{2}\|\x_q^i-\x_q^j\|^2\right)
   % \leq& \sum_{t=1}^T\sum_{j=1}^n\left(\alpha f_{t}^j(\x)-f_{t}^j(\x_t^j)\right)+\sum_{t=1}^T\sum_{j=1}^nG\|\x_t^j-\x_t^i\|\\
  \end{split}
\end{equation}
Moreover, the convexity of $f_t^i(\x)$ implies that $f_{t}^j(\x_q^j)- f_{t}^j(\x)\leq \langle\nabla f_{t}^j(\x_q^j), \x_q^j-\x \rangle$ for any $\x\in\K$. By substituting it into \eqref{thm-DSPF-eq1} and summing over $q=1,\dots,T/L$, for any $\x\in\K$, we have
\begin{equation}
\label{thm-DSPF-eq2}
\begin{split}
\sum_{q=1}^{T/L}\sum_{t\in\mathcal{T}_q}\left( f_{t}(\x_q^i)-f_{t}(\x)\right)\leq \sum_{q=1}^{T/L}\sum_{t\in\mathcal{T}_q}\sum_{j=1}^n\left(\langle\nabla f_{t}^j(\x_q^j), \x_q^i-\x \rangle+\frac{\beta}{2}\|\x_q^i-\x_q^j\|^2\right).
   % \leq& \sum_{t=1}^T\sum_{j=1}^n\left(\alpha f_{t}^j(\x)-f_{t}^j(\x_t^j)\right)+\sum_{t=1}^T\sum_{j=1}^nG\|\x_t^j-\x_t^i\|\\
  \end{split}
\end{equation}
Combining \eqref{thm-DSPF-eq2} with Lemma \ref{lem1-AD-OSPF:label}, we have
\begin{equation}
\label{thm-DSPF-eq3}
\begin{split}
\sum_{q=1}^{T/L}\sum_{t\in\mathcal{T}_q}\left( f_{t}(\x_q^i)-f_{t}(\x)\right)\leq &2\eta nR+\frac{ dnRTLG^2}{2\eta}+\sum_{q=1}^{T/L}\sum_{t\in\mathcal{T}_q}\sum_{j=1}^n\langle \nabla f_{t}^j(\x_q^j),\x_q^i-\bar{\y}_q\rangle\\
&+\sum_{q=1}^{T/L}\sum_{t\in\mathcal{T}_q}\sum_{j=1}^n\frac{\beta}{2}\|\x_q^i-\x_q^j\|^2
   % \leq& \sum_{t=1}^T\sum_{j=1}^n\left(\alpha f_{t}^j(\x)-f_{t}^j(\x_t^j)\right)+\sum_{t=1}^T\sum_{j=1}^nG\|\x_t^j-\x_t^i\|\\
  \end{split}
\end{equation}
where $\bar{\y}_q$ follows the definition in Lemma \ref{lem1-AD-OSPF:label}.

Due to the $\beta$-smoothness of $f_t^j(\x)$, the definition of $G$, and the oblivious assumption, we can use Lemma \ref{lem2-AD-OSPF:label}  and $\langle \nabla f_{t}^j(\x_q^j),\x_q^i-\bar{\y}_q\rangle\leq 2GR$ for any $q\leq2$ to derive that
\begin{equation}
\label{thm-DSPF-eq4}
\begin{split}
\E\left[\sum_{q=1}^{T/L}\sum_{t\in\mathcal{T}_q}\sum_{j=1}^n\langle \nabla f_{t}^j(\x_q^j),\x_q^i-\bar{\y}_q\rangle\right]\leq nT\left(\frac{4\beta  R^2}{L}+\frac{5 dLRG^2}{\eta}\right)+4nLGR.
  \end{split}
\end{equation}
Moreover, by using Lemma \ref{lem2-AD-OSPF:label} again and $\|\x_q^i-\x_q^j\|^2\leq 4R^2$ for any $q\leq2$, we also have
\begin{equation}
\label{thm-DSPF-eq5}
\begin{split}
\E\left[\sum_{q=1}^{T/L}\sum_{t\in\mathcal{T}_q}\sum_{j=1}^n\frac{\beta}{2}\|\x_q^i-\x_q^j\|^2\right]\leq n\beta T\left(\frac{16R^2}{L}+\frac{24 dLGR^2}{\eta}\right)+4n\beta LR^2.
  \end{split}
\end{equation}
Finally, combining \eqref{thm-DSPF-eq3} with \eqref{thm-DSPF-eq4}, \eqref{thm-DSPF-eq5}, and $R^\prime=R(G+\beta R)$, we have
\begin{equation*}
\begin{split}
\E\left[\Reg(T,i)\right]\leq& 2\eta nR+\frac{6 dnRTLG\left(G+4\beta R\right)}{\eta}+\frac{20\beta nTR^2}{L}+4nLR(G+\beta R)\\
\leq &24n(\sqrt{2}T^{2/3}+\sqrt{TC})\sqrt{d}R^\prime+20\beta nT^{2/3}R^2+4n(2T^{1/3}+C)R^\prime
  \end{split}
\end{equation*}
where the last inequality is due to $\eta=G\sqrt{dTL}$ and $T^{1/3}\leq L=\max\{\lceil T^{1/3}\rceil,C\}\leq 2T^{1/3}+C$.

\subsection{Proof of Theorem \ref{thm2-AD-OSPF}}
Due to $f_t(\x)=\sum_{j=1}^nf_t^j(\x)$ and $f_t^i(\x)=\langle\c_t^i,\x\rangle$, for any $\x\in\K$, we have
\begin{equation}
\label{thm2-DSPF-eq1}
\begin{split}
\sum_{q=1}^{T/L}\sum_{t\in\mathcal{T}_q}\left( f_{t}(\x_q^i)-f_{t}(\x)\right)=&\sum_{q=1}^{T/L}\sum_{t\in\mathcal{T}_q}\sum_{j=1}^n\langle\c_t^j, \x_q^i-\x \rangle\\
=&\sum_{q=1}^{T/L}\sum_{t\in\mathcal{T}_q}\sum_{j=1}^n\langle\nabla f_{t}^j(\x_q^j), \x_q^i-\x \rangle.
   % \leq& \sum_{t=1}^T\sum_{j=1}^n\left(\alpha f_{t}^j(\x)-f_{t}^j(\x_t^j)\right)+\sum_{t=1}^T\sum_{j=1}^nG\|\x_t^j-\x_t^i\|\\
  \end{split}
\end{equation}
Combining \eqref{thm2-DSPF-eq1} with Lemma \ref{lem1-AD-OSPF:label}, for any $\x\in\K$, we have
\begin{equation}
\label{thm2-DSPF-eq2}
\begin{split}
\sum_{q=1}^{T/L}\sum_{t\in\mathcal{T}_q}\left( f_{t}(\x_q^i)-f_{t}(\x)\right)\leq &2\eta nR+\frac{ dnRTLG^2}{2\eta}+\sum_{q=1}^{T/L}\sum_{t\in\mathcal{T}_q}\sum_{j=1}^n\langle \nabla f_{t}^j(\x_q^j),\x_q^i-\bar{\y}_q\rangle
   % \leq& \sum_{t=1}^T\sum_{j=1}^n\left(\alpha f_{t}^j(\x)-f_{t}^j(\x_t^j)\right)+\sum_{t=1}^T\sum_{j=1}^nG\|\x_t^j-\x_t^i\|\\
  \end{split}
\end{equation}
where $\bar{\y}_q$ follows the definition in Lemma \ref{lem1-AD-OSPF:label}.

Note that $f_t^j(\x)$ now is $0$-smooth. By further recalling the definition of $G$ and the oblivious assumption, we can use Lemma \ref{lem2-AD-OSPF:label} and $\langle \nabla f_{t}^j(\x_q^j),\x_q^i-\bar{\y}_q\rangle\leq 2GR$ for any $q\leq2$ to derive that
\begin{equation}
\label{thm2-DSPF-eq3}
\begin{split}
\E\left[\sum_{q=1}^{T/L}\sum_{t\in\mathcal{T}_q}\sum_{j=1}^n\langle \nabla f_{t}^j(\x_q^j),\x_q^i-\bar{\y}_q\rangle\right]\leq\frac{5nT dLRG^2}{\eta}+4nLGR.
  \end{split}
\end{equation}
Finally, combining \eqref{thm2-DSPF-eq2} with \eqref{thm2-DSPF-eq3}, we have
\begin{equation*}
\begin{split}
\E\left[\Reg(T,i)\right]\leq& 2\eta nR+\frac{6 dnRTLG^2}{\eta}+4nLRG\\
\leq& 8n\sqrt{dTC}RG+4n C RG
  \end{split}
\end{equation*}
where the last inequality is due to $\eta=G\sqrt{dTL}$ and $L=C$.

\section{Analysis of Algorithm \ref{ADMFW}}
In this section, we present detailed proofs of the theoretical guarantees for Algorithm \ref{ADMFW}.

 % provide the detailed proofs of theoretical guarantees of Algorithm \ref{ADMFW}.

\subsection{Proof of Theorem \ref{thm1-DMFW}}
We first introduce the following lemma regrading the convergence property of the iterations in \eqref{iterations-MFW}.
\begin{lemma}
\label{lem1-DMFW}
Suppose Assumptions \ref{assum:bounded-set} and \ref{assum:DCsets} hold. Let $\x_1=\ze$ and $\x_{k+1}=\x_{k}+(1/L)\v_k\odot(\mathbf{1}-\x_{k})$, for $k=1,\dots,L$, where $\v_k$ can be any vector in $\K$. Then, it holds that $\x_k\in\K$ for any $k\in[L+1]$. Moreover, for any continuous DR-submodular, non-negative, and $\beta$-smooth function $f(\x):\X\mapsto\mathbb{R}$, and any $\x\in\X$, it holds that
\begin{equation*}
% \label{main-lem1-DMFW}
\begin{split}
f(\x_{L+1})\geq \frac{1}{e}f(\x)+\sum_{k=1}^L\frac{1}{L}\left(1-\frac{1}{L}\right)^{L-k}\langle\nabla f(\x_{k})\odot(\onv-\x_{k}),\v_{k}-\x\rangle-\frac{\beta R^2}{2L}.
\end{split}
\end{equation*}
\end{lemma}
% For each block $q\in[T/L]$, let $F_q(\x)=(1/L)\sum_{t\in\mathcal{T}_q}f_t(\x)$. Under , it is easy to verify that $F_q(\x)$ is continuous DR-submodular, non-negative, and $(n\beta)$-smooth over $\X$. Let $\alpha=1/e$ and $\epsilon=1/L$. Then, 
Due to Lemma \ref{lem1-DMFW},  $f_t(\x)=\sum_{j=1}^nf_t^j(\x)$, and Assumptions \ref{assum:DR-submodular} and \ref{assum:smooth}, for any $\x\in\K$, it is not hard to verify that
\begin{equation}
\label{thm1-DMFW-eq1}
\begin{split}
&\sum_{q=1}^{T/L}\sum_{t\in\mathcal{T}_q}\left(\frac{1}{e} f_t(\x)-f_t(\x_q^i)\right)=\sum_{q=1}^{T/L}\sum_{t\in\mathcal{T}_q}\left(\frac{1}{e} f_t(\x)-f_t(\x_{q,L+1}^i)\right)\\
\leq &\sum_{q=1}^{T/L}\sum_{k=1}^L\frac{1}{L}\left(1-\frac{1}{L}\right)^{L-k}\sum_{t\in\mathcal{T}_q}\sum_{j=1}^n\langle\nabla f_t^j(\x_{q,k}^i)\odot(\x_{q,k}^i-\onv),\v_{q,k}^i-\x\rangle+\frac{n\beta TR^2}{2L}.
\end{split}
\end{equation}
For any $j\in[n]$ and $t\in\mathcal{T}_q$, we have
\begin{equation}
\label{thm1-DMFW-eq2-pre-pre}
\begin{split}
&\|\nabla f_t^j(\x_{q,k}^i)\odot(\x_{q,k}^i-\onv)-\nabla f_t^j(\x_{q,k}^j)\odot(\x_{q,k}^j-\onv)\|\\
\leq&\|(\nabla f_t^j(\x_{q,k}^i)-\nabla f_t^j(\x_{q,k}^j))\odot(\x_{q,k}^i-\onv)\|+\|\nabla f_t^j(\x_{q,k}^j)\odot(\x_{q,k}^j-\x_{q,k}^i)\|.
\end{split}
\end{equation}
Due to $\x_{q,k}^i\in\K$ shown in Lemma \ref{lem1-DMFW}, and $\K\subseteq\X=[0,1]^d$ in Assumption \ref{assum:bounded-set}, it is easy to verify that $\ze\leq \onv-\x_{q,k}^i\leq \onv$, which implies that 
\begin{equation}
\label{thm1-DMFW-eq2-pre}
\|\y\odot(\x_{q,k}^i-\onv)\|\leq\|\y\|,\forall\y\in\mathbb{R}^d.
\end{equation}
Combining  \eqref{thm1-DMFW-eq2-pre-pre} with \eqref{thm1-DMFW-eq2-pre} and \eqref{bound-gradient}, for any $j\in[n]$ and $t\in\mathcal{T}_q$, we have
\begin{equation}
\label{thm1-DMFW-eq2}
\begin{split}
&\|\nabla f_t^j(\x_{q,k}^i)\odot(\x_{q,k}^i-\onv)-\nabla f_t^j(\x_{q,k}^j)\odot(\x_{q,k}^j-\onv)\|\\
\leq &\|\nabla f_t^j(\x_{q,k}^i)-\nabla f_t^j(\x_{q,k}^j)\|+G\|\x_{q,k}^j-\x_{q,k}^i\|\\
\leq& (\beta+G)\|\x_{q,k}^j-\x_{q,k}^i\|
\end{split}
\end{equation}
where the last inequality is due to the $\beta$-smoothness of $f_t^j(\x)$. 

Then, for any $j\in[n]$ and $t\in\mathcal{T}_q$, it is easy to verify that
\begin{equation*}
\begin{split}
&\langle\nabla f_t^j(\x_{q,k}^i)\odot(\x_{q,k}^i-\onv),\v_{q,k}^i-\x\rangle-\langle\nabla f_t^j(\x_{q,k}^j)\odot(\x_{q,k}^j-\onv),\v_{q,k}^i-\x\rangle\\\leq &\|\nabla f_t^j(\x_{q,k}^i)\odot(\x_{q,k}^i-\onv)-\nabla f_t^j(\x_{q,k}^j)\odot(\x_{q,k}^j-\onv)\|\|\v_{q,k}^i-\x\|\\
\leq& 2(\beta+G)R\|\x_{q,k}^j-\x_{q,k}^i\|
\end{split}
\end{equation*}
where the last inequality is due to \eqref{thm1-DMFW-eq2} and $\|\v_{q,k}^i-\x\|\leq 2R$.

For brevity, let $c_k=\left(1-1/L\right)^{L-k}$. Combining \eqref{thm1-DMFW-eq1} with the above inequality, for any $\x\in\K$, we have
\begin{equation}
\label{thm1-DMFW-eq3}
\begin{split}
\sum_{q=1}^{T/L}\sum_{t\in\mathcal{T}_q}\left(\frac{1}{e} f_t(\x)-f_t(\x_q^i)\right)\leq &\sum_{q=1}^{T/L}\sum_{k=1}^L\frac{c_k}{L}\sum_{t\in\mathcal{T}_q}\sum_{j=1}^n\langle\nabla f_t^j(\x_{q,k}^j)\odot(\x_{q,k}^j-\onv),\v_{q,k}^i-\x\rangle\\
&+\frac{2(\beta+G)R}{L}\sum_{q=1}^{T/L}\sum_{k=1}^L\sum_{t\in\mathcal{T}_q}\sum_{j=1}^n\|\x_{q,k}^j-\x_{q,k}^i\|+\frac{n\beta TR^2}{2L}.
\end{split}
\end{equation}
% Moreover, from Step 6 of Algorithm \ref{ADMFW}, conditional  
Next, let $\xi_q$ denote the randomness introduced at each block $q$ of Algorithm \ref{ADMFW}, and $\xi_{1:q}$ denote the randomness introduced at the first $q$ blocks. Note that conditioned on the randomness $\xi_{1:q-1}$, vectors $\{\v_{q,k}^i\}_{k\in[L],i\in[n]}$ and $\{\x_{q,k}^i\}_{k\in[L+1],i\in[n]}$ are determined. Due to the random permutation of $(t_{q,1},\dots, t_{q,L})$  and the oblivious assumption, for any $q\in[T/L]$, $k\in[L]$, and $i\in [n]$, it is not hard to verify that
\begin{equation*}
\begin{split}
\E\left[\tilde{\nabla}f_{t_{q,k}}^i(\x_{q,k}^i)\odot(\x_{q,k}^i-\mathbf{1})|\xi_{1:q-1}\right]=&\E\left[{\nabla}f_{t_{q,k}}^i(\x_{q,k}^i)\odot(\x_{q,k}^i-\mathbf{1})|\xi_{1:q-1}\right]\\
=&\E\left[\left.\frac{1}{L}\sum_{t\in\mathcal{T}_q}{\nabla}f_{t}^i(\x_{q,k}^i)\odot(\x_{q,k}^i-\mathbf{1})\right|\xi_{1:q-1}\right]
\end{split}
\end{equation*}
which implies that
\begin{equation}
\label{thm1-DMFW-eq5}
\begin{split}
&\E\left[\sum_{q=1}^{T/L}\sum_{k=1}^L\frac{c_k}{L}\sum_{t\in\mathcal{T}_q}\sum_{j=1}^n\langle\nabla f_t^j(\x_{q,k}^j)\odot(\x_{q,k}^j-\onv),\v_{q,k}^i-\x\rangle\right]\\
=&\E\left[\sum_{k=1}^Lc_k\sum_{q=1}^{T/L}\sum_{j=1}^n\left(\ell_{q,k}^j(\v_{q,k}^i)-\ell_{q,k}^j(\x)\right)\right].
\end{split}
\end{equation}
Now, we only need to analyze $\|\x_{q,k}^j-\x_{q,k}^i\|$ for any $i,j\in[n]$. For $k=1$, due to $\x_{q,1}^j=\x_{q,1}^i=\ze$, it is easy to verify that 
\begin{equation}
\label{thm1-DMFW-eq6}
\begin{split}
\|\x_{q,1}^j-\x_{q,1}^i\|=0.
 \end{split}
\end{equation}
For any $k>1$, due to the update rule of $\x_{q,k}^j$ and $\x_{q,k}^i$, we have
\begin{equation*}
% \label{thm1-DMFW-eq7}
\begin{split}
&\|\x_{q,k}^j-\x_{q,k}^i\|\\
=&\left\|\x_{q,k-1}^{j}+\frac{\v_{q,k-1}^j}{L} \odot(\mathbf{1}-\x_{q,k-1}^j)-\x_{q,k-1}^i-\frac{\v_{q,k-1}^i}{L} \odot(\mathbf{1}-\x_{q,k-1}^i)\right\|\\
=&\left\|\x_{q,k-1}^{j}\odot\left(\mathbf{1}-\frac{\v_{q,k-1}^j}{L} \right)+\frac{\v_{q,k-1}^j}{L} -\x_{q,k-1}^i\odot\left(\mathbf{1}-\frac{\v_{q,k-1}^i}{L} \right)-\frac{\v_{q,k-1}^i}{L} \right\|\\
\leq &\left\|\x_{q,k-1}^{j}\odot\left(\mathbf{1}-\frac{\v_{q,k-1}^j}{L} \right)-\x_{q,k-1}^i\odot\left(\mathbf{1}-\frac{\v_{q,k-1}^i}{L} \right)\right\|+\frac{\|\v_{q,k-1}^j-\v_{q,k-1}^i\|}{L}\\
\leq &\left\|\x_{q,k-1}^{j}\odot\left(\frac{\v_{q,k-1}^i}{L} -\frac{\v_{q,k-1}^j}{L} \right)\right\|+\left\|(\x_{q,k-1}^{j}-\x_{q,k-1}^{i})\odot\left(\mathbf{1}-\frac{\v_{q,k-1}^i}{L} \right)\right\|+\frac{\|\v_{q,k-1}^j-\v_{q,k-1}^i\|}{L}.
% \leq &\frac{\|\v_{q,k-1}^i-\v_{q,k-1}^j\|}{L}+\|\x_{q,k-1}^{j}-\x_{q,k-1}^{i}\|
 \end{split}
\end{equation*}
Due to $\x_{q,k-1}^{j},\v_{q,k-1}^i\in\K$ and $\K\subseteq\X=[0,1]^d$ in Assumption \ref{assum:bounded-set}, we can verify that $\|\x_{q,k-1}^{j}\odot\y\|\leq \|\y\|$ and $\|\y\odot(1-\v_{q,k-1}^i/L)\|\leq \|\y\|$, $\forall\y\in\mathbb{R}^d$. Therefore, for any $k>1$, the above inequality can be further relaxed to
\begin{equation}
\label{thm1-DMFW-new-eq7}
\begin{split}
\|\x_{q,k}^j-\x_{q,k}^i\| \leq \frac{2\|\v_{q,k-1}^i-\v_{q,k-1}^j\|}{L}+\|\x_{q,k-1}^{j}-\x_{q,k-1}^{i}\|
 \end{split}
\end{equation}
% \begin{equation*}
% % \label{thm1-DMFW-eq7}
% \begin{split}
% &\left\|\x_{q,k-1}^{j}\odot\left(\mathbf{1}-\frac{\v_{q,k-1}^j}{L} \right)-\x_{q,k-1}^i\odot\left(\mathbf{1}-\frac{\v_{q,k-1}^i}{L} \right)\right\|\\
% \leq &\left\|\x_{q,k-1}^{j}\odot\left(\frac{\v_{q,k-1}^i}{L} -\frac{\v_{q,k-1}^j}{L} \right)\right\|+\left\|(\x_{q,k-1}^{j}-\x_{q,k-1}^{i})\odot\left(\mathbf{1}-\frac{\v_{q,k-1}^i}{L} \right)\right\|\\
% \leq &\frac{\|\v_{q,k-1}^i-\v_{q,k-1}^j\|}{L}+\|\x_{q,k-1}^{j}-\x_{q,k-1}^{i}\|
%  \end{split}
% \end{equation*}
From \eqref{thm1-DMFW-eq6} and \eqref{thm1-DMFW-new-eq7}, for any $k\in[L]$, it is easy to verify that
\begin{equation}
\label{thm1-DMFW-eq7}
\begin{split}
\|\x_{q,k}^j-\x_{q,k}^i\|\leq \frac{2}{L}\sum_{\tau=1}^{k-1}\|\v_{q,\tau}^j-\v_{q,\tau}^i\|.
 \end{split}
\end{equation}
Finally, combining \eqref{thm1-DMFW-eq3} with \eqref{thm1-DMFW-eq5}, \eqref{thm1-DMFW-eq7}, and the definition of $c_k$, for any $\x\in\K$, we have
\begin{equation*}
\begin{split}
\E\left[\sum_{q=1}^{T/L}\sum_{t\in\mathcal{T}_q}\left(\frac{f_t(\x)}{e} -f_t(\x_q^i)\right)\right]\leq& \E\left[\sum_{k=1}^L\left(1-\frac{1}{L}\right)^{L-k}\sum_{q=1}^{T/L}\sum_{j=1}^n\left(\ell_{q,k}^j(\v_{q,k}^i)-\ell_{q,k}^j(\x)\right)\right]\\
&+\E\left[\frac{4(\beta+G)R}{L}\sum_{k=1}^L\sum_{q=1}^{T/L}\sum_{j=1}^n\sum_{\tau=1}^{k-1}\|\v_{q,\tau}^j-\v_{q,\tau}^i\|\right]+\frac{n\beta TR^2}{2L}\\
\leq& \E\left[\sum_{k=1}^L\left(1-\frac{1}{L}\right)^{L-k}\sum_{q=1}^{T/L}\sum_{j=1}^n\left(\ell_{q,k}^j(\v_{q,k}^i)-\ell_{q,k}^j(\x)\right)\right]\\
&+\E\left[4(\beta+G)R\sum_{k=1}^L\sum_{q=1}^{T/L}\sum_{j=1}^n\|\v_{q,k}^j-\v_{q,k}^i\|\right]+\frac{n\beta TR^2}{2L}.
\end{split}
\end{equation*}

\subsection{Proof of Lemma \ref{lem1-DMFW}}
This lemma is a simplified version of Lemma 8 of \citet{pedramfar2024unified}, and can be proved by combining some existing results in \citet{Zhang-AISTATS23}. Specifically, under Assumptions \ref{assum:bounded-set} and \ref{assum:DCsets}, the result that $\x_k\in\K$ for any $k\in[L+1]$ is due to Lemma 4 of \citet{Zhang-AISTATS23}. Moreover, for any $\x\in\X$, and $k=2,3,\dots,L+1$, Lemma 7 of \citet{Zhang-AISTATS23} has shown that 
\begin{equation*}
\begin{split}
f(\x_{k})\geq& \left(1-\frac{1}{L}\right)f(\x_{k-1})+\frac{1}{L}f(\x_{k-1}+(\onv-\x_{k-1})\odot\x)\\
&+\frac{1}{L}\langle(\v_{k-1}-\x)\odot(\onv-\x_{k-1}),\nabla f(\x_{k-1})\rangle-\frac{\beta}{2}\|\x_{k}-\x_{k-1}\|^2.
\end{split}
\end{equation*}
Therefore, for any $\x\in\X$, it is not hard to verify that
\begin{equation}
\label{lem1-DMFW-eq1}
\begin{split}
f(\x_{L+1})\geq& \left(1-\frac{1}{L}\right)f(\x_{1})+\sum_{k=1}^L\frac{1}{L}\left(1-\frac{1}{L}\right)^{L-k}f(\x_{k}+(\onv-\x_{k})\odot\x)\\
&+\sum_{k=1}^L\frac{1}{L}\left(1-\frac{1}{L}\right)^{L-k}\langle(\v_{k}-\x)\odot(\onv-\x_{k}),\nabla f(\x_{k})\rangle\\
&-\sum_{k=1}^L\left(1-\frac{1}{L}\right)^{L-k}\frac{\beta}{2}\|\x_{k+1}-\x_{k}\|^2.
\end{split}
\end{equation}
To simplify the right side of \eqref{lem1-DMFW-eq1}, we notice that $f(\x_1)\geq 0$ and 
\begin{equation}
\label{lem1-DMFW-eq2}
\|\x_{k+1}-\x_{k}\|^2=\left\|\frac{1}{L}\v_k\odot(\mathbf{1}-\x_{k})\right\|^2\leq \left\|\frac{1}{L}\v_k\right \|^2\leq \frac{R^2}{L^2}
\end{equation}
where the first inequality is due to $\x_k\in\K\subseteq\X=[0,1]^n$ and the last inequality is due to $\|\v_k\|\leq R$. Additionally, we introduce the following lemma.
\begin{lemma}
\label{lem6:zhang-AISTATS23}
(Lemma 6 of \citet{Zhang-AISTATS23}) For any continuous DR-submodular function $f(\x):[0,1]^d\mapsto\mathbb{R}_{+}$, and any $\x,\y\in[0,1]^d$, we have
\[
  f(\y+(1-\y)\odot\x)\geq (1-\|\y\|_\infty)f(\x).
\]
\end{lemma}
Combining \eqref{lem1-DMFW-eq1} with $f(\x_1)\geq 0$, \eqref{lem1-DMFW-eq2}, and Lemma \ref{lem6:zhang-AISTATS23}, for any $\x\in\X$, we have
\begin{equation}
\label{lem1-DMFW-eq3}
\begin{split}
f(\x_{L+1})\geq& \sum_{k=1}^L\frac{1}{L}\left(1-\frac{1}{L}\right)^{L-k}(1-\|\x_{k}\|_\infty)f(\x)\\
&+\sum_{k=1}^L\frac{1}{L}\left(1-\frac{1}{L}\right)^{L-k}\langle(\v_{k}-\x)\odot(\onv-\x_{k}),\nabla f(\x_{k})\rangle-\frac{\beta R^2}{2L}.
\end{split}
\end{equation}
Note that Lemma 5 of \citet{Zhang-AISTATS23} has shown that $\x_k(i)\leq 1-\left(1-1/L\right)^{k-1}$ for any $i\in[d]$ and $k\in[L]$, where $\x_k(i)$ denotes the $i$-th element in $\x_k$. Then, due to $f(\x)\geq 0$, it is easy to verify that
\begin{equation}
\label{lem1-DMFW-eq4}
\begin{split}
\sum_{k=1}^L\frac{1}{L}\left(1-\frac{1}{L}\right)^{L-k}(1-\|\x_{k}\|_\infty)f(\x)\geq\left(1-\frac{1}{L}\right)^{L-1}f(\x)\geq \frac{1}{e}f(\x).
\end{split}
\end{equation}
Finally, combining \eqref{lem1-DMFW-eq3} with \eqref{lem1-DMFW-eq4} and $\langle(\v_{k}-\x)\odot(\onv-\x_{k}),\nabla f(\x_{k})\rangle=\langle\nabla f(\x_{k})\odot(\onv-\x_{k}),\v_{k}-\x\rangle$, we have
\begin{equation*}
\begin{split}
f(\x_{L+1})\geq \frac{1}{e}f(\x)+\sum_{k=1}^L\frac{1}{L}\left(1-\frac{1}{L}\right)^{L-k}\langle\nabla f(\x_{k})\odot(\onv-\x_{k}),\v_{k}-\x\rangle-\frac{\beta R^2}{2L}.
\end{split}
\end{equation*}

\subsection{Proof of Corollary \ref{cor5}}
Due to \eqref{thm1-DMFW-eq2-pre} and Assumption \ref{assumption3}, it is easy to verify that
\begin{equation}
\label{bound-estimated-stochastic-DMFW}
\left\|\tilde{\nabla}f_{t}^i(\x_{q,k}^i)\odot(\x_{q,k}^i-\mathbf{1})\right\|\leq \left\|\tilde{\nabla}f_{t}^i(\x_{q,k}^i)\right\|\leq G.
\end{equation}
Note that as previously discussed, for any $k\in[L]$, linear losses $\{\ell_{q,k}^i(\x)\}_{q\in[T/L],i\in[n]}$ in Algorithm \ref{ADMFW} actually are oblivious to the $k$-th instance of the D-OCO algorithm $\D$, i.e., $\D_k$. 

Combining this fact with \eqref{bound-estimated-stochastic-DMFW}, and Assumptions \ref{assum:bounded-set} and \ref{assumption4}, for any $\x\in\K$, we can directly apply Theorem \ref{thm-DFTPL} to derive the following upper bound
\begin{equation}
\label{cor5-eq1}
\begin{split}
\sum_{q=1}^{T/L}\sum_{j=1}^n\left(\ell_{q,k}^j(\v_{q,k}^i)-\ell_{q,k}^j(\x)\right)\leq& 5nRG\sqrt{\frac{dTC^\prime}{L}}+4nC^\prime GR+6nGR\\
\leq& 15nRG\sqrt{\frac{dTC^\prime}{L}}
%\frac{27nKTG^2}{2hL}+nhR^2+\frac{9nGRT}{2L\sqrt{K+2}}
%\leq \frac{2nLTG^2}{h}+nhR^2
\end{split}
\end{equation}
where $C^\prime=2\lceil \ln(\sqrt{n}T/L)\rho^{-1}\rceil$.
% where $C=\lceil 2\ln(T/L\sqrt{n})/(1-\sigma_2(A))\rceil\leq T/L$.

Moreover, from Lemma \ref{lem1:D-FTPL}, it is also easy to derive the following consensus error bound
\begin{equation}
\label{cor5-eq2}
\sum_{k=1}^L\sum_{q=1}^{T/L}\sum_{j=1}^n\|\v_{q,k}^j-\v_{q,k}^i\|\leq 4nRL.
\end{equation}
By substituting \eqref{cor5-eq1} and \eqref{cor5-eq2} into Theorem \ref{thm1-DMFW}, for any $\x\in\K$, we have
\begin{equation*}
% \label{cor5-eq3}
\begin{split}
\E\left[\sum_{q=1}^{T/L}\sum_{t\in\mathcal{T}_q}\left(\alpha f_t(\x)-f_t(\x_q^i)\right)\right]
\leq& 15nRG\sqrt{dTC^\prime L}+16(\beta+G)nR^2L+\frac{n\beta TR^2}{2L}.
\end{split}
\end{equation*}

\section{Analysis of Algorithm \ref{D-FTPL}}
In this section, we present detailed proofs of the theoretical guarantees for Algorithm \ref{D-FTPL}.

% provide the detailed proofs of theoretical guarantees of Algorithm \ref{D-FTPL}.

\subsection{Proof of Lemma \ref{lem1:D-FTPL}}
This lemma can be simply proved by using the consensus error bound of the standard gossip step \citep{Xiao-Gossip04}. Here, we provide the detailed proof for completeness. 

Specifically, let $X_k=[(\g_{q,k}^1)^\top;(\g_{q,k}^2)^\top;\dots;(\g_{q,k}^n)^\top]\in\mathbb{R}^{n\times d}$ for any $k=1,2,\dots,L/2+1$, and $\bar{X}=[\bar{\g}_q^\top;\bar{\g}_q^\top;,\dots;\bar{\g}_q^\top]\in\mathbb{R}^{n\times d}$. From step 9 of Algorithm \ref{D-FTPL}, for any $k\in[L/2]$, it is easy to verify that $X_{k+1}=AX_{k}$. Note that according to Assumption \ref{assumption4}, the matrix $A$ is doubly stochastic. Therefore, for any $k\in[L/2]$, we further have
\[
  \frac{\onv\onv^\top}{n}X_{k+1}=\frac{\onv\onv^\top}{n}AX_{k}=\frac{\onv\onv^\top}{n}X_{k}=\dots=\frac{\onv\onv^\top}{n}X_{1}=\bar{X}.
\]
Combining with $\bar{X}=A\bar{X}$, for any $k\in[L/2]$, we have
\begin{align*}
\|X_{k+1}-\bar{X}\|_F=&\|A(X_{k}-\bar{X})\|_F=\left\|\left(A-\frac{\onv\onv^\top}{n}\right)(X_k-\bar{X})\right\|_F\\
\leq&\left\|A-\frac{\onv\onv^\top}{n}\right\|_2\left\|X_k-\bar{X}\right\|_F=\sigma_2(A)\left\|X_k-\bar{X}\right\|_F
\end{align*}
where the last equality is also due to the doubly stochastic property of $A$.

Combining the above inequality with $\g_{q}^i=\g_{q,L/2+1}^i$, for any $i\in[n]$ and $q\in[T/L]$, we can derive that
\begin{equation}
\label{lem1-DFTPL-eq1}
\|\g_{q}^i-\bar{\g}_q\|\leq \|X_{L/2+1}-\bar{X}\|_F\leq(\sigma_2(A))^{L/2}\left\|X_1-\bar{X}\right\|_F.
\end{equation}
Due to the value of $L$, it is easy to verify that 
\begin{equation}
\label{lem1-DFTPL-eq2}
(\sigma_2(A))^{L/2}\leq  (\sigma_2(A))^{\frac{\ln(T\sqrt{n})}{1-\sigma_2(A)}}\leq (\sigma_2(A))^{\frac{\ln(T\sqrt{n})}{\ln(1/\sigma_2(A))}}=\frac{1}{T\sqrt{n}}
\end{equation}
where the second inequality is due to $\ln(1/x)\geq 1-x$ for any $x\geq 0$. Moreover, we have $\|X_1-\bar{X}\|_F\leq\|X_1\|_F+\|\bar{X}\|_F$ and thus
\begin{equation}
\label{lem1-DFTPL-eq3}
\left\|X_1-\bar{X}\right\|_F\leq 2\sqrt{\sum_{i=1}^n\|\g_{q,1}^i\|^2}\\
\leq 2\sqrt{L\sum_{i=1}^n\sum_{t\in\mathcal{T}_q}\|\nabla f_{t}^i(\x_q^i)\|^2}\leq2\sqrt{n}LG
\end{equation}
where the last inequality is due to the definition of $G$.

By substituting \eqref{lem1-DFTPL-eq2} and \eqref{lem1-DFTPL-eq3} into \eqref{lem1-DFTPL-eq1}, for any $i\in[n]$ and $q\in[T/L]$, we have $\|\g_{q}^i-\bar{\g}_q\|\leq 2LG/T$. Next, by repeating the above processes with $\x_q^i$ and noticing that $\x_1^i=\hat{\x}_1^i$, for any $i\in[n]$ and $q\in[T/L]$, it is easy to verify that
\begin{equation*}
\|\x_{q}^i-\bar{\x}_q\|\leq \frac{2\sqrt{\sum_{i=1}^n\|\hat{\x}_q^i\|^2}}{T\sqrt{n}}\leq \frac{2R}{T}.
\end{equation*}
\subsection{Proof of Theorem \ref{thm-DFTPL}}
In Lemma \ref{lem1:D-FTPL}, we have defined $\bar{\x}_q=(1/n)\sum_{i=1}^n\hat{\x}_q^i$  and $\bar{\g}_q=(1/n)\sum_{i=1}^n\sum_{t\in\mathcal{T}_q}\nabla f_t^i(\x_q^i)$, where $\bar{\g}_q$ now can be rewritten as $\bar{\g}_q=(1/n)\sum_{i=1}^n\sum_{t\in\mathcal{T}_q}\c_t^i$ due to $f_t^i(\x)=\langle\c_t^i,\x\rangle$. For any $\x\in\K$, due to $f_t(\x)=\sum_{j=1}^nf_t^j(\x)=\sum_{j=1}^n\langle\c_t^j,\x\rangle$, it is not hard to verify that
\begin{equation}
\label{thm1-DFTPL-eq1}
\begin{split}
&\sum_{q=1}^{T/L}\sum_{t\in\mathcal{T}_q}\left(f_t(\x_q^i)-f_t(\x)\right)=\sum_{q=1}^{T/L}\sum_{t\in\mathcal{T}_q}\sum_{j=1}^n\langle\c_t^j,\x_q^i-\x\rangle\\
\leq &\sum_{q=1}^{T/L}\sum_{t\in\mathcal{T}_q}\sum_{j=1}^n\langle\c_t^j,\bar{\x}_q-\x\rangle+\sum_{q=1}^{T/L}\sum_{t\in\mathcal{T}_q}\sum_{j=1}^n\|\c_t^j\|\|\x_q^i-\bar{\x}_q\|\\
=&\sum_{q=1}^{T/L}n\langle\bar{\g}_q,\bar{\x}_q-\x\rangle+\sum_{q=1}^{T/L}\sum_{t\in\mathcal{T}_q}\sum_{j=1}^n\|\c_t^j\|\|\x_q^i-\bar{\x}_q\|\\
=&\sum_{q=1}^{T/L}\sum_{j=1}^n\langle\bar{\g}_q,\hat{\x}_q^j-\x\rangle+\sum_{q=1}^{T/L}\sum_{t\in\mathcal{T}_q}\sum_{j=1}^n\|\c_t^j\|\|\x_q^i-\bar{\x}_q\|\\
\leq &\sum_{q=1}^{T/L}\sum_{j=1}^n\langle \g_q^j,\hat{\x}_q^j-\x\rangle+\sum_{q=1}^{T/L}\sum_{j=1}^n\|\bar{\g}_q-\g_q^j\|\|\hat{\x}_q^j-\x\|+\sum_{q=1}^{T/L}\sum_{t\in\mathcal{T}_q}\sum_{j=1}^n\|\c_t^j\|\|\x_q^i-\bar{\x}_q\|\\
\leq &\sum_{q=1}^{T/L}\sum_{j=1}^n\langle \g_q^j,\hat{\x}_q^j-\x\rangle+6nGR
\end{split}
\end{equation}
where the last inequality is due to Lemma \ref{lem1:D-FTPL}, $\|\hat{\x}_q^j-\x\|\leq 2R$, and $\|\c_t^j\|\leq G$.

Then, let $\y_q^j=\E_{\v\in\B}[\argmax_{\x\in\K}\langle-\sum_{\tau=1}^{q-1}\g_{\tau}^j+\eta\v,\x\rangle]$ for any $j\in[n]$ and $q\in[T/L]$. Note that since the adversary is oblivious, for any $t\in[T]$ and $i\in[n]$, $\c_t^i$ is independent of any randomness of Algorithm \ref{D-FTPL}. Due to the computation of $\g_q^i$ is determined for any $q\in[T/L]$ and $i\in[n]$, it is also independent of any randomness of Algorithm \ref{D-FTPL}. This implies that for any $j\in[n]$ and $q\geq 3$, $\hat{\x}_q^j$ is an unbiased estimation of $\y_{q-2}^{j}$. 

Therefore, for any $\x\in\K$, we have
\begin{equation}
\label{thm1-DFTPL-eq3}
\begin{split}
&\E\left[\sum_{q=1}^{T/L}\langle \g_q^j,\hat{\x}_q^j-\x\rangle\right]=\E\left[\sum_{q=3}^{T/L}\langle \g_q^j,\y_{q-2}^j-\x\rangle\right]+\E\left[\sum_{q=1}^{2}\langle \g_q^j,\hat{\x}_q^j-\x\rangle\right]\\
=&\E\left[\sum_{q=1}^{T/L}\langle \g_q^j,\y_{q}^j-\x\rangle\right]+\E\left[\sum_{q=3}^{T/L}\langle \g_q^j,\y_{q-2}^j-\y_{q}^j\rangle\right]+\E\left[\sum_{q=1}^{2}\langle \g_q^j,\hat{\x}_q^j-\y_{q}^j\rangle\right].
% \leq & \frac{2R}{\eta}+\frac{\eta dRTLG^2}{2}+\E\left[\sum_{q=3}^{T/L}\langle \g_q^j,\y_{q-2}^j-\y_{q}^j\rangle\right]+\E\left[\sum_{q=1}^{2}\langle \g_q^j,\hat{\x}_q^j-\y_{q}^j\rangle\right]
\end{split}
\end{equation}
Since the standard gossip step can be viewed as a convex combination, for any $j\in[n]$ and $q\in[T/L]$, it is easy to verify that 
\begin{equation}
\label{thm1-DFTPL-eq5-pre}
\|\g_q^j\|\leq \max_{i\in[n]}\|\g_{q,1}^i\|\leq \max_{i\in[n]}\sum_{t\in\mathcal{T}_q}\left\|\c_t^i\right \|\leq LG.
\end{equation}
Combining \eqref{thm1-DFTPL-eq3} with \eqref{thm1-DFTPL-eq5-pre} and Lemma \ref{lem-opfs-hazan}, for any $\x\in\K$,  we have
\begin{equation}
\label{thm1-DFTPL-eq5}
\begin{split}
\E\left[\sum_{q=1}^{T/L}\langle \g_q^j,\hat{\x}_q^j-\x\rangle\right]\leq & {2\eta R}+\frac{ dRTLG^2}{2\eta}+\E\left[\sum_{q=3}^{T/L}LG\|\y_{q-2}^j-\y_{q}^j\|\right]+4LGR\\
\leq &  {2\eta R}+\frac{ dRTLG^2}{2\eta}+\E\left[\frac{2dTRG}{\eta}\max_{q\in[T/L]}\|\g_{q}^j\|\right]+4LGR\\
\leq &{2\eta R}+\frac{3 dRTLG^2}{\eta}+4LGR
\end{split}
\end{equation}
where the second inequality is due to Lemma \ref{lem1_ospf}.

Finally, combining \eqref{thm1-DFTPL-eq1} with \eqref{thm1-DFTPL-eq5} and $\eta=G\sqrt{dTL}$, for any $\x\in\K$, we have
\begin{align*}
  \E\left[\sum_{q=1}^{T/L}\sum_{t\in\mathcal{T}_q}\left(f_t(\x_q^i)-f_t(\x)\right)\right]\leq& 2\eta nR+\frac{3n dRTLG^2}{\eta}+4nLGR+6nGR\\
  \leq& 5nRG\sqrt{dTL}+4nLGR+6nGR.
\end{align*}
% where the last inequality is due to .
%and $L=\lceil 2\ln(T\sqrt{n})/(1-\sigma_2(A))\rceil$.

\section{Analysis of Algorithm \ref{mono-DNSCR}}
The theoretical guarantees for Algorithm \ref{mono-DNSCR} can be proved by following the analysis of Algorithm \ref{DNSCR}. In this section, we present detailed proofs for completeness. 

\subsection{Proof of Theorem \ref{thm: mono-DSCR}}
We first notice that \eqref{thm1-eq1} in the proof of Theorem \ref{thm: DSCR} still holds, though $\alpha=1-1/e$ and $\x_t^i=\hat{\x}_t^i$ here. Combining \eqref{thm1-eq1} with Lemma \ref{lem:non-oblvious-mon}, for any $\x\in\K$, we have
\begin{equation}
\label{mono-thm1-eq2}
\begin{split}
\alpha\sum_{t=1}^Tf_{t}(\x)-\sum_{t=1}^Tf_{t}(\x_t^i)  {\leq}& \sum_{t=1}^T\sum_{j=1}^n\left\langle\nabla F_t^j(\hat{\x}_t^j),\x-\hat{\x}_t^j\right\rangle+\sum_{t=1}^T\sum_{j=1}^nG\|\hat{\x}_t^j-\hat{\x}_t^i\|\\
  \leq &\sum_{t=1}^T\sum_{j=1}^n\left(\left\langle\nabla F_t^j(\hat{\x}_t^j),\x-\hat{\x}_t^i\right\rangle+\left(\left\|\nabla F_t^j(\hat{\x}_t^j)\right\|+G\right)\|\hat{\x}_t^j-\hat{\x}_t^i\|\right).
  % \leq &\sum_{t=1}^T\sum_{j=1}^n\left\langle\nabla F_t^j(\hat{\x}_t^j),\x-\hat{\x}_t^i\right\rangle+\sum_{t=1}^T\sum_{j=1}^nG\|\hat{\x}_t^j-\hat{\x}_t^i\|
  %+\sum_{t=1}^T\sum_{j=1}^n\frac{G}{2}\|\hat{\x}_t^j-\hat{\x}_t^i\|.
  % \leq &\sum_{t=1}^T\sum_{j=1}^n\left\langle\nabla F_t^j(\hat{\x}_t^j),\x-\hat{\x}_t^i\right\rangle+\sum_{t=1}^T\sum_{j=1}^n\left\|\nabla F_t^j(\hat{\x}_t^j)\right\|\|\hat{\x}_t^j-\hat{\x}_t^i\|+\sum_{t=1}^T\sum_{j=1}^nG\|\x_t^j-\x_t^i\|\\
   % \leq& \sum_{t=1}^T\sum_{j=1}^n\left(\alpha f_{t}^j(\x)-f_{t}^j(\x_t^j)\right)+\sum_{t=1}^T\sum_{j=1}^nG\|\x_t^j-\x_t^i\|\\
  \end{split}
\end{equation}
Moreover, from the definition of $\nabla F_t^i(\x)$ given in Lemma \ref{lem:non-oblvious-mon}, and Assumption \ref{assum:bounded-set}, for any $\x\in \mathcal{K}$, it is not hard to verify that 
\begin{equation}
\label{mono-thm1-eq3}
\nabla F_t^i(\x)= \mathbb{E}_{z_t^i\sim\Z^\prime}\left.\left[\alpha \tilde{\nabla} f_t^i(z_t^i\x)\right|\x\right].
\end{equation}
Combining \eqref{mono-thm1-eq3} with Jensen's inequality and Assumption \ref{assumption3}, for any $\x\in \mathcal{K}$, we have
\begin{equation}
\label{mono-thm1-eq4}
\left\|\nabla F_t^i(\x)\right\|\leq \mathbb{E}_{z_t^i\sim\Z^\prime}\left.\left[\left\|\alpha \tilde{\nabla} f_t^i(z_t^i\x)\right\|\right|\x\right]\leq \alpha G.
\end{equation}
% Due to \eqref{definition: non-monotone gradient F}, an unbiased estimation of $\nabla F_t^i(\x)$ can be computed by sampling $z_t^i$ from $Z$ and setting $\tilde{\nabla} F_t^i(\x)=\frac{3}{8}\tilde{\nabla} f_t^i(\x^\prime)$, where $\x^\prime=(z_t^i/2)(\x-\xx)+\xx$. 
Then, combining \eqref{mono-thm1-eq2} with \eqref{mono-thm1-eq4}, for any $\x\in \mathcal{K}$, we have
\begin{equation}
\label{mono-thm1-eq5}
\begin{split}
\alpha\sum_{t=1}^Tf_{t}(\x)-\sum_{t=1}^Tf_{t}(\x_t^i)  
  \leq \sum_{t=1}^T\sum_{j=1}^n\left\langle\nabla F_t^j(\hat{\x}_t^j),\x-\hat{\x}_t^i\right\rangle+\sum_{t=1}^T\sum_{j=1}^n2G\|\hat{\x}_t^j-\hat{\x}_t^i\|.
  % \leq &\sum_{t=1}^T\sum_{j=1}^n\left\langle\nabla F_t^j(\hat{\x}_t^j),\x-\hat{\x}_t^i\right\rangle+\sum_{t=1}^T\sum_{j=1}^nG\|\hat{\x}_t^j-\hat{\x}_t^i\|
  %+\sum_{t=1}^T\sum_{j=1}^n\frac{G}{2}\|\hat{\x}_t^j-\hat{\x}_t^i\|.
  % \leq &\sum_{t=1}^T\sum_{j=1}^n\left\langle\nabla F_t^j(\hat{\x}_t^j),\x-\hat{\x}_t^i\right\rangle+\sum_{t=1}^T\sum_{j=1}^n\left\|\nabla F_t^j(\hat{\x}_t^j)\right\|\|\hat{\x}_t^j-\hat{\x}_t^i\|+\sum_{t=1}^T\sum_{j=1}^nG\|\x_t^j-\x_t^i\|\\
   % \leq& \sum_{t=1}^T\sum_{j=1}^n\left(\alpha f_{t}^j(\x)-f_{t}^j(\x_t^j)\right)+\sum_{t=1}^T\sum_{j=1}^nG\|\x_t^j-\x_t^i\|\\
  \end{split}
\end{equation}
Finally, due to \eqref{mono-thm1-eq3}, \eqref{mono-thm1-eq5}, and $\ell_t^i(\x)=\langle-\tilde{\nabla} {F}_t^i(\hat{\x}_t^i), \x\rangle$, for any $\x\in \mathcal{K}$, we have
\begin{equation*}
\begin{split}
\E\left[\alpha\sum_{t=1}^Tf_{t}(\x)-\sum_{t=1}^Tf_{t}(\x_t^i) \right] 
  \leq \E\left[\sum_{t=1}^T\sum_{j=1}^n\left(\ell_t^j(\hat{\x}_t^i)-\ell_t^j(\x)\right)\right]+\E\left[\sum_{t=1}^T\sum_{j=1}^n2G\|\hat{\x}_t^j-\hat{\x}_t^i\|\right].
  \end{split}
\end{equation*}

\subsection{Proof of Corollary \ref{mono-cor1:label}}
Due to Assumption \ref{assumption3} and $\alpha=1-1/e$, it is easy to verify that
\begin{equation}
\label{mono-bound-estimated-stochastic}
\left\|\tilde{\nabla} {F}_t^i(\hat{\x}_t^i)\right\|= \alpha \left\|\tilde{\nabla} f_t^i(z_t^i\hat{\x}_t^i)\right\|\leq \alpha G
\end{equation}
which implies that the linear loss $\ell_t^i(\x)$ is $(\alpha G)$-Lipschitz over $\K$. Because of this Lipschitz continuity, and Assumptions \ref{assum:bounded-set} and \ref{assumption4}, for any $\x\in\K$, we can directly use Theorem 1 of \citet{wan2024optimal} to derive the following upper bound
\begin{equation}
\label{mono-cor1-eq1}
\sum_{t=1}^T\sum_{j=1}^n\left(\ell_t^j(\hat{\x}_t^i)-\ell_t^j(\x)\right)\leq \frac{21n\alpha^2CTG^2}{2h}+nhR^2
%\leq \frac{2nLTG^2}{h}+nhR^2
\end{equation}
where $C=\lceil\sqrt{2}\ln(\sqrt{14n})/((\sqrt{2}-1)\sqrt{\rho})\rceil$, and $h$ can be any positive constant.

Moreover, from (39), (40), and (41) in the proof of Theorem 1 of \citet{wan2024optimal}, it is also easy to derive the following consensus error bound
\begin{equation}
\label{mono-cor1-eq2}
\sum_{t=1}^T\sum_{j=1}^n\|\hat{\x}_t^j-\hat{\x}_t^i\|\leq \frac{7n\alpha CTG}{h}.
\end{equation}
By substituting \eqref{mono-cor1-eq1} and \eqref{mono-cor1-eq2} into Theorem \ref{mono-DNSCR}, for any $\x\in\K$, we have
\begin{equation}
\label{mono-cor1-eq3}
  \E\left[\alpha\sum_{t=1}^Tf_{t}(\x)-\sum_{t=1}^Tf_{t}(\x_t^i) \right] 
  \leq \frac{25n CTG^2}{h}+nhR^2.
\end{equation}
Finally, we can complete this proof by substituting $h=5\sqrt{ CT}G/R$ into \eqref{mono-cor1-eq3}.

\subsection{Proof of Corollary \ref{mono-cor2:label}}
Similar to \eqref{mono-cor1-eq1}, we can directly use Theorem 5 of \citet{wan2024optimal} to derive the following upper bound
\begin{equation}
\label{mono-cor2-eq1}
\sum_{t=1}^T\sum_{j=1}^n\left(\ell_t^j(\hat{\x}_t^i)-\ell_t^j(\x)\right)\leq \frac{27n\alpha^2 LTG^2}{2h}+nhR^2+\frac{12n\alpha GRT}{\sqrt{L+2}}
%\leq \frac{2nLTG^2}{h}+nhR^2
\end{equation}
where $L$ is an integer such that $T\geq L\geq C=\lceil\sqrt{2}\ln(\sqrt{14n})/((\sqrt{2}-1)\sqrt{\rho})\rceil$, and $h$ can be any positive constant.

Moreover, from the proof of Theorem 5 of \citet{wan2024optimal}, it is also easy to derive the following consensus error bound
\begin{equation}
\label{mono-cor2-eq2}
\sum_{t=1}^T\sum_{j=1}^n\|\hat{\x}_t^j-\hat{\x}_t^i\|\leq \frac{9n\alpha LTG}{h}+\frac{8nRT}{\sqrt{L+2}}.
\end{equation}
By substituting \eqref{mono-cor2-eq1} and \eqref{mono-cor2-eq2} into Theorem \ref{mono-DNSCR}, for any $\x\in\K$, we have
\begin{equation}
\label{mono-cor2-eq3}
\begin{split}
  \E\left[\alpha\sum_{t=1}^Tf_{t}(\x)-\sum_{t=1}^Tf_{t}(\x_t^i) \right] 
  \leq& \frac{32n LTG^2}{h}+nhR^2+\frac{28nGRT}{\sqrt{L}}.
  % \leq& \frac{14n LTG^2}{h}+nhR^2+\frac{12nGRT}{\sqrt{L}}.
\end{split}
\end{equation}
Finally, by substituting $h=4\sqrt{2LT}G/R$ and $\sqrt{T}\leq L=\max\{\lceil\sqrt{T}\rceil,C\}\leq2\sqrt{T}+C$ into \eqref{mono-cor2-eq3}, we have
\[
  \E\left[\Reg(T,i,\alpha) \right]\leq 8nGR\sqrt{2TC}+44nGRT^{3/4}.
\]

\subsection{Proof of Theorem \ref{thm: mono-DSCR-smooth}}
Note that \eqref{thm1-smooth-eq0} in the proof of Theorem \ref{thm: DSCR-smooth} still holds, though $\alpha=1-1/e$ and $\x_t^i=\hat{\x}_t^i$ here. Combining \eqref{thm1-smooth-eq0} with Lemma \ref{lem:non-oblvious-mon}, for any $\x\in\K$, we have
\begin{equation} 
\label{mono-thm1-smooth-eq1}
\begin{split}
&\alpha\sum_{t=1}^Tf_{t}(\x)-\sum_{t=1}^Tf_{t}(\x_t^i)  \\
\leq & \sum_{t=1}^T\sum_{j=1}^n\left(\left\langle\nabla F_t^j(\hat{\x}_t^j),\x-\hat{\x}_t^j\right\rangle+\langle\nabla f_t^j(\hat{\x}_t^j),\hat{\x}_t^j-\hat{\x}_t^i\rangle+\frac{\beta}{2}\|\hat{\x}_t^i-\hat{\x}_t^j\|^2\right)\\
\leq & \sum_{t=1}^T\sum_{j=1}^n\left(\left\langle\nabla F_t^j(\hat{\x}_t^j),\x-\hat{\x}_t^i\right\rangle+\left\langle\nabla f_t^j(\hat{\x}_t^j)-\nabla F_t^j(\hat{\x}_t^j),\hat{\x}_t^j-\hat{\x}_t^i\right\rangle+\frac{\beta}{2}\|\hat{\x}_t^i-\hat{\x}_t^j\|^2\right).
  \end{split}
\end{equation}
Then, it is easy to complete this proof by taking expectation on two sides of \eqref{mono-thm1-smooth-eq1}, and combining  with \eqref{mono-thm1-eq3} and $\ell_t^i(\x)=\langle-\tilde{\nabla} {F}_t^i(\hat{\x}_t^i), \x\rangle$.
% , for any $\x\in \mathcal{K}$, we have
% \begin{equation*}
% \begin{split}
% \E\left[\alpha\sum_{t=1}^Tf_{t}(\x)-\sum_{t=1}^Tf_{t}(\x_t^i) \right]\leq& \E\left[\sum_{t=1}^T\sum_{j=1}^n\left(\ell_t^j(\hat{\x}_t^i)-\ell_t^j(\x)\right)\right]+\E\left[\sum_{t=1}^T\sum_{j=1}^n\frac{\beta}{2}\|\hat{\x}_t^i-\hat{\x}_t^j\|^2\right]\\
% &+\E\left[\sum_{t=1}^T\sum_{j=1}^n\left\langle\nabla f_t^j(\hat{\x}_t^j)-\nabla F_t^j(\hat{\x}_t^j),\hat{\x}_t^j-\hat{\x}_t^i\right\rangle\right].
%   \end{split}
% \end{equation*}

\subsection{Proof of Corollary \ref{mono-cor3:label}}
To distinguish the decisions of Algorithms \ref{DNSCR} and \ref{AD-OSPA}, in this proof, we  denote the decision $\x_q^i$ in Algorithm \ref{AD-OSPA} for any $q\in[T/L]$ and $i\in[n]$ as $\u_q^i$, where $L$ will be specified later. This implies that $\hat{\x}_t^i$ in Algorithm \ref{DNSCR} equals to $\u_q^i$ for any $t\in \mathcal{T}_q$, where $\mathcal{T}_q=\{(q-1)L+1,\dots,qL\}$. Then, due to \eqref{mono-bound-estimated-stochastic}, and Assumption \ref{assum:bounded-set}, for any $\x\in\K$, we can apply Lemma \ref{lem1-AD-OSPF:label} to derive that
\begin{equation}
\label{mono-cor3-eq1}
\begin{split}
&\sum_{t=1}^T\sum_{j=1}^n\left(\ell_t^j(\hat{\x}_t^i)-\ell_t^j(\x)\right)=\sum_{q=1}^{T/L}\sum_{t\in\mathcal{T}_q}\sum_{j=1}^n\langle \nabla \ell_{t}^j(\u_q^j),\u_q^i-\x\rangle\\
\leq& 2\eta nR+\frac{dn\alpha^2 RTLG^2}{2\eta}+\sum_{q=1}^{T/L}\sum_{t\in\mathcal{T}_q}\sum_{j=1}^n\langle \nabla \ell_{t}^j(\u_q^j),\u_q^i-\bar{\y}_q\rangle\\
=&2\eta nR+\frac{dn\alpha^2 RTLG^2}{2\eta}+\sum_{q=1}^{T/L}\sum_{t\in\mathcal{T}_q}\sum_{j=1}^n\langle \tilde{\nabla} {F}_t^j(\u_q^j),\bar{\y}_q-\u_q^i\rangle
\end{split}
\end{equation}
where $\bar{\y}_q$ follows the definition in Lemma \ref{lem1-AD-OSPF:label}, $\eta$ will be specified later, and the first and last equalities are due to the definition of $\ell_t^j(\x)$.

By substituting \eqref{mono-cor3-eq1} into Theorem \ref{thm: mono-DSCR-smooth}, for any $\x\in\K$, we have
\begin{equation}
\label{mono-cor3-eq2}
\begin{split}
&\E\left[\alpha\sum_{t=1}^Tf_{t}(\x)-\sum_{t=1}^Tf_{t}(\x_t^i) \right]-\left(2\eta nR+\frac{dn\alpha^2 RTLG^2}{2\eta}\right)\\
\leq& 
\E\left[\sum_{q=1}^{T/L}\sum_{t\in\mathcal{T}_q}\sum_{j=1}^n\langle \tilde{\nabla} {F}_t^j(\u_q^j),\bar{\y}_q-\u_q^i\rangle\right]+\E\left[\sum_{t=1}^T\sum_{j=1}^n\frac{\beta}{2}\|\hat{\x}_t^i-\hat{\x}_t^j\|^2\right]\\
&+\E\left[\sum_{t=1}^T\sum_{j=1}^n\left\langle\nabla f_t^j(\hat{\x}_t^j)-\nabla F_t^j(\hat{\x}_t^j),\hat{\x}_t^j-\hat{\x}_t^i\right\rangle\right]\\
=& 
\E\left[\sum_{q=1}^{T/L}\sum_{t\in\mathcal{T}_q}\sum_{j=1}^n\langle {\nabla} {F}_t^j(\u_q^j),\bar{\y}_q-\u_q^j\rangle\right]+\E\left[\sum_{q=1}^{T/L}\sum_{t\in\mathcal{T}_q}\sum_{j=1}^n\frac{\beta}{2}\|\u_q^i-\u_q^j\|^2\right]\\
&+\E\left[\sum_{q=1}^{T/L}\sum_{t\in\mathcal{T}_q}\sum_{j=1}^n\left\langle\nabla f_t^j(\u_q^j),\u_q^j-\u_q^i\right\rangle\right]
  \end{split}
\end{equation}
where the equality is due to the definition of $\u_q^i$ and $\E[\tilde{\nabla} {F}_t^j(\u_q^j),\bar{\y}_q-\u_q^i]=\E[\nabla {F}_t^j(\u_q^j),\bar{\y}_q-\u_q^i]$ derived from \eqref{mono-thm1-eq3} for any $t\in\mathcal{T}_q$.

For the second term in the right side of \eqref{mono-cor3-eq2}, due to \eqref{mono-bound-estimated-stochastic}, Lemma \ref{lem2-AD-OSPF:label} and $\|\u_q^i-\u_q^j\|^2\leq 4R^2$ for any $q\leq 2$, we have
\begin{equation}
\label{mono-cor3-eq3}
\begin{split}
\E\left[\sum_{q=1}^{T/L}\sum_{t\in\mathcal{T}_q}\sum_{j=1}^n\frac{\beta}{2}\|\u_q^i-\u_q^j\|^2\right]\leq \beta nT\left(\frac{16R^2}{L}+\frac{24 d\alpha LGR^2}{\eta}\right) + 4n\beta LR^2.
\end{split}
\end{equation}
To analyze the first term in the right side of \eqref{mono-cor3-eq2}, we introduce a new lemma about $F_t^i(\x)$.
\begin{lemma}
\label{lem2:non-oblvious-mon}
(Theorem 9 of \citet{Zhang-Arxiv24}) Under Assumptions \ref{assum:bounded-set}, \ref{assum:DR-submodular}, \ref{assum:smooth}, and \ref{assum:mono}, for any $i\in[n]$ and $t\in[T]$, the function $F_t^i(\x)$ defined in Lemma \ref{lem:non-oblvious-mon} can be written as $F_t^i(\x)= \int_0^1\frac{e^{z-1}}{z}f_t^i(z\x)dz$ and it is $(\beta/e)$-smooth over $\mathcal{X}$.
\end{lemma}
Due to the oblivious assumption of $f_t^i(\x)$, the function $F_t^i(\x)$ given by Lemma \ref{lem2:non-oblvious-mon} is also oblivious. Combining this property, Lemma \ref{lem2:non-oblvious-mon}, \eqref{mono-thm1-eq4}, and \eqref{mono-bound-estimated-stochastic}, we can use Lemma \ref{lem2-AD-OSPF:label} and $\langle {\nabla} {F}_t^j(\u_q^j),\bar{\y}_q-\u_q^j\rangle\leq 2\alpha RG$ for any $q\leq 2$ to derive that
\begin{equation}
\label{mono-cor3-eq4}
\begin{split}
\E\left[\sum_{q=1}^{T/L}\sum_{t\in\mathcal{T}_q}\sum_{j=1}^n\langle {\nabla} {F}_t^j(\u_q^j),\bar{\y}_q-\u_q^j\rangle\right]\leq nT\left(\frac{4\beta R^2}{eL}+\frac{5 d\alpha^2 LRG^2}{\eta}\right)+4n\alpha RGL.
\end{split}
\end{equation}
Similarly, for any $q\geq 3$ and $t\in\mathcal{T}_q$, due to the oblivious assumption and $\beta$-smoothness of $f_t^i(\x)$, \eqref{bound-gradient}, and \eqref{mono-bound-estimated-stochastic}, we can use Lemma \ref{lem2-AD-OSPF:label} again to derive that
\begin{equation}
\label{mono-cor3-eq5}
\begin{split}
\E\left[\left\langle\nabla f_t^j(\u_q^j),\u_q^j-\u_q^i\right\rangle\right]\leq&\E\left[\left\langle\nabla f_t^j(\u_q^j),\u_q^j-\bar{\y}_q\right\rangle\right]+\E\left[\left\langle\nabla f_t^j(\u_q^j),\bar{\y}_q-\u_q^i\right\rangle\right]\\
\leq &\frac{8\beta R^2}{L}+\frac{10 d\alpha LRG^2}{\eta}.
\end{split}
\end{equation}
Combining \eqref{mono-cor3-eq5} with $\langle\nabla f_t^j(\u_q^j),\u_q^j-\u_q^i\rangle\leq 2RG$ for any $q\leq 2$, it is easy to verify that
\begin{equation*}
% \label{cor3-eq6}
\begin{split}
\E\left[\sum_{q=1}^{T/L}\sum_{t\in\mathcal{T}_q}\sum_{j=1}^n\left\langle\nabla f_t^j(\u_q^j),\u_q^j-\u_q^i\right\rangle\right]\leq nT\left(\frac{8\beta R^2}{L}+\frac{10 d\alpha LRG^2}{\eta}\right)+4nRGL.
\end{split}
\end{equation*}
Finally, by substituting \eqref{mono-cor3-eq3}, \eqref{mono-cor3-eq4}, and the above inequality into \eqref{mono-cor3-eq2}, we have
\begin{equation*}
\begin{split}
\E\left[\Reg(T,i,\alpha) \right]\leq &2\eta nR+\frac{dn\alpha RTLG}{\eta}\left(16G+24\beta R\right)+\frac{26\beta nTR^2}{L}+nRL\left(8G+4\beta R\right)\\
\leq &24n(\sqrt{2}T^{2/3}+\sqrt{TC})\sqrt{d}R^\prime+26\beta nT^{2/3}R^2+8n(2T^{1/3}+C)R^\prime
  \end{split}
\end{equation*}
where the last inequality is due to $\eta=G^\prime \sqrt{dTL}$, $G^\prime = \alpha G$, $T^{1/3}\leq L=\max\{\lceil T^{1/3}\rceil,C\}\leq 2T^{1/3}+ C$, and $R^\prime=R(G+\beta R)$. Note that these parameters are following Theorem \ref{thm1-AD-OSPF} with a simple adjustment about the upper bound on the gradient norm in \eqref{mono-bound-estimated-stochastic}.

\end{document}